\documentclass[11pt]{article}
\usepackage[utf8]{inputenc}
\usepackage[margin=1in]{geometry}

\usepackage{microtype}
\usepackage{graphicx}
\usepackage{subfigure}
\usepackage{booktabs} 
\usepackage{natbib}
\setcitestyle{open={(},close={)}}

\usepackage[colorlinks=true,citecolor=blue]{hyperref}

\usepackage{amsmath,amsthm,amsfonts,amssymb,mathdots,array,mathrsfs,bm,bbm,stmaryrd,graphicx,subfigure,xcolor}
\usepackage{bbold}
\usepackage{breakcites}
\usepackage[linesnumbered, ruled, vlined]{algorithm2e}
\let\oldnl\nl
\newcommand{\nonl}{\renewcommand{\nl}{\let\nl\oldnl}}

\usepackage[T1]{fontenc}
\usepackage{enumerate}
\usepackage{inputenc}

\usepackage{graphicx} 
\usepackage{subfigure}

\usepackage{booktabs,balance}
\usepackage{rotating}
\usepackage{boldline}
\usepackage{makecell}
\usepackage{multirow}
\usepackage{balance}

\usepackage{tikz}

\newtheorem{theorem}{Theorem}

\newtheorem{lemma}[theorem]{Lemma}

\newtheorem{definition}{Definition}[section]

\newcommand{\bsmat}{\begin{bmatrix} }
\newcommand{\esmat}{\end{bmatrix} }

\begin{document}

\title{\bf OPORP: One Permutation + One Random Projection}

\author{
\textbf{Ping Li},\ \ \textbf{Xiaoyun Li}\\\\
LinkedIn Ads\\
700 Bellevue Way NE, Bellevue, WA 98004, USA\\
 \texttt{\{pinli,\ xiaoyli\}@linkedin.com}
}
 \date{\vspace{-0.2in}}
\maketitle
  
\begin{abstract}

\noindent OPORP is an improved variant of the  count-sketch data structure by using a \textbf{fixed-length binning} scheme and a \textbf{normalization} step for the estimation.  In our experience, we find engineers like the name ``one permutation + one random projection'' as it tells the exact steps. 

\vspace{0.1in}

\noindent Consider two data vectors (e.g., embeddings): $u, v\in\mathbb{R}^D$. In many embedding-based applications  where  vectors are generated from trained models, $D=256\sim 1024$ are common and $D>1024$ is not rare (e.g., GPT models). $D$ can be much larger in  applications where the vectors are generated without  training.  With OPORP, we first apply a permutation on the data vectors. A random vector $r\in\mathbb{R}^D$ is generated with moments: $E(r_i) = 0, E(r_i^2)=1, E(r_i^3) =0, E(r_i^4)=s$. Note that $s=3$ if $r_i$ follows the standard Gaussian distribution. We multiply  $r$ (element-wise) with all permuted data vectors. Then we break the $D$ columns into $k$ equal-length bins and aggregate (i.e., sum) the values in each bin to obtain $k$ samples from each data vector. One key step is to normalize the $k$ samples to the unit $l_2$ norm. In this way, for the two  original data vectors $u, v\in\mathbb{R}^D$, we obtain two new vectors $x,y\in\mathbb{R}^D$ with unit $l_2$ norms. The  inner product of $x,y$ approximates the original correlation $\rho$ (i.e., the cosine)  between $u$ and $v$. Our main contribution  is to show that the estimation variance has essentially the following expression: 
\begin{align}\notag
(s-1)A
+ \frac{D-k}{D-1}\frac{1}{k}\left[ (1-\rho^2)^2 -2A\right],\hspace{0.2in} \text{where }\  s\geq 1 \ , \  A\geq 0. 
\end{align}This variance formula reveals several key properties of the proposed scheme and estimator:
\begin{itemize}
\item We need $s=1$, otherwise the variance has a term which does not decrease with increasing sample size $k$. There is only one such distribution: $r_i\in\{-1,+1\}$ with equal probabilities.

\vspace{0.05in}

\item The factor $\frac{D-k}{D-1}$ can be quite beneficial in reducing variances. When $k=D$, the variance is zero. When $k=D/2$ (and $s=1$), the variance is reduced by $50\%$. When $k=D/4$ (which is also a common practical situation), the variance is reduced by $25\%$. \vspace{0.05in}

\item The term $(1-\rho^2)^2$ represents another (drastic) variance reduction due to normalization, compared to $(1+\rho^2)$, which is the corresponding term without normalization. Interestingly, $(1-\rho^2)^2$ also matches the classical asymptotic variance of the classical correlation estimator. 
\end{itemize}

\vspace{0.05in}

\noindent The OPORP procedure can be repeated $m$ times to  improve the estimate. We illustrate that, by letting the $k$ in OPORP to be $k=1$  and repeat the procedure $m$ times, we exactly recover the work of ``very sparse random projections'' (VSRP)~\citep{li2006very}. This  immediately leads to a normalized  estimator for VSRP which substantially improves  the original estimator of VSRP. 

\vspace{0.15in}

\noindent In summary, the two key steps in OPORP: (i)  \textbf{normalization} and (ii)  \textbf{fixed-length binning}, have  considerably improved the accuracy in estimating the cosine similarity, which is a routine (and crucial) task in modern embedding-based retrieval (EBR) applications.  Count-sketch type of data structures have been widely used in feature hashing for AI model compression, heavy-hitter detection, efficient communication in federated learning, and privacy~\citep{li2023differential}.
\end{abstract}

\newpage

\tableofcontents

\newpage

\section{Introduction}

Given two $D$-dimensional vectors, $u,v\in\mathbb{R}^D$, a common task is to compute the ``cosine'' similarity: 
\begin{align}
\rho = \frac{\sum_{i=1}^D u_i v_i}{\sqrt{\sum_{i=1}^D u_i^2}\sqrt{\sum_{i=1}^D v_i^2}}. 
\end{align}
Some applications also need to compute the inner product $a$ and the $l_2$ distance $d$: 
\begin{align}
a = \sum_{i=1}^D u_iv_i,\hspace{0.4in} d = \sum_{i=1}^D |u_i - v_i|^2.
\end{align}
The data vectors can be the ``embeddings'' learned from deep learning models such as the celebrated ``two-tower'' model~\citep{huang2013learning}. They can also be data vectors processed without training, for example, the $n$-grams (shingles), which can be extremely high-dimensional, e.g., $D$ is million or billion or even higher depending on the choice of ``$n$'' in $n$-grams~\citep{broder1997resemblance,broder1997syntactic,li2005using, das2007google,chierichetti2009compressing,li2008one,li2012one,tamersoy2014guilt,nargesian2018table,wang2019memory,li2022c}. 

\vspace{0.1in}

It is often the case that the embedding vectors generated from deep learning models are relatively short (e.g., $D=256$ or $D=1024$), often dense, and  typically normalized, i.e., $\sum_{i=1}^D u_i^2 = \sum_{i=1}^D v_i^2 = 1$.   (In this study, we will not assume the original data vectors are normalized.)  For example, for BERT-type of embeddings~\citep{devlin2019bert}, the embedding size $D$ is typically 768 or 1024; and Applications with BERT models may also use higher embedding dimensions, e.g., $D=4096$~\citep{giorgi2021declutr}. For GLOVE word embeddings~\citep{pennington2014glove}, $D=300$ is often the default choice. In recent EBR (embedding based retrieval) applications~\citep{chang2020pre,yu2022egm,yu2022boost}, using $D=256$ or $D=512$ appears common.  For knowledge graph embeddings, we see the use of embedding size $D=256\sim 768$~\citep{huang2019knowledge,spillo2022knowledge}. In many computer vision applications, the embedding sizes are often  larger, e.g., 4096, 8192 or even larger~\citep{karpathy2014deep,yu2018hierarchical,lanchantin2021general}. The recent advances in GPT-3 models~\citep{brown2020language} for NLP tasks (text classification, semantic search, etc.) learn word embeddings with $D=1024\sim 12288$~\citep{neelakantan2022text}. 

\vspace{0.1in}

In practical scenarios, the cost for storing the embeddings is usually expensive. In fact, even with merely $D= 256$, the storage cost for the embeddings can be prohibitive in industrial  applications. For example, suppose an app has 100 million (active) users and each user is represented by a $D=256$ embedding vector. Then storing the embeddings (assuming each dimension is a 4-byte real number) would cost 100GB.  It will make the deployment much easier if the storage can be reduced to, say 25GB (a 4-fold reduction) or 12.5GB (a 8-fold reduction). Reducing the embedding size will, of course, also translate into the reductions in the computational and communication costs.  

\vspace{0.1in}

In this paper, we study a compression scheme based on the idea of ``one permutation + one random projection'', or OPORP for short. It basically uses the (variant of) count-sketch data structure~\citep{charikar2004finding}, with several differences: (i) we focus on one permutation and one random projection (while it is straightforward to extend the analysis to multiple projections); (ii) we use a \textbf{fixed-length binning} scheme; (iii) we adopt a \textbf{normalization} step in the estimation stage. Compared with the previous works~\citep{weinberger2009feature,li2011hashing} which used count-sketch type data structures for building large-scale machine learning models, the normalization step very significantly reduces the estimation variance, as shown by our theoretical analysis. In addition, the fixed-binning scheme brings in a multiplicative term $\frac{D-k}{D-1}$ in the variance which also substantially reduces the estimation error when $k=D/2$ (i.e., a 50\% variance reduction) or even just $k=D/4$. 

\newpage

\subsection{Count-Sketch and Variants}

We briefly review the count-sketch data structure~\citep{charikar2004finding}. Count-sketch first uses a hash function $h:[D]\mapsto [k]$ to uniformly map each data coordinate to one of $k$ bins, and then aggregates the coordinate values within the bin. Here, each coordinate $i\in[1,D]$ is further multiplied by a  Rademacher variable $r_i$ with $P(r_i=-1)=P(r_i=1)=1/2$. The binning procedure of count-sketch can be interpreted, in a probabilistically equivalent manner, as the ``variable-length binning scheme''. That is, we first apply a random permutation on the data vector and splits the coordinates into $k$  bins whose lengths follow a multinomial distribution. Also, in the original count-sketch, the above procedure is repeated $m$ times (for identifying heavy hitters, another term for ``compressed sensing''). The count-sketch data structure and variants have been widely used in  applications. Recent examples include graph embedding~\citep{wu2019demo}, word \& image embedding~\citep{chen2017reading,zhang2020dynamic,abdullah2021finding,singhal2021federated,zhang2022kernelized},  model \& communication compression~\citep{weinberger2009feature,li2011hashing, chen2015compressing,rothchild2020fetchsgd,haddadpour2020fedsketch}, etc. Note that in many  applications, only $m=1$ repetition is used. Our study will  focus on $m=1$ and the analysis can be extended to $m>1$. In fact, we can recover ``very sparse random projections'' (VSRP)~\citep{li2006very} if we let $m>1$ (and $k=1$, i.e., using just one bin for each repetition). This is an interesting insight/connection.

\subsection{Random Projections (RP) and Very Sparse Random Projections (VSRP)}

To a large extent, the work of OPORP is also closely related to  random projections (RP), especially the ``sparse'' or ``very sparse'' random projections~\citep{achlioptas2003database,li2006very}. The basic idea of random projections is to multiply the original data vectors, e.g., $u\in\mathbb{R}^D$ with a random  matrix $R\in\mathbb{R}^{D\times k}$ to generate new vectors, e.g., $x\in\mathbb{R}^k$, as samples from which we can recover the original similarities (e.g., the inner products or cosines). The entries of the random  matrix $R$ are typically sampled i.i.d. from the standard Gaussian distribution or the Rademacher distribution.  The projection matrix can also be made (very) sparse to facilitate the  computation. For instance, the entries  in $R$ take values in $\{-1,0,1\}$ with probabilities $\{1/(2s), 1-1/s, 1/(2s)\}$,  and we can control the sparsity by altering $s$. In many cases, $R$ can be considerably sparse while maintaining good learning capacity/utility. For example, in our experiments (Section~\ref{sec:experiment}), the learning performance does not drop much when the projection matrix contains around 90\% zeros (i.e., $s=10$) on average. 

As an effective tool for dimensionality reduction and geometry preservation, the methods of  (very sparse) random projections have been widely adopted by numerous applications in data mining, machine learning, computational biology, databases, compressed sensing, etc.~\citep{johnson1984extensions,goemans1995improved,dasgupta2000experiments,bingham2001random,buher2001effcient,charikar2002similarity,fern2003random,achlioptas2003database,datar2004locality,candes2006robust,donoho2006compressed,li2006very,rahimi2007random,dasgupta2008random,li2014coding,li2019random,li2019generalization,rabanser2019failing,tomita2020sparse,li2021one}.

\subsection{OPORP versus VSRP}

We will demonstrate that we can utilize OPORP to recover ``very sparser random projections'' (VSRP)~\citep{li2006very}. Basically, we have the option of repeating the OPORP procedure $m$ times, which will reduce the variance while increasing the sample size. Interestingly, by using $m$ repetitions and letting the $k$ (number of bins) in OPORP to be $k=1$, we exactly recover VSRP with $m$ projections. This means that the theory we develop for OPORP also applies to VSRP. In particular, we immediately obtain the normalized estimator for VSRP and its theoretical variance. Therefore, OPORP and VSRP are the two extreme examples of the family of (sparse) random projections. In this paper, we show that with merely $m=1$ repetition, OPORP has already achieved smaller variances than the standard random projections and very sparse random projections. If we hope to achieve the same level of sparsity of the projection matrix, OPORP could be substantially more accurate than VSRP (depending on data distributions).

\section{The Proposed Algorithm of OPORP}

As the name ``OPORP'' suggests, the proposed algorithm mainly consists of applying ``one permutation'' then ``one random projection'' on the data vectors $\in\mathbb{R}^D$, for the purpose of reducing the dimensionality, the memory/disk space, and the computational cost. The dimensionality  $D$ varies significantly, depending on applications. As discussed in the Introduction, for embedding vectors generated from learning models,  using $D=256\sim 1024$ is fairly common although some applications use $D=8192$ or even larger.  As long as the embedding size $D$ is not too large, it is affordable (and convenient) to simply generate and store the permutation vector and the random projection vector. In fact, even when $D$ is as large as a billion ($D=10^9$), storing two $10^9$-dimensional dense vectors is often affordable. On the other hand, for applications which need $D\gg 10^9$,  we might have to resort to various approximations to generate the permutation/projection vectors such as the standard ``universal hashing''~\citep{carter1977universal}. In particular, in the literature of ($b$-bit) minwise hashing and related techniques~\citep{broder1997resemblance,broder1997syntactic,broder1998min, indyk1999sublinear,charikar2002similarity,li2008one, li2012one, shrivastava2016simple, li2022c}, there are abundance of discussions about generating (high quality) permutations in extremely high-dimensional space. In this paper, we will simplify the discussion by assuming a random permutation vector and a random projection vector.

\subsection{The Procedure of OPORP}

In summary, the procedure of OPORP has the following steps: 
\begin{itemize}
\item Generate a permutation $\pi: [D] \longrightarrow [D]$.
\item Apply the same permutation to all data vectors, e.g., $u, v\in\mathbb{R}^D$. 
\item Generate a random vector $r$ of size $D$, with i.i.d. entries $r_i$ of the following first four moments: 
\begin{align}
E(r_i) = 0,\hspace{0.3in}  E(r_i^2) = 1,\hspace{0.3in}  E(r_i^3) = 0,\hspace{0.3in}    E(r_i^4) = s.
\end{align}
Our calculation will show that $s=1$ leads to the smallest variance of OPORP. There exists only one such distribution if we let $s=1$, that is, $r_i\in\{-1, 1\}$ with equal probabilities, i.e., the Rademacher distribution.  We carry out the calculations for general $s$, for the convenience of  comparing with ``very sparse random projections'' (VSRP)~\citep{li2006very}. In fact, this will also develop the new theory and estimator for VSRP.

\item Divide the $D$ columns into $k$ bins. There are two binning strategies:
\begin{enumerate}
    \item Fixed-length binning scheme: every bin has a length of $D/k$. We  assume  $D$ is divisible by $k$; if not, we can always pad zeros. In the practice of embedding-based retrieval (EBR), as it is often to let $D=2^8 = 256$ or $D=2^{10}=1024$,  we can conveniently choose (e.g.,) $k=2^6=64$. Our  analysis will show that using this fixed-length scheme will result in a variance reduction by a factor of $\frac{D-k}{D-1}$, which is quite significant for typical EBR applications, compared to the commonly analyzed variable-length binning scheme.  \vspace{0.1in}
    
    \item Variable-length binning scheme: the bin lengths follow a multinomial distribution\\ $multinomial(D, 1/k, 1/k, ...., 1/k)$ with $k$ bins. Note that $k$ can be larger than $D$, i.e., some bins will be empty. When $k\rightarrow \infty$, it essentially recovers the fixed-length binning scheme with $k=D$. The variable-length binning  scheme is the strategy in the previous literature~\citep{charikar2004finding,weinberger2009feature,li2011hashing}. 
\end{enumerate}

\item For each bin and each data vector, we generate a sample as follows: 
\begin{align}
    x_j = \sum_{i=1}^{D} u_i r_i I_{ij},\hspace{0.3in}
    y_j = \sum_{i=1}^{D} v_i r_i I_{ij}, \hspace{0.2in} j = 1, 2,  ..., k
\end{align}
where $I_{ij}$ is an indicator: $I_{ij}=1$ if the original coordinate $i$ is mapped to bin $j$, and $I_{ij}=0$ otherwise. Because there are two binning schemes, wherever necessary, we will use $I_{1,ij}$ (fixed-length) and $I_{2,ij}$ (variable-length) to differentiate these two binning schemes. 
\end{itemize}

\vspace{0.2in}

After we have obtained the samples (e.g., $x_j$, $y_j$), we can estimate the inner product $a$, the $l_2$ distance $d$, and the cosine $\rho$ of the original data vectors, as follows:
\begin{align}
\hat{a} = \sum_{j=1}^k x_jy_j,\hspace{0.2in} \hat{d} = \sum_{j=1}^k |x_j - y_j|^2, \hspace{0.2in} \hat{\rho} = \frac{\sum_{j=1}^k x_jy_j}{\sqrt{\sum_{j=1}^k x_j^2}\sqrt{\sum_{j=1}^k y_j^2}}.
\end{align}
Note that, for the  estimator $\hat{\rho}$, the normalization step is not needed at the estimation time if we pre-normalize and store the data, e.g., $x_j^\prime = \frac{x_j^\prime}{\sqrt{\sum_{t=1}^kx_t^2}}$. This is a notable advantage. Also,  wherever necessary, we will again use $\hat{a}_1$, $\hat{a}_2$, $\hat{d}_1$, $\hat{d}_2$, $\hat{\rho}_1$, $\hat{\rho}_2$, to differentiate the two binning schemes. We should mention  that we will not assume the original data vectors are normalized to the unit $l_2$ norms, although in the practice of embedding-based retrieval (EBR), the embedding vectors are typically normalized. 

\vspace{0.1in}

If the original data vectors ($u$, $v$) are normalized, then $\hat{a}$ also provides an estimate of the original cosine because the original inner product is identical to the cosine in normalized data. One of the main contributions in this paper is to show that using $\hat{\rho}$ would be substantially more accurate than using $\hat{a}$ even when the original data are already normalized. Basically, the estimation variance of $\hat{\rho}$ is proportional to $(1-\rho^2)^2$ while the estimation of $\hat{a}$ (in normalized data) is proportional to $1+\rho^2$. The difference between $(1-\rho^2)^2$ and  $1+\rho^2$ can be highly substantial, especially for $|\rho|$ close to 1.

\subsection{The Choice of $r$} 

For the random projection vector $r\in\mathbb{R}^D$, we have only specified that its entries are i.i.d. and obey the following moment conditions:  
\begin{align}\notag
E(r_i) = 0,\hspace{0.3in}  E(r_i^2) = 1,\hspace{0.3in}  E(r_i^3) = 0,\hspace{0.3in}    E(r_i^4) = s, \hspace{0.1in} s\geq 1.
\end{align}
Note that $s\geq 1$ is needed because $E(r_i^4)\geq E^2(r_i^2)=1$ (the Cauchy-Schwarz Inequality). Typically, users who are familiar with random projections might attempt to sample $r$ from the Gaussian distribution. Our analysis, however, will show that the Gaussian distribution should not be used for OPORP. This is quite different from the standard random projections for which using either the Gaussian distribution or the Rademacher distribution (i.e., $r_i\in \{-1, +1\}$ with equal probabilities) would not make an essential difference. For OPORP, our analysis will show that we should use $s=1$ (i.e., the Rademacher distribution), by carrying out the calculations for general $s\geq 1$.

\newpage

Here, we list some common distributions, which satisfy the moment conditions, as follows: 
\begin{itemize}
\item The standard Gaussian distribution $N(0,1)$. This is the popular choice in the literature of random projections. The fourth moment of the standard Gaussian is 3, i.e., $s=3$.

\vspace{0.1in}

\item The uniform distribution, $\sqrt{3}\times unif[-1,1]$. We need the $\sqrt{3}$ factor in order to have $E(r_i^2)=1$. For this choice of distribution, we have $E(r_i^4)=s=9/5$. 

\vspace{0.1in}

\item The ``very sparse'' distribution, as used in~\citet{li2006very}: 
\begin{align}\label{eqn:vsrp}
r_i = \sqrt{s}\times \left\{\begin{array}{rrl}
-1 & \text{with prob.} &1/(2s)\\
0 & \text{with prob.} &1-1/s\\
+1 & \text{with prob.} &1/(2s)
\end{array}\right.
\end{align}
which generalizes~\citet{achlioptas2003database} (for $s=1$ and $s=3$).
\end{itemize}


\subsection{Comparison with Very Sparse Random Projections (VSRP)}

Note that for OPORP, even though it only effectively uses just one random projection, we can still view that as a random projection ``matrix'' $\in\mathbb{R}^{D\times k}$ with exactly one 1 on each row.   In comparison, the ``very sparse random projections'' (VSRP)~\citep{li2006very} uses a random projection matrix $\in\mathbb{R}^{D\times k}$ with entries sampled i.i.d. from the ``very sparse'' distribution~\eqref{eqn:vsrp}. Interestingly, for VSRP, if we let its ``$s$'' parameter to be $s=k$, then OPORP (with its $s=1$) and VSRP will have the same sparsity on average in the projection ``matrix''. In terms of the implementation, suppose we store the projection matrix, then it would be much more convenient to store the one projection vector for OPORP because it is really just a vector of length $D$. In comparison, storing the sparse random projection matrix would incur an additional overhead because we will have to store the locations (coordinates) of each non-zero entries. Thus, OPORP would be much more convenient to use. 

\vspace{0.1in}

In terms of the estimation variance, OPORP (with $s=1$) would be more accurate than VSRP, for several reasons. Firstly, OPORP with the fixed-length binning scheme has the $\frac{D-k}{D-1}$ variance reduction term, as will be shown in our theoretical analysis. Secondly, if we do not consider the $\frac{D-k}{D-1}$ term and we choose $s=1$ for both OPORP and VSRP, then their theoretical variances are identical for the un-normalized estimators. As long as $s>1$ for VSRP, the theoretical variance is larger than that of OPORP (for $s=1$). If we choose $s=k$ for VSRP (to achieve the same average sparsity as OPORP), then its variance might be significantly much larger, depending on the original data (e.g., $u$ and $v$). In addition, in this paper, we derive the variance formula for the normalized estimator of OPORP, which substantially improves the un-normalized estimator. 

\vspace{0.1in}

Finally, we should mention that we can actually recover VSRP if we  just use one bin for OPORP and repeat the procedure $k$ times. This means that theory and estimators we develop for OPORP can be directly utilized to develop new theory and new estimator for VSRP. In particular, the normalized estimator for VSRP is developed whose variance can be directly inferred from OPORP.

\vspace{0.1in}

In summary, OPORP and VSRP can be viewed as the two extreme examples of a family of sparse random projections. OPORP is more convenient to use and can be substantially more accurate than VSRP especially if we hope to maintain the same level of sparsity for the projection matrix.

\newpage

\section{Theoretical Analysis of OPORP and Numerical Verification}

In this section, we conduct the theoretical analysis to derive the estimation variances for OPORP. Recall that, 
 we generate $k$ samples as follows
\begin{align}\notag
    x_j = \sum_{i=1}^{D} u_i r_i I_{ij},\hspace{0.3in}
    y_j = \sum_{i=1}^{D} v_i r_i I_{ij}, \hspace{0.2in} j = 1, 2,  ..., k
\end{align}
where $I_{ij}$ is a random variable determined by one of the following  two binning schemes: 
\begin{enumerate}
    \item ({\em First binning scheme}) Fixed-length binning scheme: every bin has a length of $D/k$. We assume that $D$ is divisible by $k$, if not, we can always pad zeros.
    \item ({\em Second binning scheme}) Variable-length binning scheme: the bin lengths follow a multinomial distribution $multinomial(D, 1/k, 1/k, ...., 1/k)$ with $k$ bins. 
\end{enumerate}
Specifically, $I_{ij} = 1$ if the original coordinate $i\in[1,D]$ is mapped to bin $j\in[1,k]$; $I_{ij}=0$ otherwise. Wherever necessary, we will use $I_{1,ij}$ and $I_{2,ij}$ to differentiate the two binning schemes. 

\begin{lemma}\label{lem:I}
$\forall i\in[1,D]$, $j\in[1,k]$, $i\neq i^\prime$, $j\neq j^\prime$, 
\begin{align*}
&E(I_{1,ij}) = E(I_{1,ij}^n) = \frac{1}{k}, n=1,2,3,...\\
&E(I_{1,ij}I_{1,ij^\prime}) = 0,\\
&E(I_{1,ij}I_{1,i^\prime j^\prime}) = \frac{D}{D-1}\frac{1}{k^2},\\
&E(I_{1,ij}I_{1,i^\prime j}) = \frac{D-k}{D-1}\frac{1}{k^2},
\end{align*}
\begin{align*}
&E(I_{2,ij}) = E(I_{2,ij}^n) =  \frac{1}{k}, n=1,2,3,...\\
&E(I_{2,ij}I_{2,ij^\prime}) = 0,\\
&E(I_{2,ij}I_{2,i^\prime j^\prime}) = \frac{1}{k^2},\\
&E(I_{2,ij}I_{2,i^\prime j}) = \frac{1}{k^2},
\end{align*}
 \begin{align*}
kE\left( I_{ij} I_{i^\prime j}\right) + k(k-1)E\left(I_{ij}I_{i^\prime j^\prime}\right) = 1,
\end{align*}
\end{lemma}

\noindent\textbf{Proof of Lemma~\ref{lem:I}:} Consider the first binning scheme, where all $k$ bins have the same length $D/k$. Thus, $E(I^n_{1,ij}) = E(I_{1,ij}) =\frac{D/k}{D} = \frac{1}{k}$. Each coordinate $i$ can only be mapped to one bin, hence $E(I_{1,ij}I_{1,ij^\prime}) = 0, \forall j\neq j^\prime$.  To understand $E(I_{1,ij}I_{1,i^\prime j^\prime}) = \frac{1}{k}\frac{D/k}{D-1} = \frac{D}{D-1}\frac{1}{k^2}$, we first assign $i$ to $j$ which occurs with probability $1/k$; then assign $i^\prime$ to $j^\prime$, which occurs with probability $\frac{D/k}{D-1}$ because the bin length is $D/k$ and there are $D-1$ locations left (as one is taken). Finally, to understand $E(I_{1,ij}I_{1,i^\prime j}) = \frac{1}{k}\frac{D/k-1}{D-1} = \frac{D-k}{D-1}\frac{1}{k^2}$, we only have $D/k-1$ (instead of $D/k$) choices because one location in bin $j$ is already taken. 

Next, we consider the second binning scheme. As the $k$ bin lengths follow the multinomial distribution, the results follow using properties of multinomial moments after some algebra. $\hfill\qed$

\subsection{The Un-normalized Estimators}

Once we have samples $x_j$, $y_j$, we can estimate the original inner product $a$ by $\hat{a} = \sum_{j=1}^k x_jy_j$.  The results in Lemma~\ref{lem:I} can assist us to derive the variances of the inner product estimators, $\hat{a}_1$ and $\hat{a}_2$ for two binning schemes, respectively. 

\begin{theorem}\label{thm:a}
\begin{align*}
&E(\hat{a}) = a, \\
&Var(\hat{a}_1)
=\left(s-1\right) \sum_{i=1}^D u_i^2 v_i^2 + \frac{1}{k}\left(a^2+ \sum_{i=1}^Du_i^2 \sum_{i=1}^Dv_i^2 -2\sum_{i=1}^D u_i^2v_i^2\right)\frac{D-k}{D-1}, \\
&Var(\hat{a}_2)
=\left(s-1\right) \sum_{i=1}^D u_i^2 v_i^2 +\frac{1}{k}\left(a^2+ \sum_{i=1}^Du_i^2 \sum_{i=1}^Dv_i^2 -2\sum_{i=1}^D u_i^2v_i^2\right).
\end{align*}
\end{theorem}
\noindent\textbf{Proof of Theorem~\ref{thm:a}:}  See Appendix~\ref{proof:thm:a}. $\hfill\qed$

\vspace{0.2in}

Compared to  $Var(\hat{a}_2)$ for the variable-bin-length scheme (which appeared in the prior work~\citep{li2011hashing}), the additional factor $\frac{D-k}{D-1}$ in $Var(\hat{a}_1)$ demonstrates the benefit of the proposed fixed-bin-length strategy. Also, it is clear that we should choose $s=1$.  What if we only use one bin, i.e., $k=1$? In this case $\frac{D-k}{D-1} = 1$, i.e., two binning scheme becomes identical. This is of course expected and also explains why in $\frac{D-k}{D-1}$ we have $D-1$ instead of just $D$.

\vspace{0.2in}

What will happen if we repeat OPORP $m$ times? In that case, the variances will be reduced by a factor of $\frac{1}{m}$, i.e., 
\begin{align*}
&Var(\hat{a}_1; m \text{ repetitions})
=\frac{1}{m}\left[\left(s-1\right) \sum_{i=1}^D u_i^2 v_i^2 + \frac{1}{k}\left(a^2+ \sum_{i=1}^Du_i^2 \sum_{i=1}^Dv_i^2 -2\sum_{i=1}^D u_i^2v_i^2\right)\frac{D-k}{D-1}\right], \\
&Var(\hat{a}_2; m \text{ repetitions})
=\frac{1}{m}\left[\left(s-1\right) \sum_{i=1}^D u_i^2 v_i^2 +\frac{1}{k}\left(a^2+ \sum_{i=1}^Du_i^2 \sum_{i=1}^Dv_i^2 -2\sum_{i=1}^D u_i^2v_i^2\right)\right].
\end{align*}

\vspace{0.2in}

Furthermore,  if we let $k=1$ and still repeat $m$ times, then the two estimators become the same one and the variance would be
\begin{align*}
&Var(\hat{a}; m \text{ repetitions and } k=1)
=\frac{1}{m}\left(a^2+ \sum_{i=1}^Du_i^2 \sum_{i=1}^Dv_i^2 +(s-3)\sum_{i=1}^D u_i^2v_i^2\right),
\end{align*}
which  is exactly the variance formula for the inner product estimator of ``very sparse random projections'' (VSRP)~\citep{li2006very}. In retrospect, this is  expected because with $k=1$ for OPORP and $m$ repetitions, we recover the regular random projections with a projection matrix of size $D\times m$. We can also change the notation from $D\times m$ to $D\times k$ if the latter is more familiar to readers.

Once we have the variances for the inner products, it is straightforward to derive the variances for the distance estimators. To see this, 
\begin{align}\notag
\hat{d} = \sum_{j=1}^k |x_j - y_j|^2 = \sum_{j=1}^k \left|\sum_{i=1}^D u_i r_i I_{ij} - \sum_{i=1}^D v_i r_i I_{ij}\right|^2 =  \sum_{j=1}^k \left|\sum_{i=1}^D (u_i - v_i) r_i I_{ij}\right|^2.
\end{align}
Clearly,  we just need to replace, in Theorem~\ref{thm:a}, both $u_i$ and $v_i$ by $u_i-v_i$, in order to derive Theorem~\ref{thm:d}. 

\begin{figure}[h!]
    \centering

\mbox{    
    \includegraphics[width=2.7in]{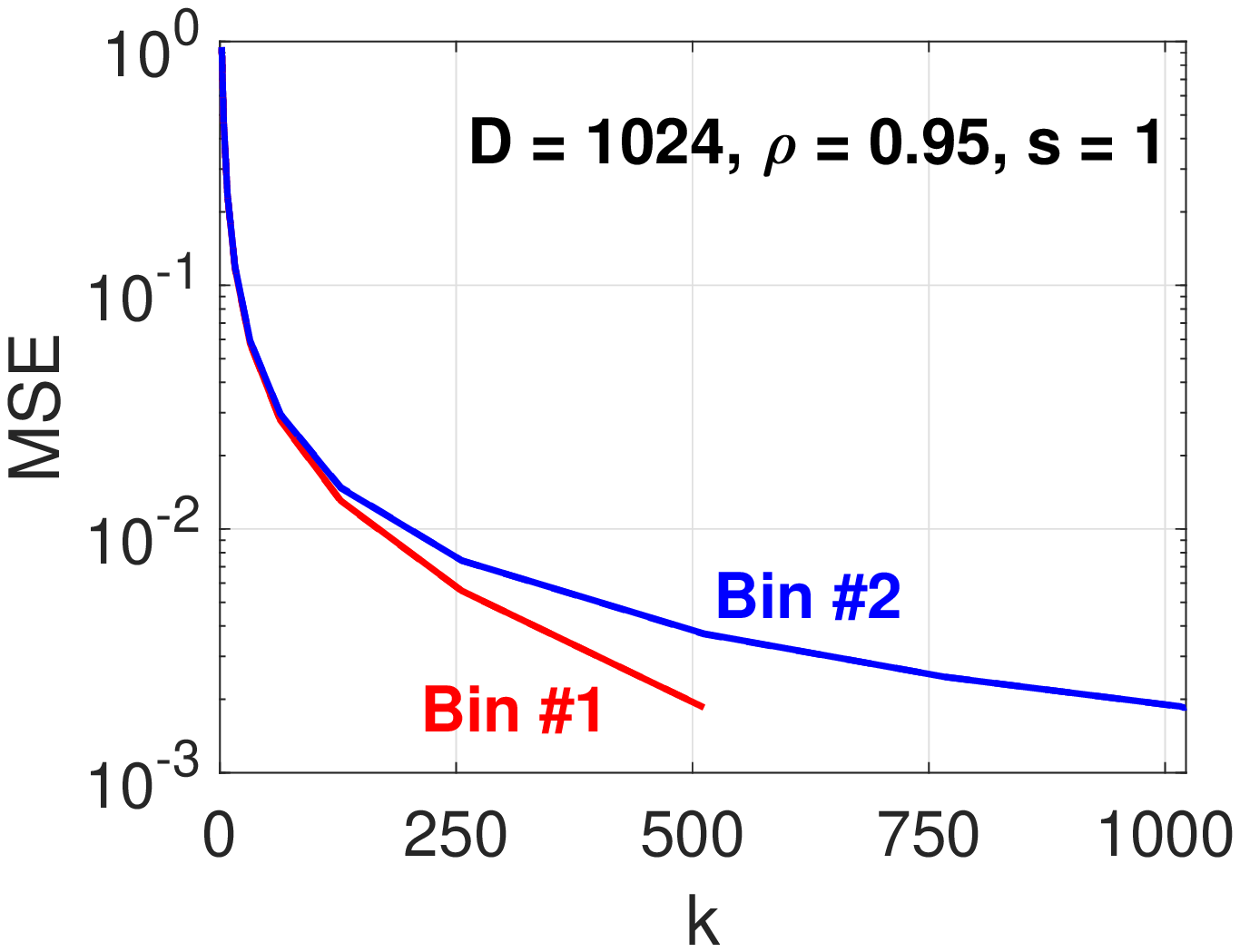}
    \includegraphics[width=2.7in]{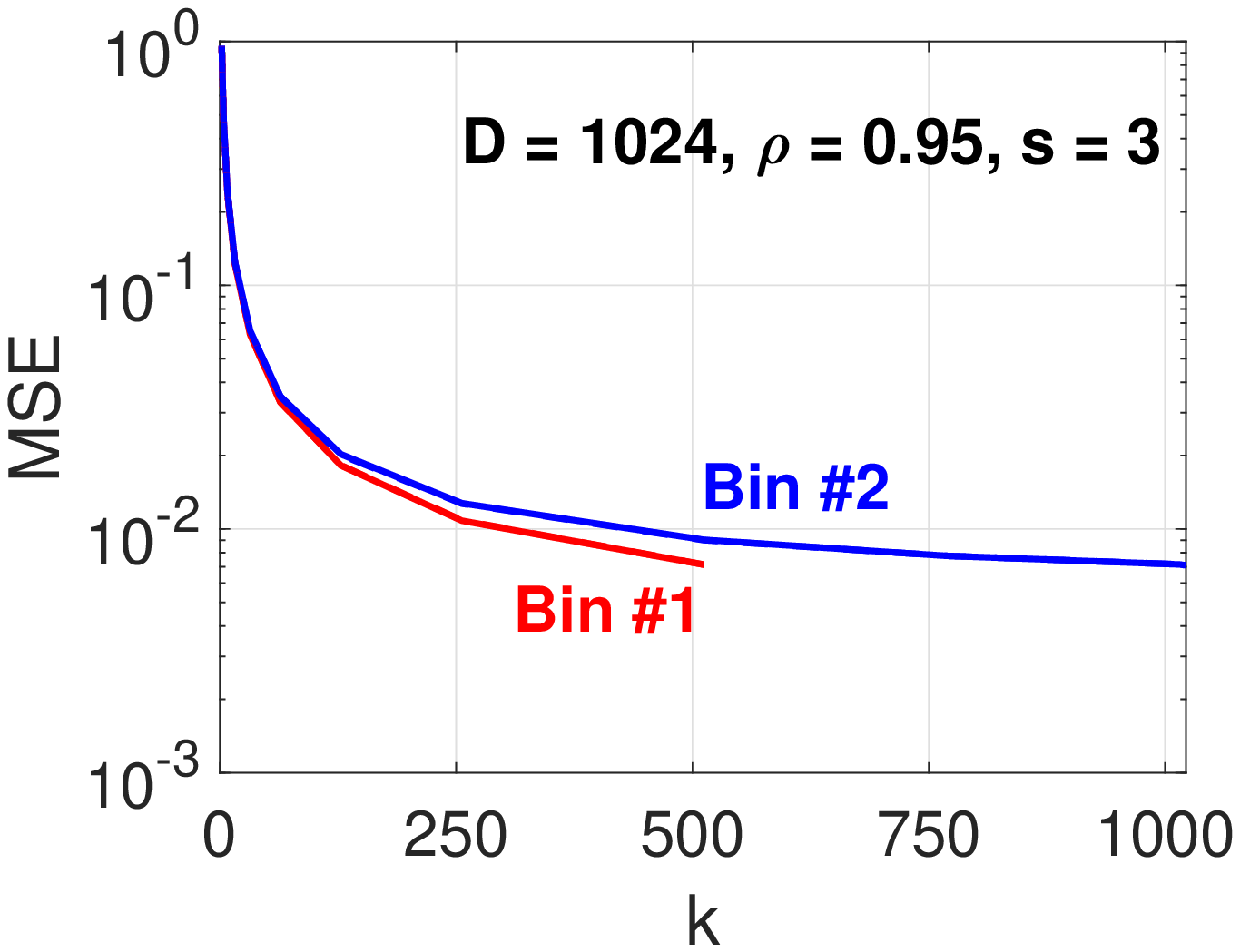}
    }
\mbox{    
    \includegraphics[width=2.7in]{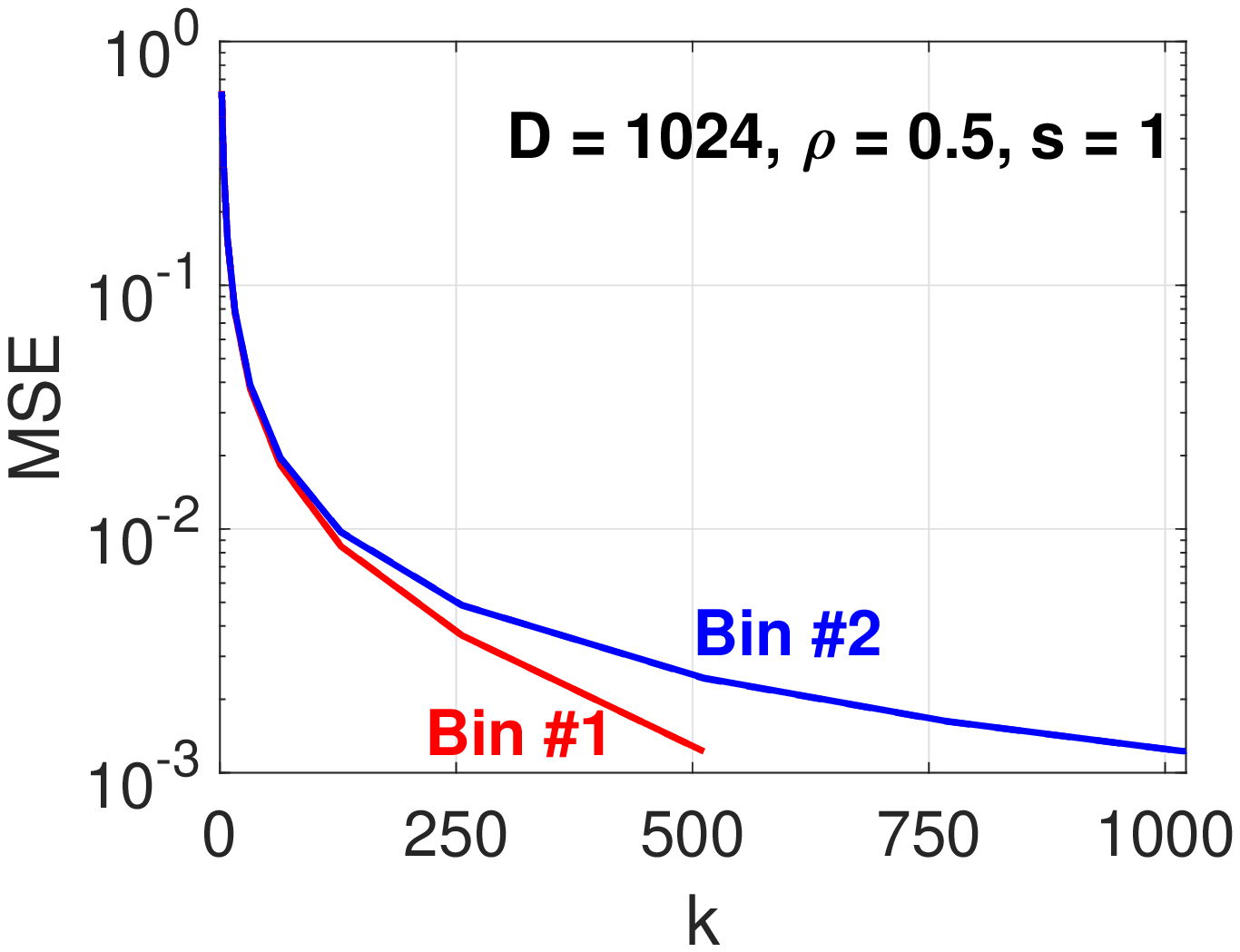}
    \includegraphics[width=2.7in]{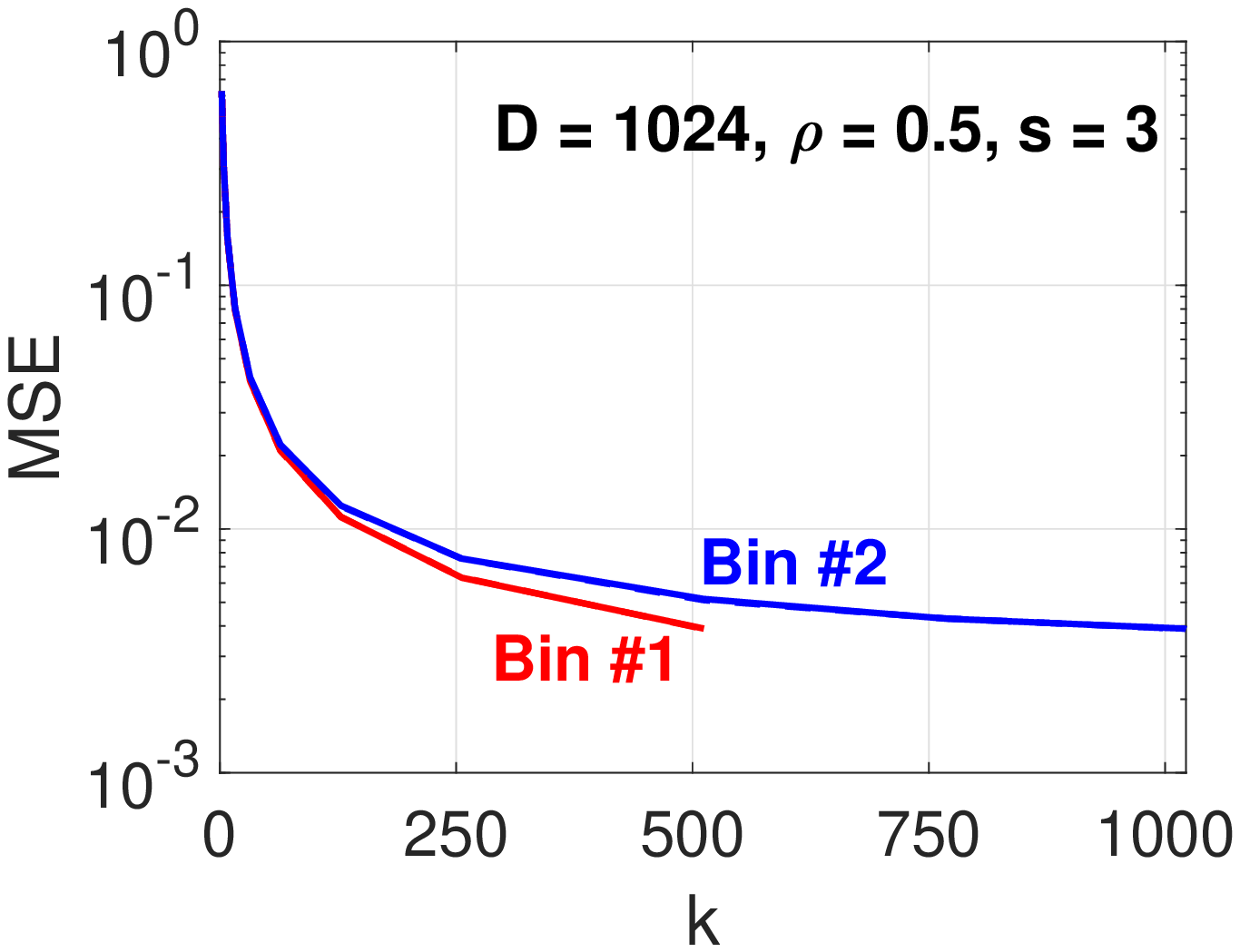}
    }    
\mbox{    
    \includegraphics[width=2.7in]{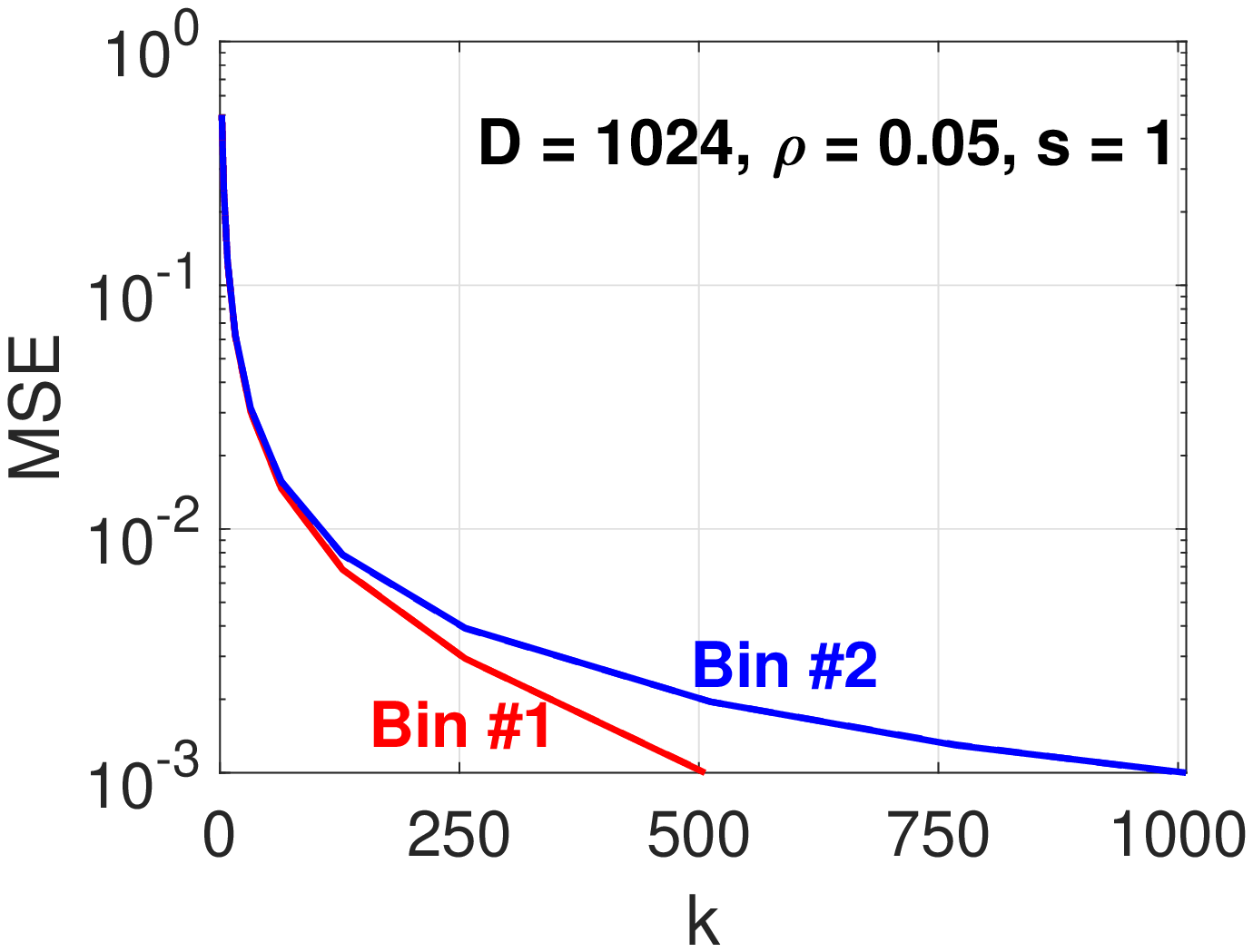}
    \includegraphics[width=2.7in]{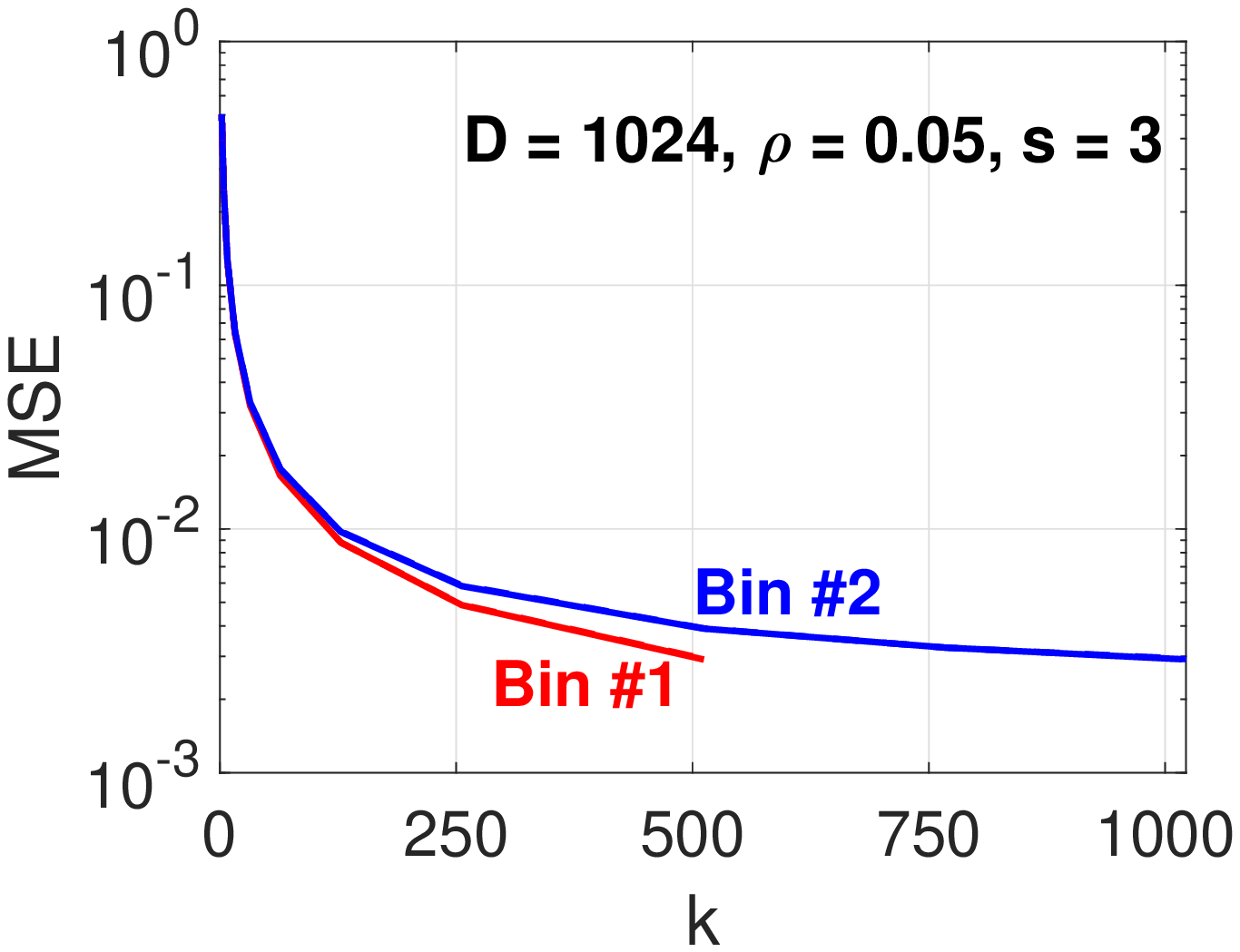}
    }        
\mbox{    
    \includegraphics[width=2.7in]{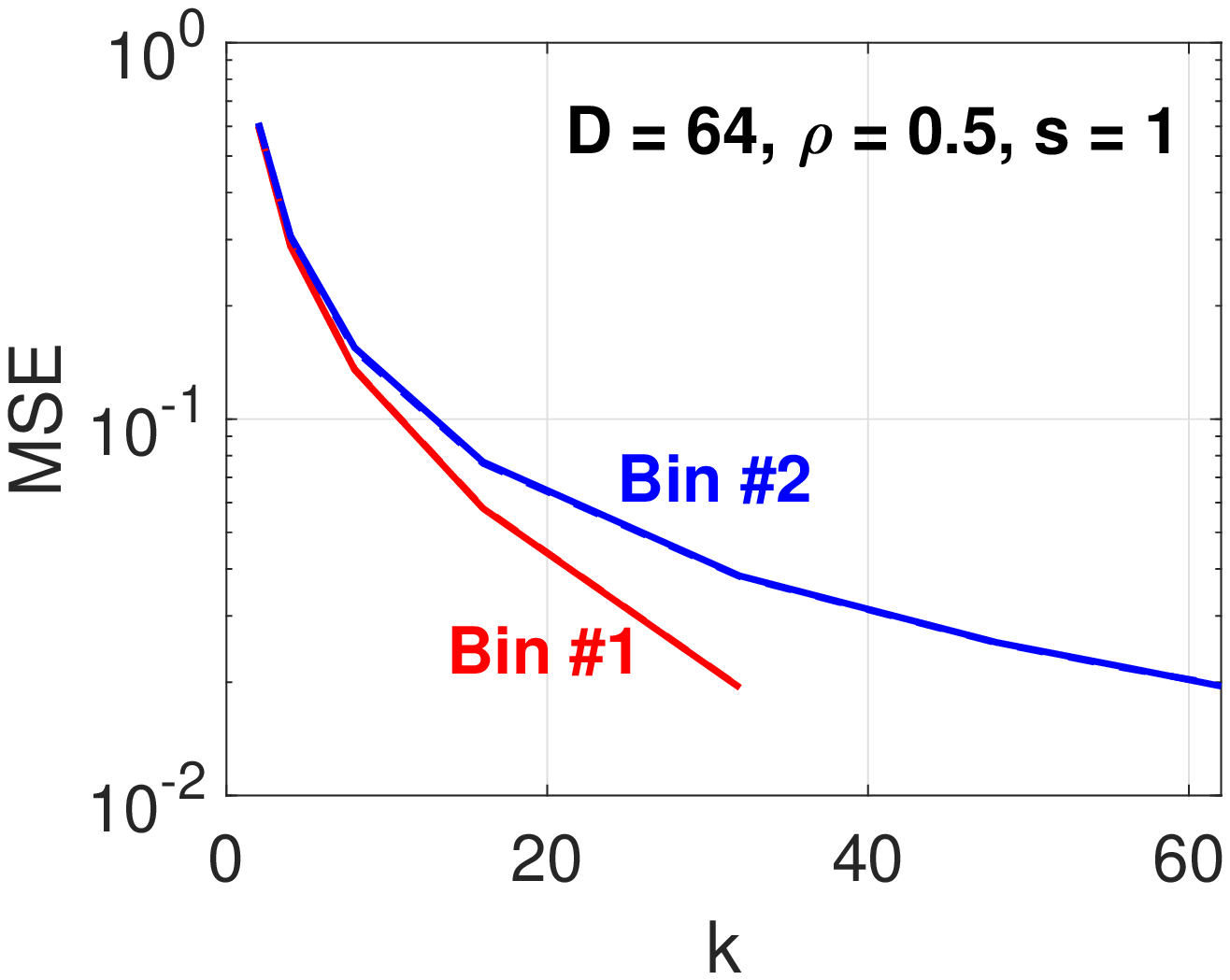}
    \includegraphics[width=2.7in]{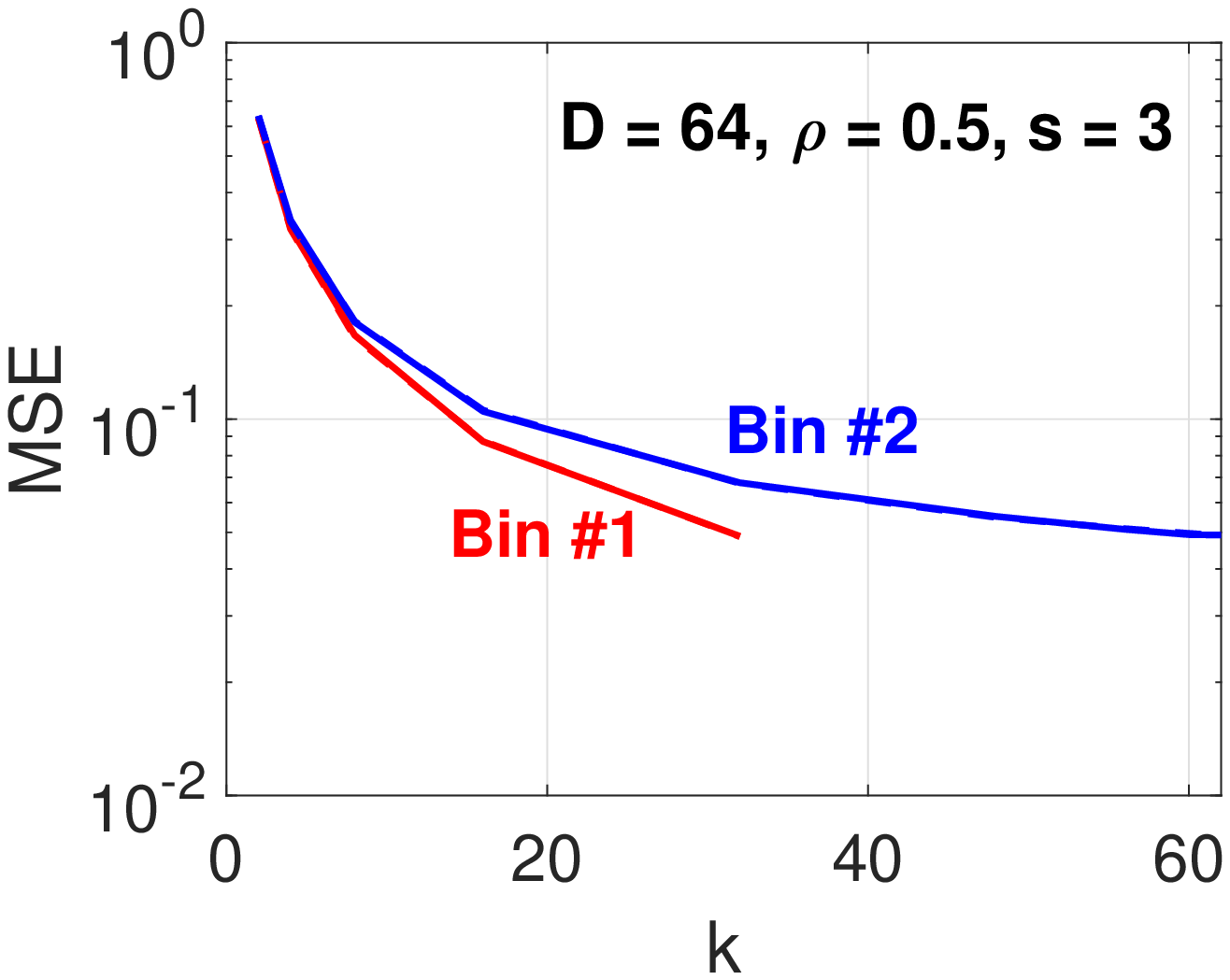}
    }    
    
\vspace{-0.2in}

    \caption{In each panel, we simulated two (normalized) vectors with the target $\rho$ value. Then we conduct OPORP $10^5$ times for each $k$, and both binning schemes. In each panel, the two solid curves represent the empirical mean square errors (MSE) and the two dashed curves for the theoretical variances. The dashed curves are not visible because they overlap with the solid curves. Note that for the fixed-length binning scheme (``Bin \#1''), we cannot choose a $k$ in between $D/2$ and $D$.}
    \label{fig:a}
\end{figure}

\newpage\clearpage 

\begin{theorem}\label{thm:d}
\begin{align*}
&E(\hat{d}) = d, \\
&Var(\hat{d}_1) = 
\left(s-1\right) \sum_{i=1}^D |u_i-v_i|^4 + \frac{1}{k}\left(2d^2- 2\sum_{i=1}^D|u_i-v_i|^4\right)\frac{D-k}{D-1}, \\
&Var(\hat{d}_2) = 
\left(s-1\right) \sum_{i=1}^D |u_i-v_i|^4 +\frac{1}{k}\left(2d^2- 2\sum_{i=1}^D|u_i-v_i|^4\right).
\end{align*}
\end{theorem}

\vspace{0.2in}

In the variance formulas, the term  $\frac{D-k}{D-1}$ of the fixed-length binning scheme, would be very beneficial if $k$ is a considerable fraction of $D$. This is possible in EBR (embedding-based retrieval) applications where $D = 256\sim 1024$ is typical. For example, when $D=256$ and $k=64$, we have $\frac{D-k}{D} =0.75$. A variance reduction by $25\%$ would be quite considerable especially as the fixed-length binning scheme is actually easier to implement than the variable-length binning scheme. The ``only disadvantage'' of the fixed-length scheme is that we cannot choose a $k$ value between $D/2$ and $D$. 

\vspace{0.1in}

Here, we provide a simulation study to verify Theorem~\ref{thm:a} and present the simulation results in Figure~\ref{fig:a}. For each panel (for a specific target $\rho$) of Figure~\ref{fig:a}, we first generate two vectors from the standard bivariate  Gaussian distribution with the target correlation $\rho$.  To avoid ambiguity, we generate the vectors many times until we have two vectors whose cosine value is very close to the target $\rho$ before we store the vectors. Otherwise the empirical cosine value can be quite different from the target $\rho$. After we generate the two vectors, we normalize them to simplify the presentation of the results because otherwise the results would be related to the norms too. Then we conduct OPORP $10^5$ times for each $k$  in $\{2, 4, 8, 16, 32, ..., D/2\}$. For convenience, we choose $D$ to be powers of 2. We only present results for $D=1024$ and $D=64$ because the other plots are pretty similar. Note that for the variable-length binning scheme, we also add simulations for $D/2<k<D$.

\vspace{0.1in}
We report the simulations for both $s=1$ and $s=3$. In each panel, we plot four curves: the empirical mean square errors (MSE = variance + bias$^2$) for both binning schemes, and the theoretical variance curves (in dashed lines) for both binning schemes.  The dashed lines are not visible because they overlap with the empirical MSEs, which verify that the correctness of the variance formulas. We can also see that, with the fixed-length binning scheme (Bin\#1), the variance is noticeably smaller than the variance of the variable-length scheme at the same $k$, confirming the benefits due to the $\frac{D-k}{D-1}$ term. Note that for $s=3$, the difference  between the two binning scheme becomes smaller, because in the formulas the $\frac{D-k}{D-1}$ term does not apply to the term involving $(s-1)$.

\subsection{The Normalized Estimators}

One can (substantially) improve the estimation accuracy via the ``normalization'' trick. That is, once we have the samples ($x_j$, $y_j$), $j=1, 2, ..., k$, we can use the following normalized estimator:
\begin{align}\notag
\hat{\rho} = \frac{\sum_{j=1}^k x_j y_j}{\sqrt{\sum_{j=1}^k x_j^2}{\sqrt{\sum_{j=1}^k y_j^2}} }.
\end{align}
Again, we use $\hat{\rho}_1$ and $\hat{\rho}_2$ to denote the estimates for the fixed-length binning and variable-length binning, respectively. As explained earlier, the normalization step will not be needed at the estimation time if we pre-normalize and store the data, e.g., $x_j^\prime = \frac{x_j^\prime}{\sqrt{\sum_{t=1}^kx_t^2}}$.

\begin{theorem}\label{thm:rho}
For large $k$, $\hat{\rho}$ converges to $\rho$, almost surely, with 
\begin{align*}
Var(\hat{\rho}_1)
=&(s-1)A
+\left\{\frac{1}{k}\left[ (1-\rho^2)^2 -2A\right]+O\left(\frac{1}{k^2}\right)\right\}\frac{D-k}{D-1}, \\
Var(\hat{\rho}_2)
=&(s-1)A
+\left\{\frac{1}{k}\left[ (1-\rho^2)^2 -2A\right]+O\left(\frac{1}{k^2}\right)\right\}.
\end{align*}
where 
\begin{align*}
A = \sum_{i=1}^D\left(u_i^\prime v_i^\prime-\rho/2({u_i^\prime}^2+{v_i^\prime}^2)\right)^2, \hspace{0.2in} u_i^\prime = \frac{u_i}{\sqrt{\sum_{t=1}^D u_t^2}}, \hspace{0.2in} 
v_i^\prime = \frac{v_i}{\sqrt{\sum_{t=1}^D v_t^2}}.
\end{align*}
\end{theorem}
\noindent\textbf{Proof of Theorem~\ref{thm:rho}:}  See Appendix~\ref{proof:thm:rho}. $\hfill\qed$

\vspace{0.2in}

The variance expressions in Theorem~\ref{thm:rho}  hold for large $k$ (i.e., $k\rightarrow D$ for the fixed-length binning and $k\rightarrow\infty$ for the variable-length binning).  Note that the term $\frac{1}{k}\left(1-\rho^2\right)^2$ inside the variances of $\hat{\rho}$ is exactly the classical asymptotic variance of the correlation estimator for the bivariate Gaussian distribution~\citep{anderson2003introduction}. Because $A\geq0$, we know that OPORP achieves smaller (asymptotic) variance than the classical estimator in statistics, even without considering the $\frac{D-k}{D-1}$ factor.

\begin{figure}[h]

    \centering
\mbox{
   \includegraphics[width=2.7in]{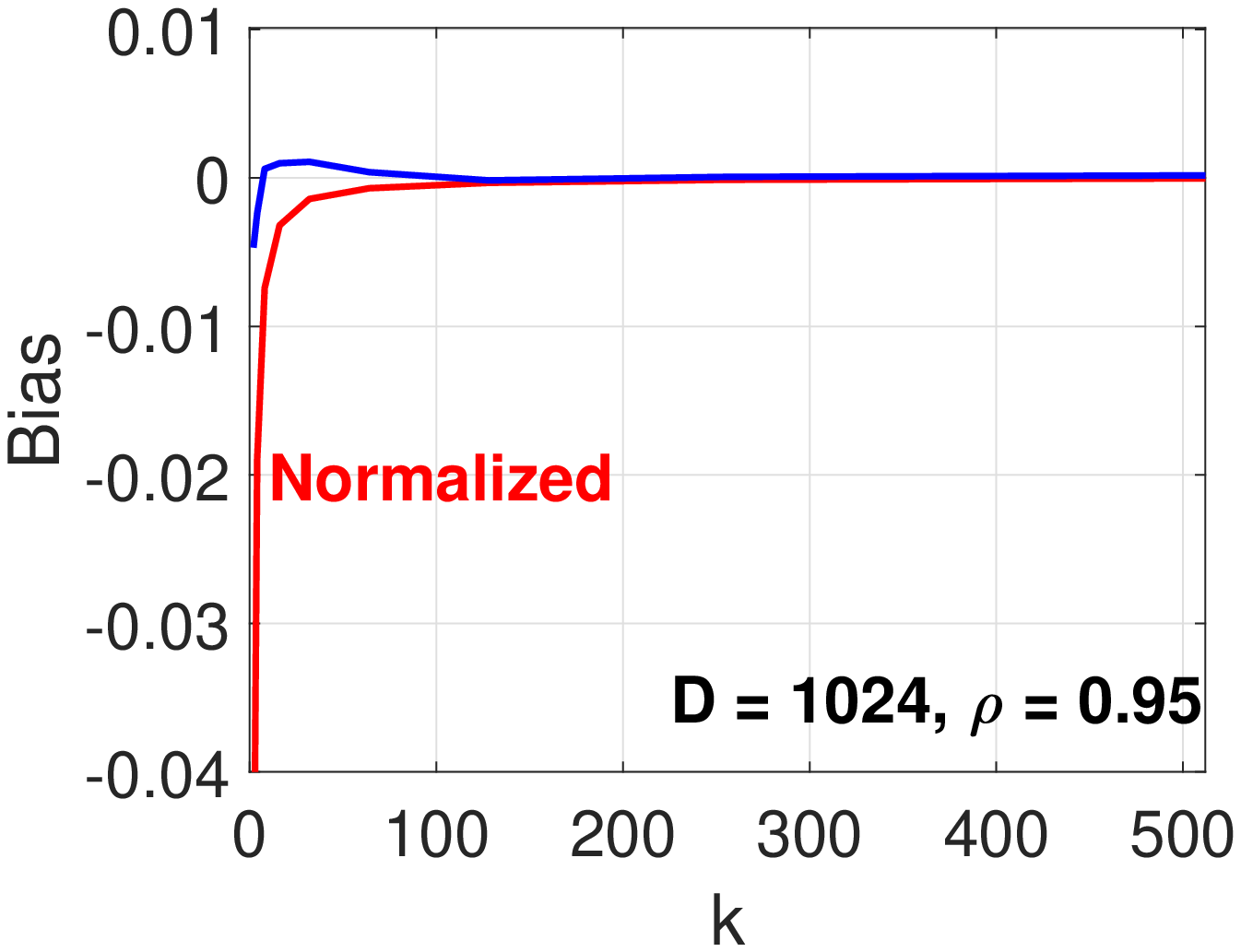}
   \includegraphics[width=2.7in]{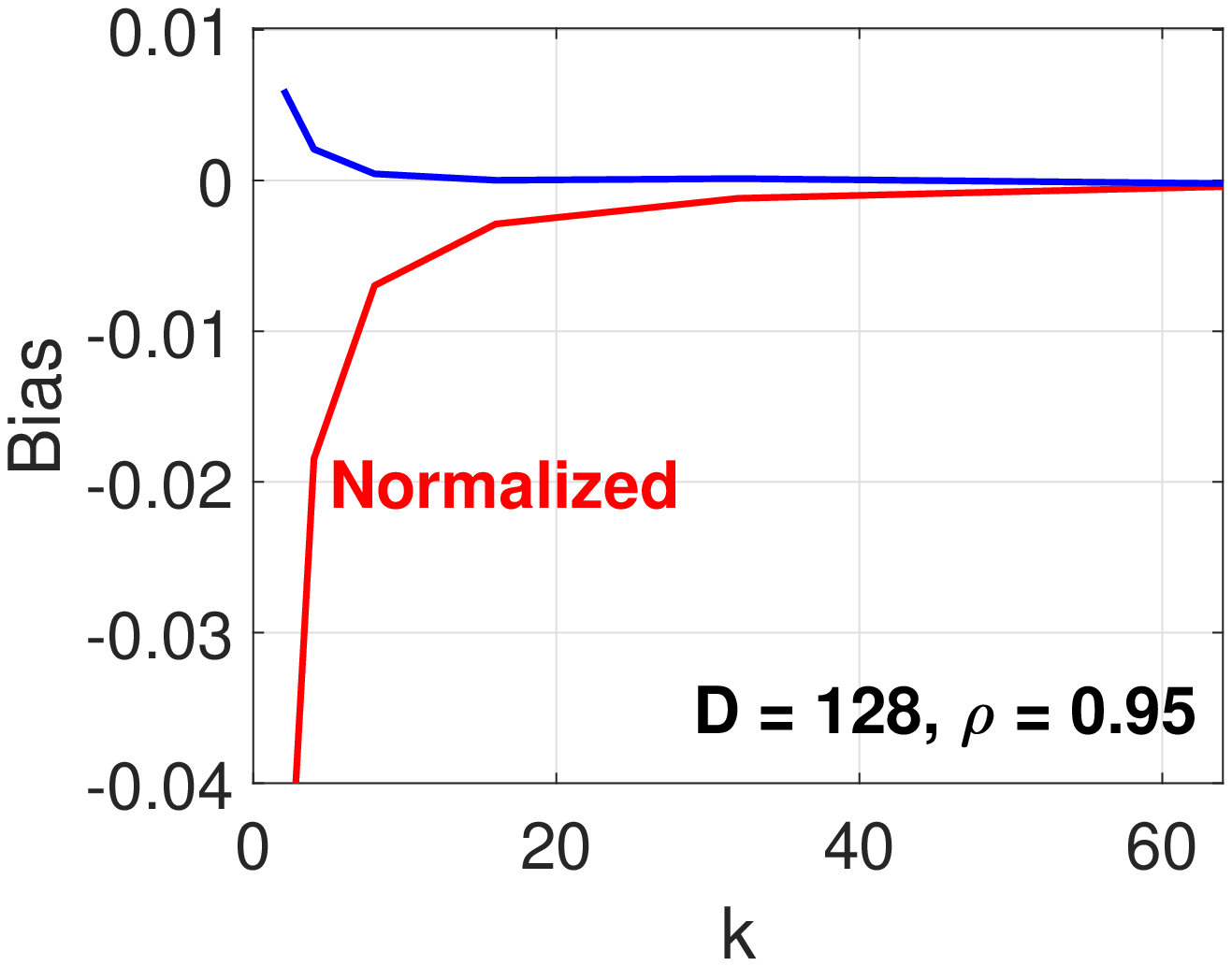}
}

\vspace{-0.15in}

   \caption{Empirical biases ($E\left(\hat{\rho}\right)-\rho$) of the normalized estimator $\hat{\rho}$ as well as the un-normalized estimator $\hat{a}$, evaluated on the same normalized data vectors in Figure~\ref{fig:a}, for $s=1$ and the fixed-length binning scheme. The empirical biases are very small (and bias$^2$ would be much smaller).}
    \label{fig:rho_bias}
\end{figure}

\begin{figure}[h!]
    \centering
\mbox{
   \includegraphics[width=2.7in]{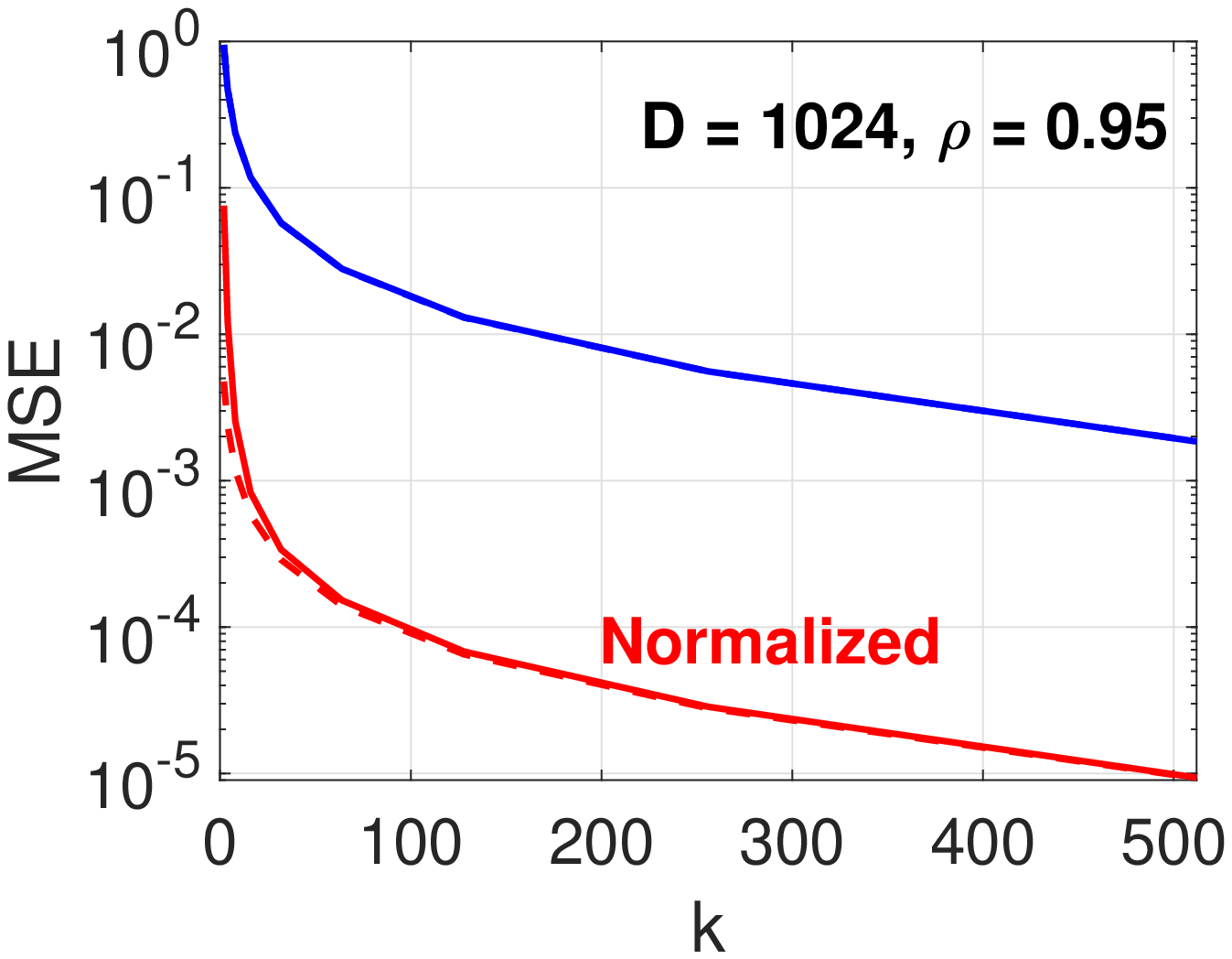}
   \includegraphics[width=2.7in]{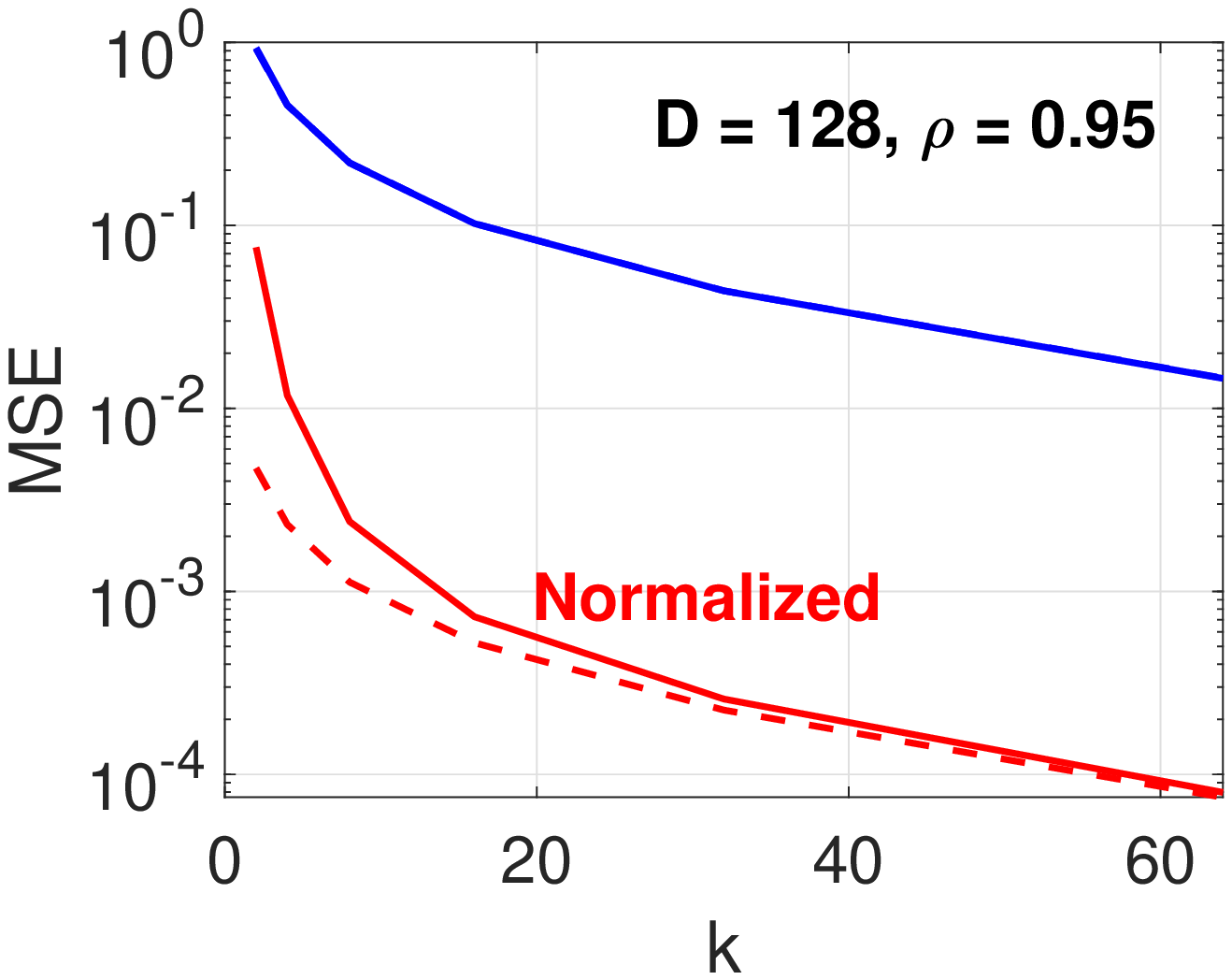}
}

\mbox{
   \includegraphics[width=2.7in]{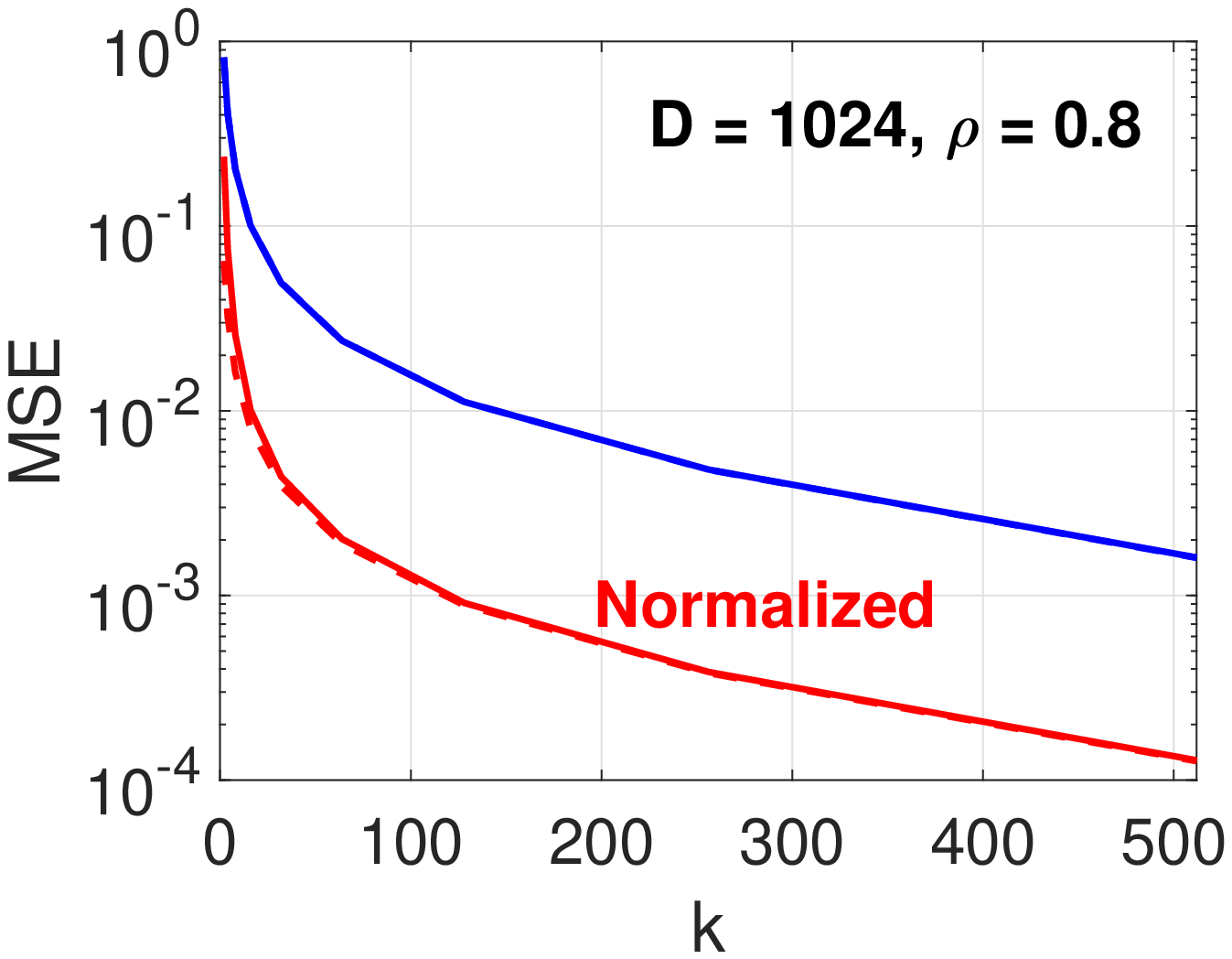}
   \includegraphics[width=2.7in]{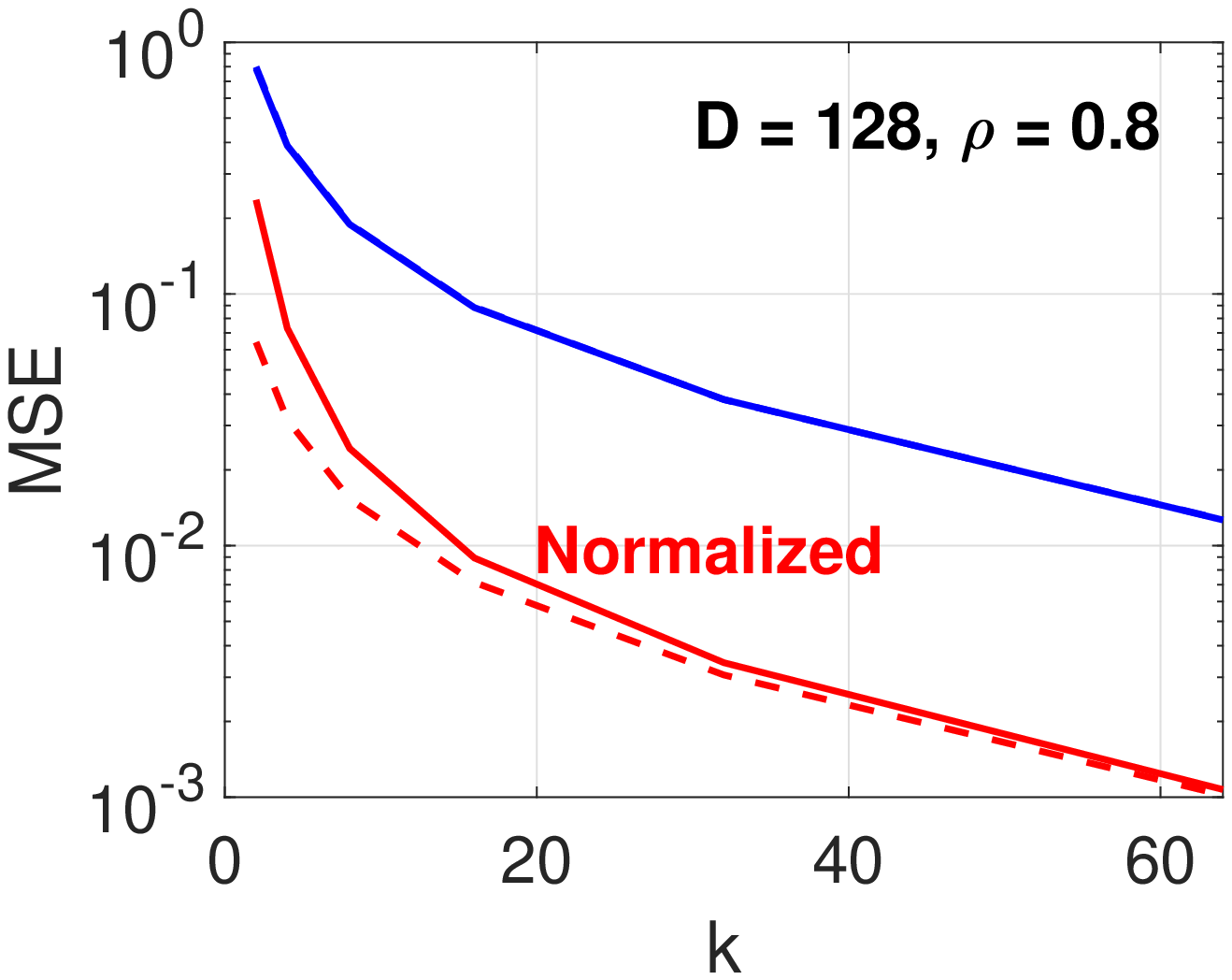}
}

\mbox{
   \includegraphics[width=2.7in]{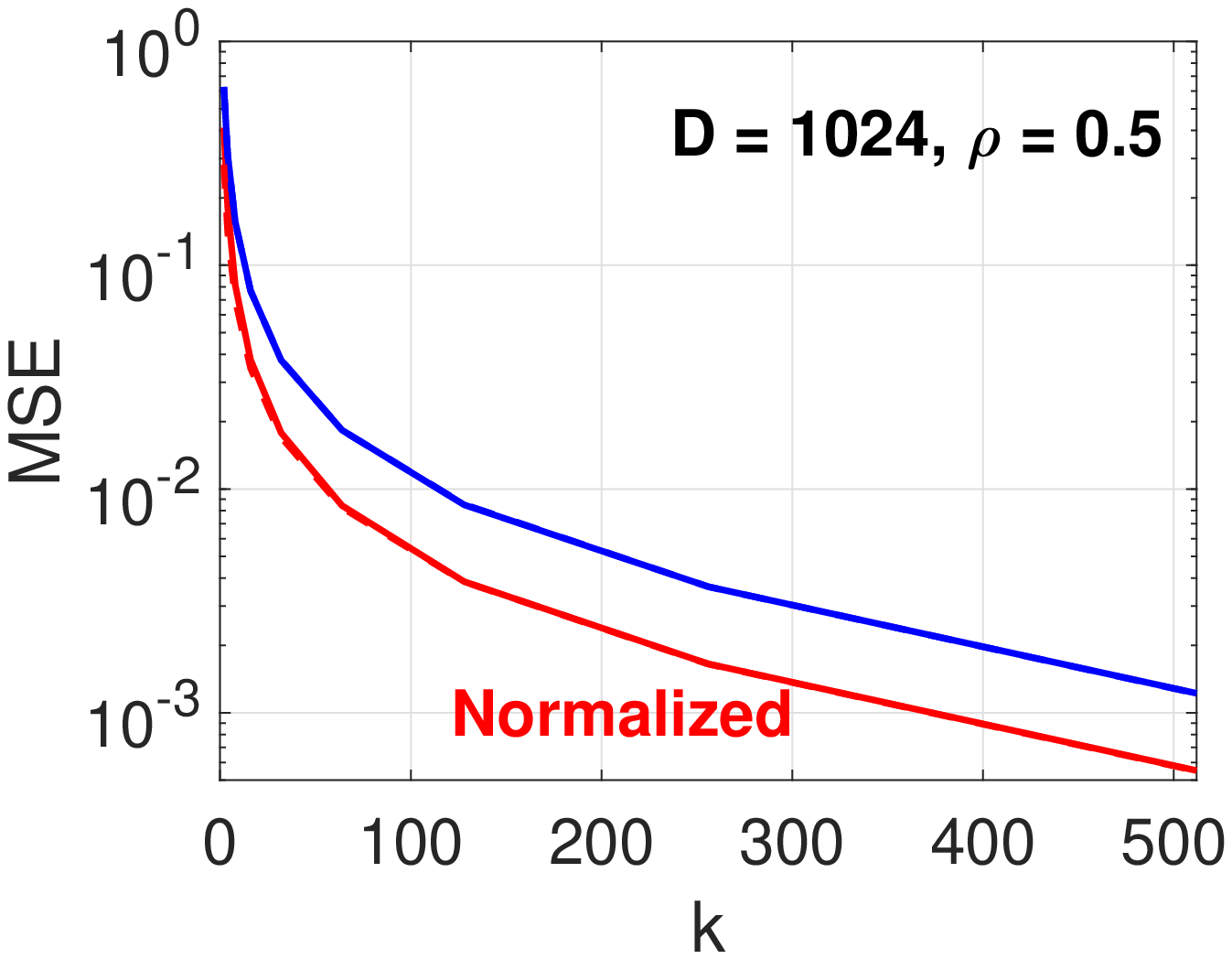}
   \includegraphics[width=2.7in]{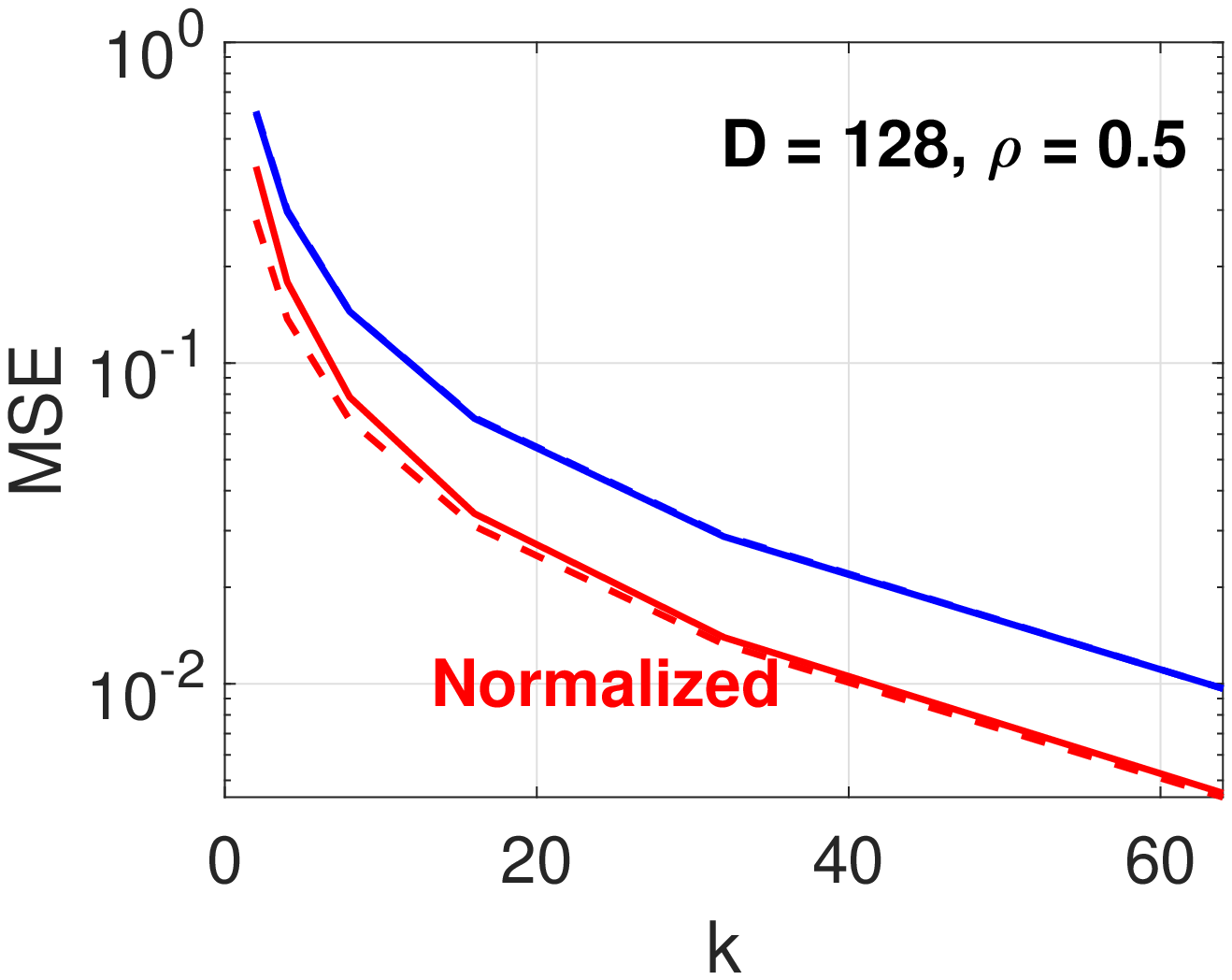}
}

\mbox{
   \includegraphics[width=2.7in]{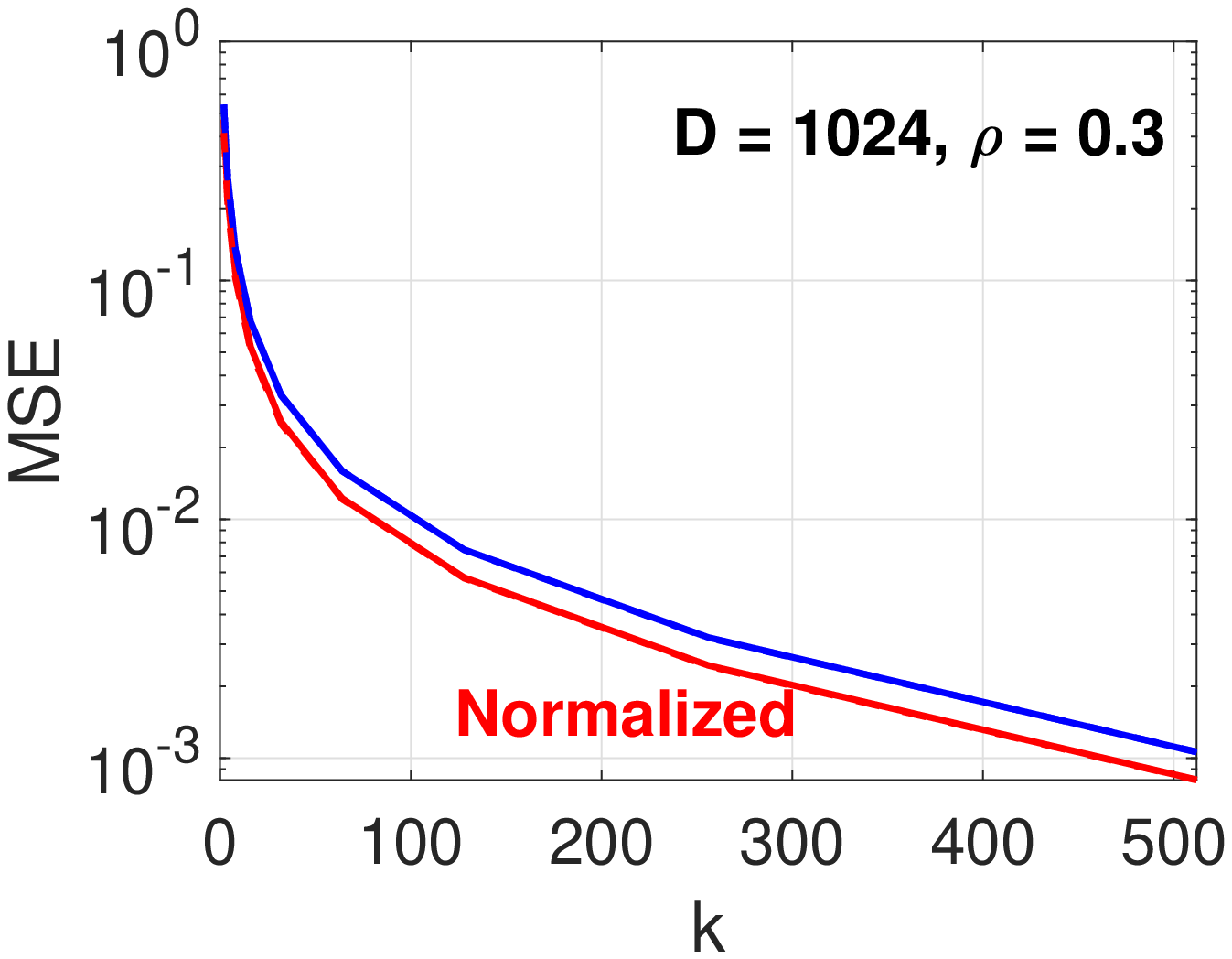}
   \includegraphics[width=2.7in]{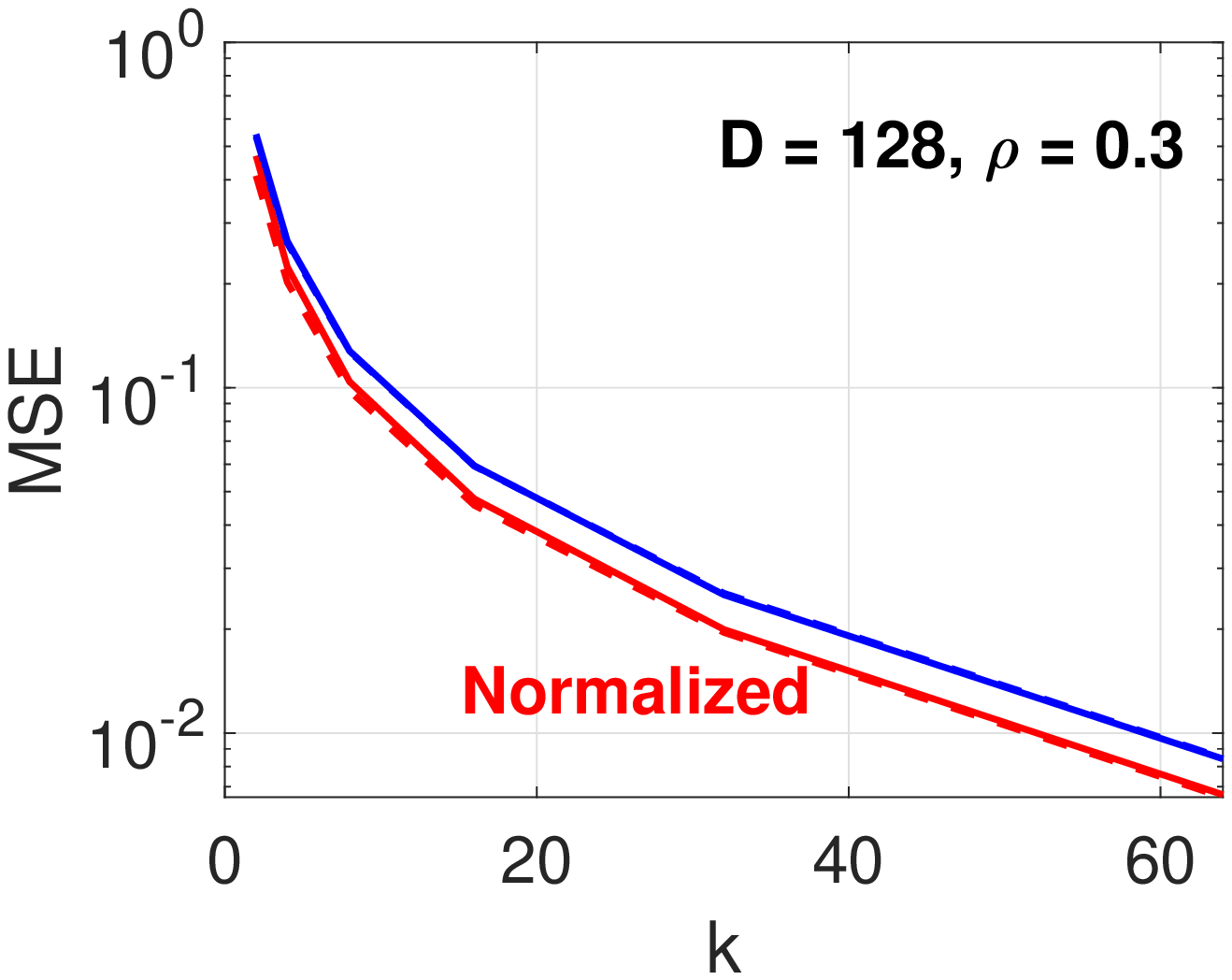}
}

\vspace{-0.2in}

   \caption{Empirical MSEs for both un-normalized and normalized estimators of OPORP, for $s=1$ and the fixed-length binning scheme, using the same normalized data vectors as in Figure~\ref{fig:a}. The normalization step reduces the MSEs considerably especially for large $\rho$ (i.e., more similar pairs). The dashed curves for the theoretical (asymptotic) variance of $\hat{\rho}$ in Theorem~\ref{thm:rho} differ slightly from the empirical MSEs (solid curves) if $k$ is small.}
    \label{fig:rho}
\end{figure}

A simulation study presented in Figure~\ref{fig:rho_bias} and Figure~\ref{fig:rho} shows that $k$ does not need to be  large in order for these variance formulas to be sufficiently accurate. 

In Figure~\ref{fig:rho_bias} and Figure~\ref{fig:rho}, we use the same data vectors as  in Figure~\ref{fig:a}, for $s=1$ and only the fixed-length binning scheme. Recall that those generated vectors are already normalized, and hence the inner product is the same as the cosine. This makes it convenient to present both the un-normalized and normalized estimators in the same plot. Recall MSE = variance + bias$^2$. Figure~\ref{fig:rho_bias} illustrates that the biases are very small (and bias$^2$ would be much smaller), as long as $k$ is not too small. The empirical MSE plots in Figure~\ref{fig:rho} confirms the significant variance reduction of the normalization step. The variance formula  in Theorem~\ref{thm:rho} is accurate, as long as $k$ is not too small.

\newpage\clearpage

\subsection{The Normalized Estimator for VSRP}

We have already explained how to recover ``very sparse random projections'' (VSRP)~\citep{li2006very} from OPORP by using $k=1$ and repeating OPORP $m$ times. We can therefore also take advantage of this finding to develop the normalized estimator for VSRP and obtain its variance.  To present the estimator and its theory for VSRP, instead of introducing new notation, we borrow the existing notation. Also, we  still use $k$ for the sample size of VSRP instead of $m$. That is, we have 
\begin{align}\notag
    x_j = \sum_{i=1}^{D} u_i r_{ij},\hspace{0.3in}
    y_j = \sum_{i=1}^{D} v_i r_{ij}, \hspace{0.2in} j = 1, 2,  ..., k.
\end{align}
where $r_{ij}$ follows the following sparse distribution parameterized by $s$: 
\begin{align}\notag
r_{ij} = \sqrt{s}\times \left\{\begin{array}{rrl}
-1 & \text{with prob.} &1/(2s)\\
0 & \text{with prob.} &1-1/s\\
+1 & \text{with prob.} &1/(2s)
\end{array}\right.
\end{align}
We have the un-normalized estimator for $a$ and the normalized for $\rho$:
\begin{align}\notag
\hat{a}_{vsrp} = \frac{1}{k}\sum_{j=1}^k x_j y_j,\hspace{0.2in}
\hat{\rho}_{vsrp} = \frac{\sum_{j=1}^k x_j y_j}{\sqrt{\sum_{j=1}^k x_j^2}{\sqrt{\sum_{j=1}^k y_j^2}} },
\end{align}
We have shown how to use the variance of $\hat{a}$ to recover the variance of $\hat{a}_{vsrp}$, as 
\begin{align*}
&Var(\hat{a}_{vsrp})
=\frac{1}{k}\left(a^2+ \sum_{i=1}^Du_i^2 \sum_{i=1}^Dv_i^2 +(s-3)\sum_{i=1}^D u_i^2v_i^2\right).
\end{align*}
As the normalized estimator and its variance for VSRP are new, we present the result as a theorem. 

\begin{theorem}\label{thm:vsrp}
As $k\rightarrow \infty$, $\hat{\rho}_{vsrp} \rightarrow \rho$ almost surely, with
\begin{align*}
Var(\hat{\rho}_{vsrp})
=&\frac{1}{k}\left( (1-\rho^2)^2 +(s-3)A\right)+O\left(\frac{1}{k^2}\right), 
\end{align*}
where 
\begin{align*}
A = \sum_{i=1}^D\left(u_i^\prime v_i^\prime-\rho/2({u_i^\prime}^2+{v_i^\prime}^2)\right)^2, \hspace{0.2in} u_i^\prime = \frac{u_i}{\sqrt{\sum_{t=1}^D u_t^2}}, \hspace{0.2in} 
v_i^\prime = \frac{v_i}{\sqrt{\sum_{t=1}^D v_t^2}}.
\end{align*}
\end{theorem}

One way to compare VSRP (for general $s$) with OPORP (for $s=1$ and $m=1$ repetition), is to evaluate the following ratios of variances:
\begin{align}\label{eqn:vsrp_ratio}
&\frac{Var(\hat{a}_{vsrp,s}) }{Var(\hat{a}) } \approx  \frac{\sum_{i=1}^Du_i^2\sum_{i=1}^Dv_i^2 + a^2+(s-3)\sum_{i=1}^Du_i^2v_i^2}
{\sum_{i=1}^Du_i^2\sum_{i=1}^Dv_i^2 + a^2-2\sum_{i=1}^Du_i^2v_i^2},\\\label{eqn:vsrp_ratio_rho}
&\frac{Var(\hat{\rho}_{vsrp,s}) }{Var(\hat{\rho}) } \approx \frac{(1-\rho^2)^2+(s-3)A}{(1-\rho^2)^2-2A},
\end{align}
where we use $\approx$ as we neglect the beneficial factor of $\frac{D-k}{D-1}$ so that the comparison would  favor VSRP.
Obviously, when $s=1$, both ratios equal 1. The ratios increase with increasing $s$ for VSRP. Because the ratio is data-dependent, it is better that we compute it using real data.

\begin{figure}[t]

\centering 

\mbox{
    \includegraphics[width=2.7in]{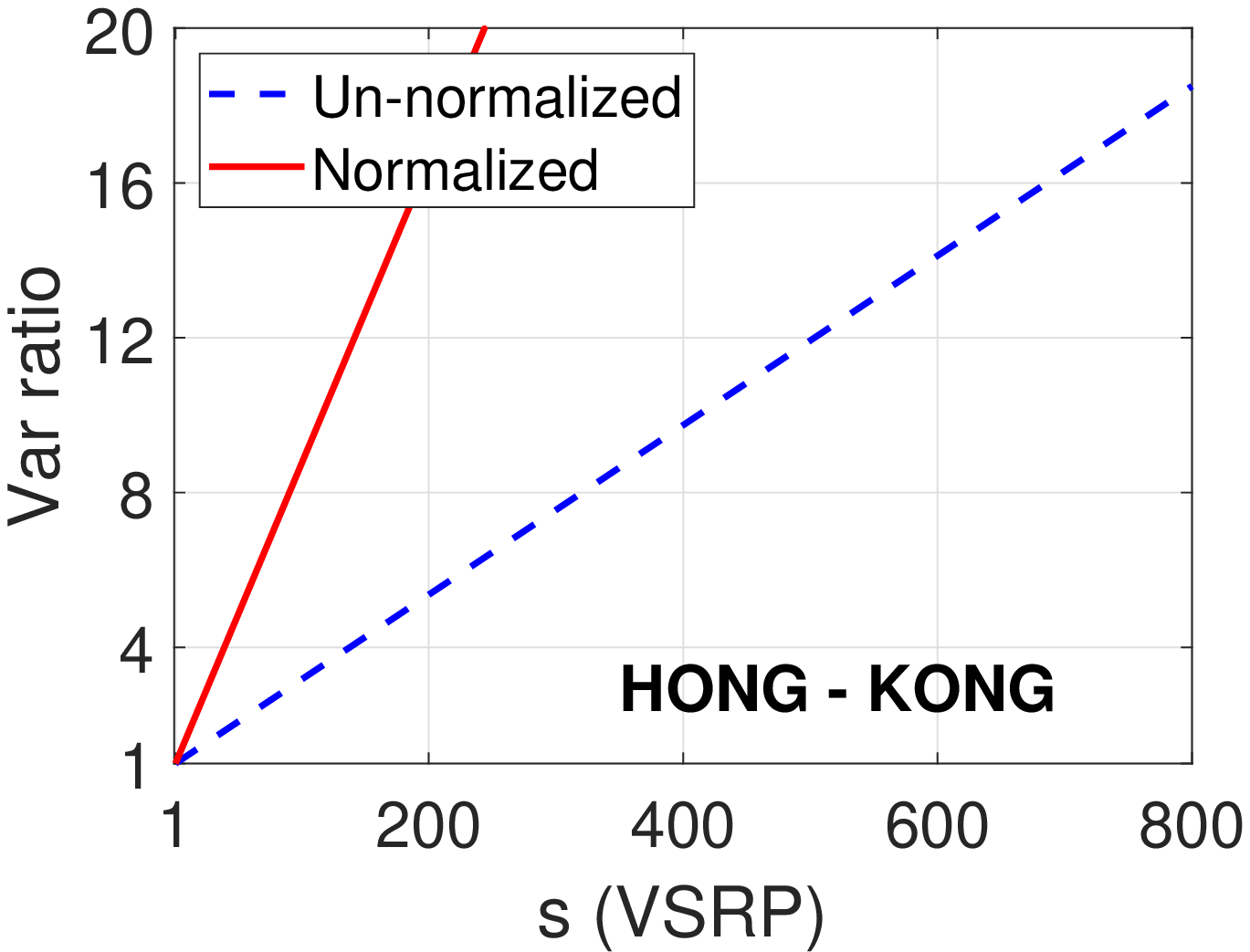}  
    \includegraphics[width=2.7in]{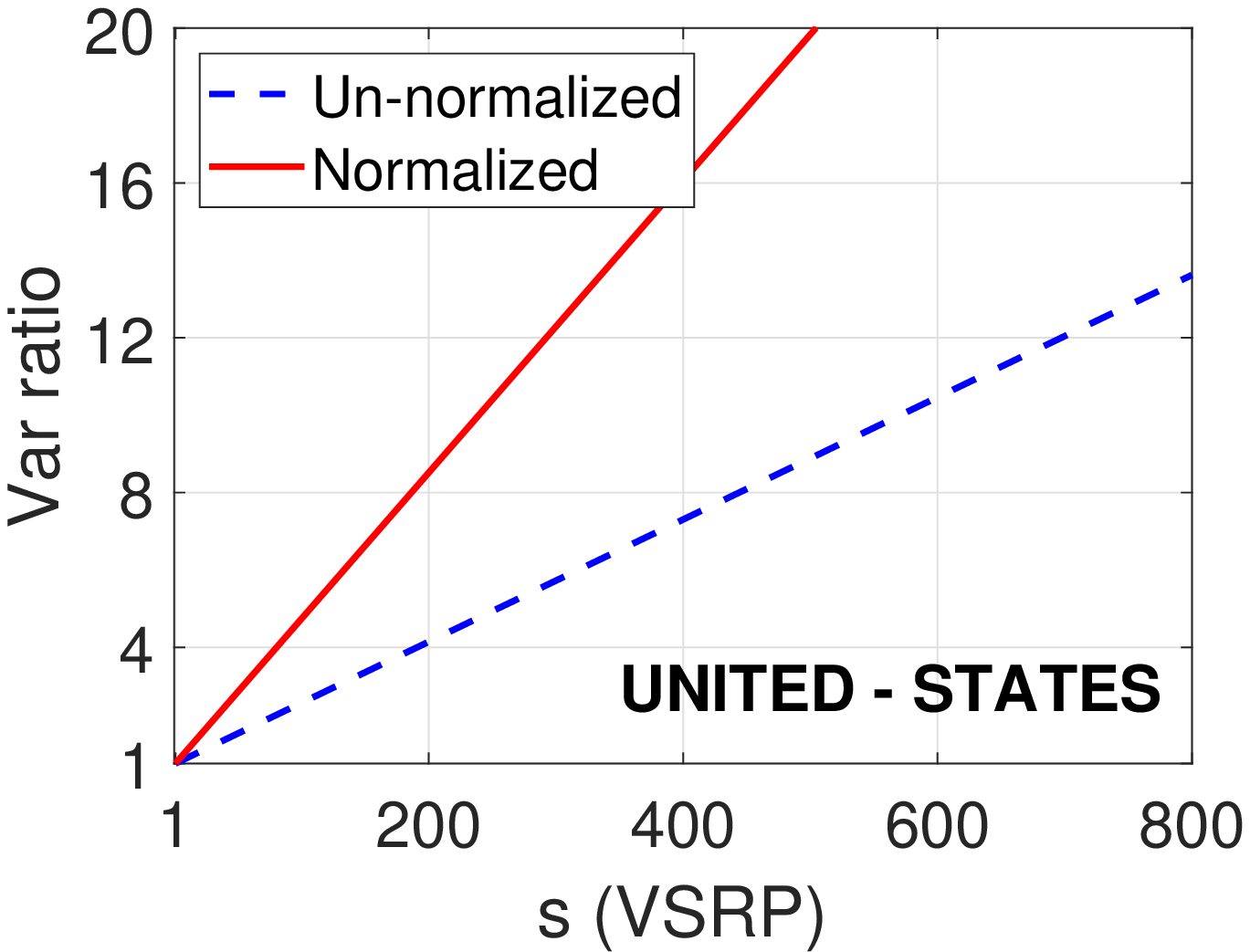}
}

\mbox{
    \includegraphics[width=2.7in]{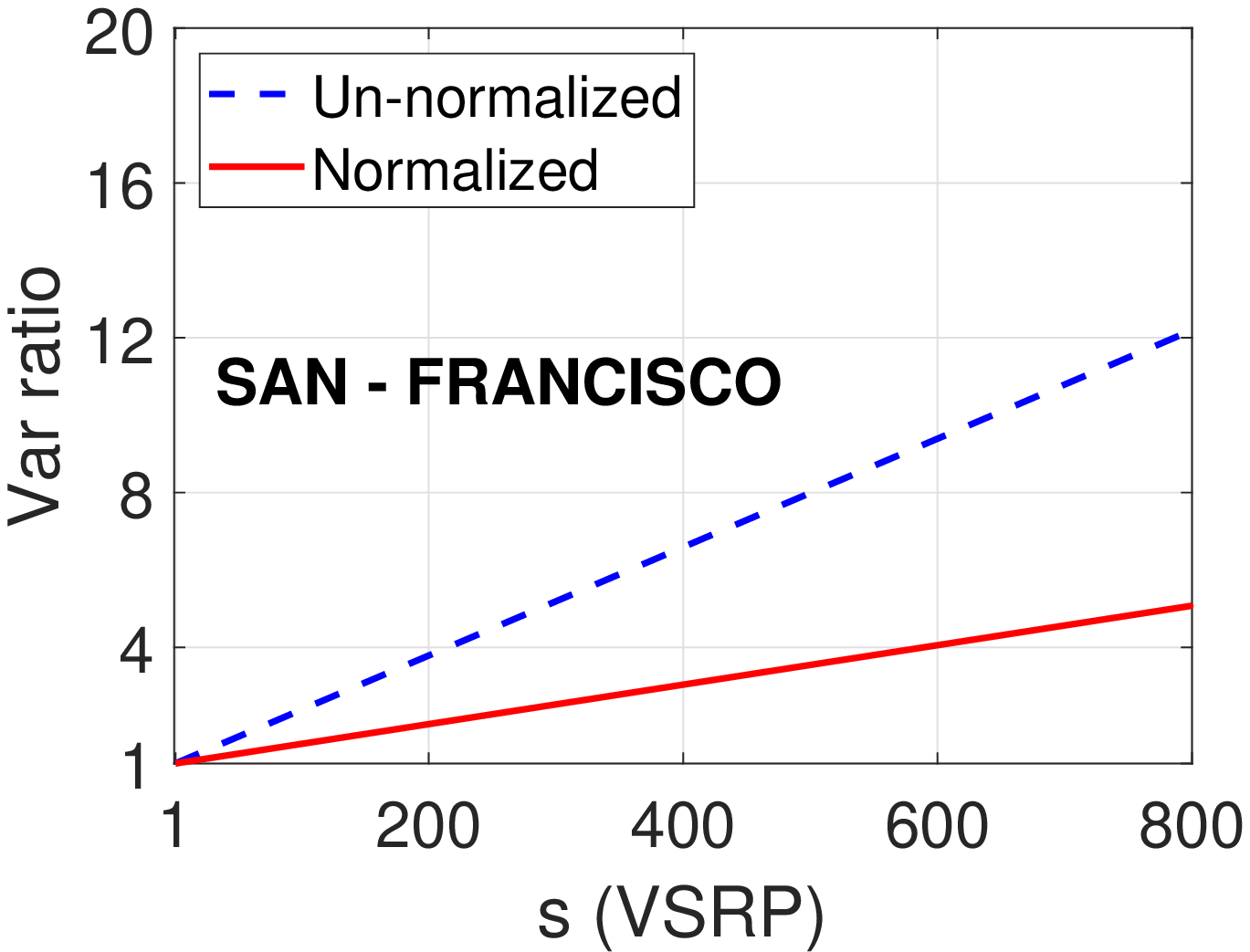}  
    \includegraphics[width=2.7in]{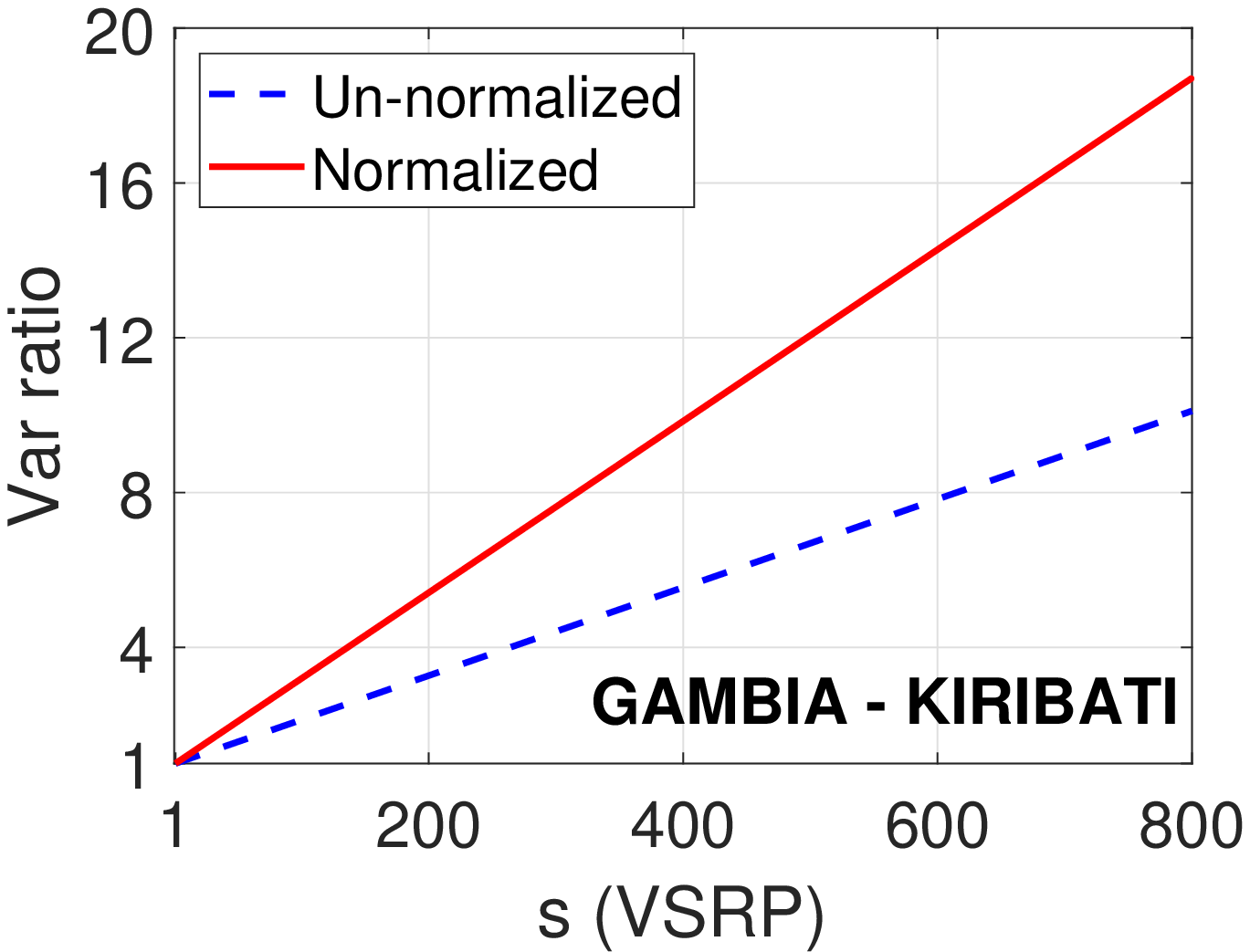}    
}

\vspace{-0.15in}

    \caption{Ratio of variances in~\eqref{eqn:vsrp_ratio} and~\eqref{eqn:vsrp_ratio_rho} to compare VSRP (parameterized by $s$) with OPORP (for its $s=1$), for both the un-normalized (dashed) and normalized (solid) estimators, on four selected word pairs from the ``Words'' dataset (see Table~\ref{tab:words}). }
    \label{fig:vsrp_ratio}
\end{figure}

\begin{table}[h!]
    \centering
\caption{Summary statistics of word-pairs from the ``Words'' dataset~\citep{li2005using}. For example, ``HONG'' represents a vector of length $2^{16}$ with each entry of the vector recording the number of documents that the ``HONG'' appears in the collection of $2^{16}$ documents.\vspace{0.1in}}\label{tab:words}
    \begin{tabular}{cc|ccccc}
    \hline\hline
        word 1 &  word 2 & $\rho$ & $a = \sum_{i=1}^D u_iv_i$ &$\sum_{i=1}^Du_i^2$ & $\sum_{i=1}^Dv_i^2$\\\hline
        HONG &KONG & 0.9623 &12967 & 13556 & 13395\\
        WEEK &MONTH & 0.8954 &281297 & 323073 & 305468\\
        OF & AND & 0.8788 & 57219161 & 69006071 &61437886\\
        UNITED &STATES &0.6693 &69201 &85934 &124415\\
        BEFORE & AFTER & 0.6633 & 59136 & 65541 &121284\\
        SAN    & FRANCISCO & 0.5623 &29386 &125109 &21832\\
        GAMBIA & KIRIBATI & 0.5250 & 228 & 360 &524\\    
        RIGHTS & RESERVED & 0.3949 & 14710 & 79527 &17449\\
        HUMAN & NATURE & 0.2992 & 14896 & 87356 &28367\\
        \hline\hline
    \end{tabular}
\end{table}

Figure~\ref{fig:vsrp_ratio} presents the variance ratios in~\eqref{eqn:vsrp_ratio} and~\eqref{eqn:vsrp_ratio_rho} on four selected word (vector) pairs from the ``Words'' dataset; see Table~\ref{tab:words} for the description of the data. In general, if the $s$ is not too large for VSRP (e.g., $s<10$), then VSRP works pretty well. For larger $s$, then the performance of VSRP largely depends on data. For example, on ``SAN-FRANCISCO'', VSRP with the normalized estimator still works  well (the variance ratio is smaller than 2) if with $s=200$. On ``HONG-KONG'', however, VSRP does not perform well: for the normalized estimator, the variance ratio $>4$ when $s>40$; and for the un-normalized estimator, the variance ratio $>4$ when $s>150$.

The variance ratio = 4 means that we need to increase the sample size of VSRP by a factor of 4 in order to maintain the same accuracy. For VSRP with a the projection matrix size of size $D\times k$, it will need $s=k$ if we hope to achieve the same level of sparsity (on average) as OPORP. Depending on applications, we typically observe that $k=100\sim 500$ might be sufficient for the standard (dense) random projections. Therefore, VSRP using a large $s$ value may lead to poor performance in terms of the required number of projections (which is the also the sample size of VSRP). 
 
In summary, VSRP should work well in general if we use a sparsity parameter $s$ around 10. VSRP may still perform well with a much larger $s$ but then that will be data-dependent. In Section~\ref{sec:experiment}, we will report the retrieval experimental results for VSRP, which also confirm the same finding.

\subsection{The Inner Product Estimators}

The simulations in Figure~\ref{fig:a}, Figure~\ref{fig:rho_bias}, and Figure~\ref{fig:rho} have used data vectors which are normalized to the unit $l_2$ norm, in part for the convenience of presenting the plots. In many EBR applications, the embedding vectors from learning models are indeed already normalized. On the other hand, there are also numerous  applications which use un-normalized data. In fact, the entire literature about ``maximum inner product search'' (MIPS)~\citep{ram2012maximum,shrivastava2014asymmetric,bachrach2014speeding,tan2021norm} is built on the fact that in many applications the norms are different and the goal is to find the maximum inner products (instead of the cosines).  Also see~\citet{fan2019mobius} for the use of MIPS on advertisement retrievals in a commercial search engine. 

\vspace{0.1in}

Recall that, once we have the samples ($x_j$, $y_j$), $j=1, 2, ..., k$, we can estimate the inner product $a$ simply by $\hat{a} = \sum_{j=1}^k x_j y_j$. To improve the estimation accuracy, we can also utilize the normalized cosine estimator $\hat{\rho} = \frac{\sum_{j=1}^k x_j y_j}{\sqrt{\sum_{j=1}^k x_j^2}{\sqrt{\sum_{j=1}^k y_j^2}} }$ 
to have a ``normalized inner product'' estimator $\hat{a}_n$:
\begin{align} \notag
\hat{a}_n = \hat{\rho}\sqrt{\sum_{i=1}^D u_i^2}\sqrt{\sum_{i=1}^D v_i^2},
\end{align}
whose variance would be directly the scaled version of the variance of $\hat{\rho}$:
\begin{align}\notag
Var\left(\hat{a}_n\right)= Var\left(\hat{\rho}\right)\sum_{i=1}^D u_i^2\sum_{i=1}^D v_i^2.
\end{align}

\begin{figure}[b!]

\vspace{-0.1in}

\mbox{\hspace{-0.2in}    
    \includegraphics[width=2.4in]{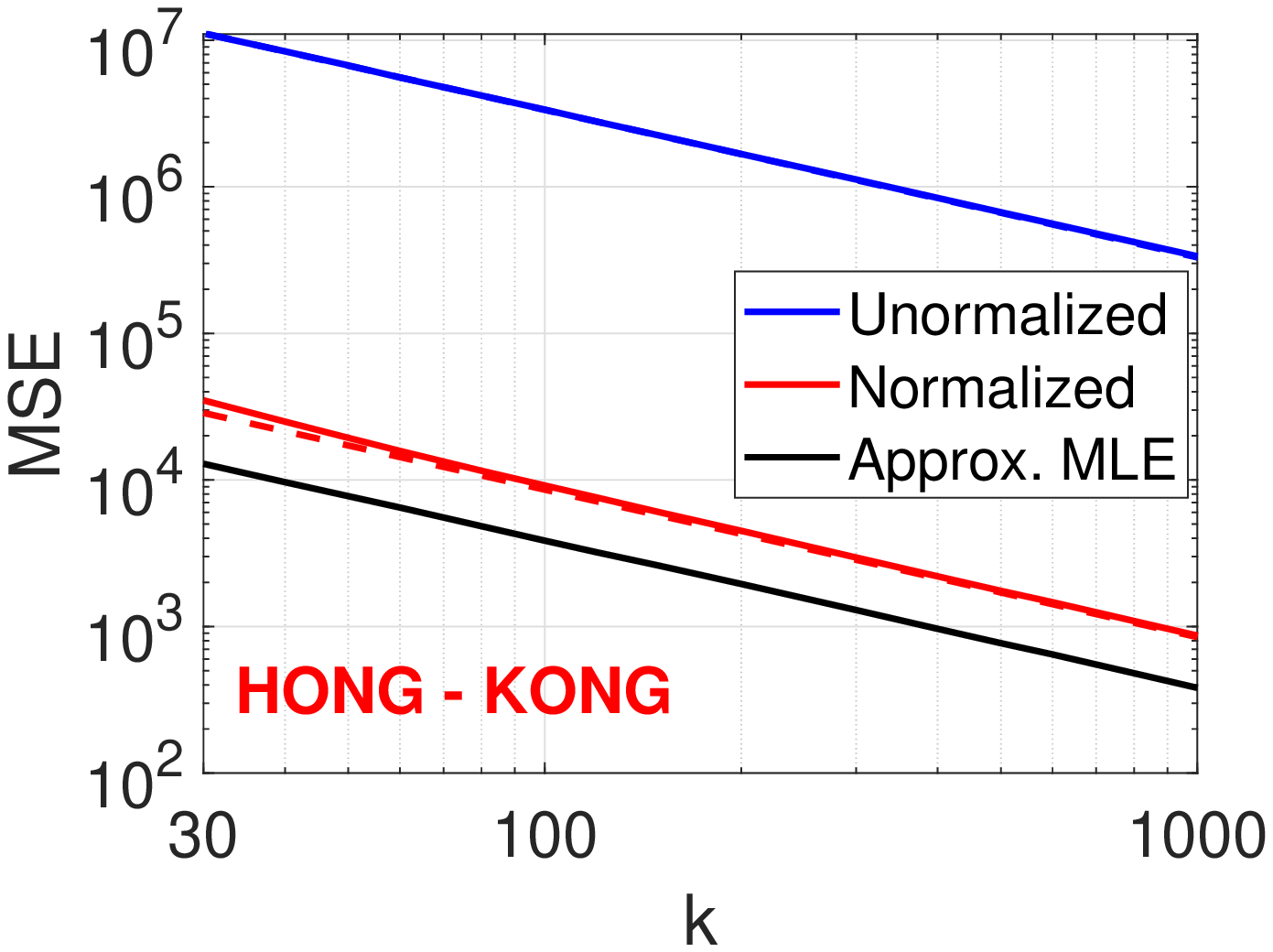}\hspace{-0.15in}   
    \includegraphics[width=2.4in]{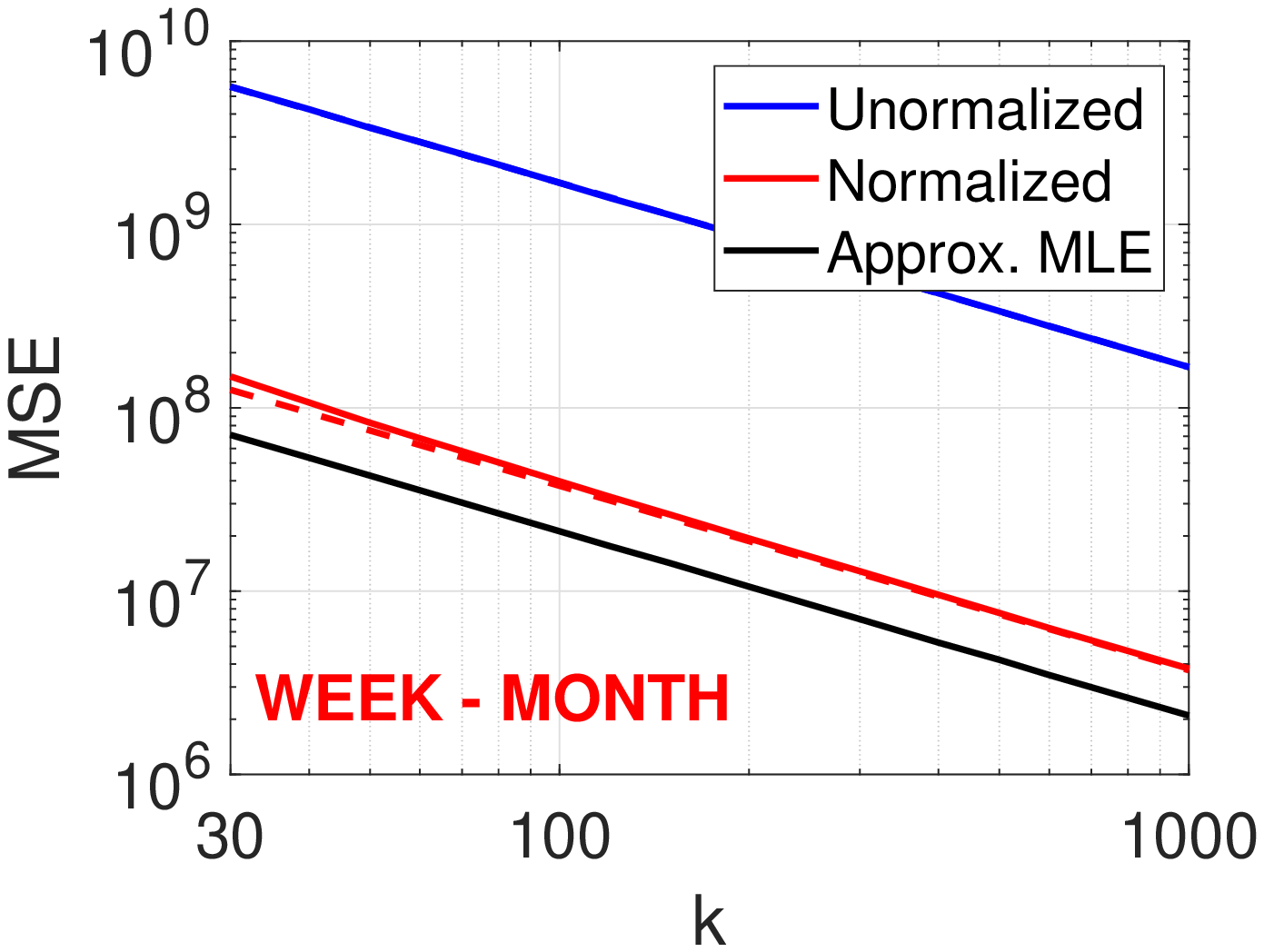}\hspace{-0.15in}   
    \includegraphics[width=2.4in]{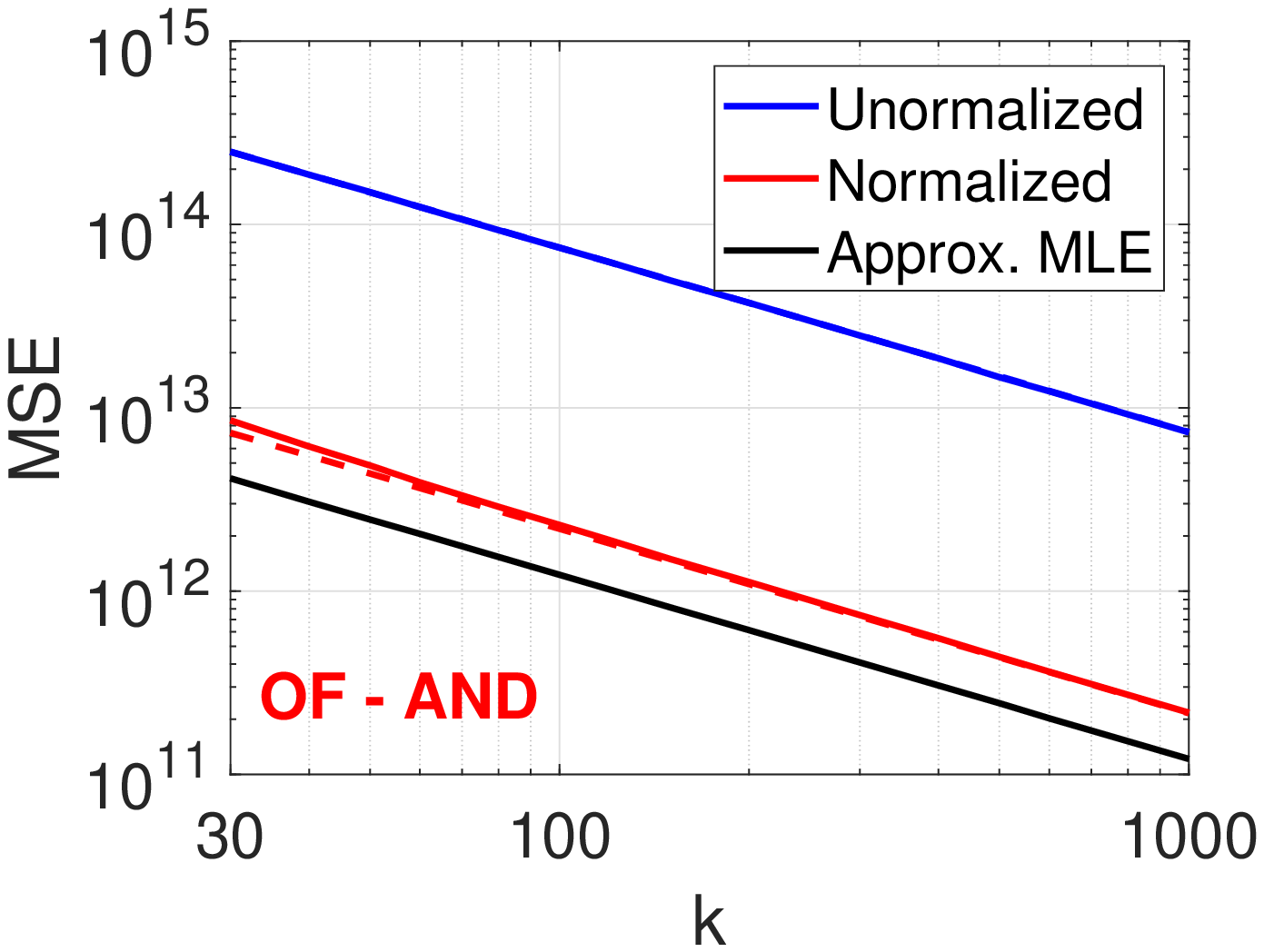}
}
\mbox{\hspace{-0.2in} 
    \includegraphics[width=2.4in]{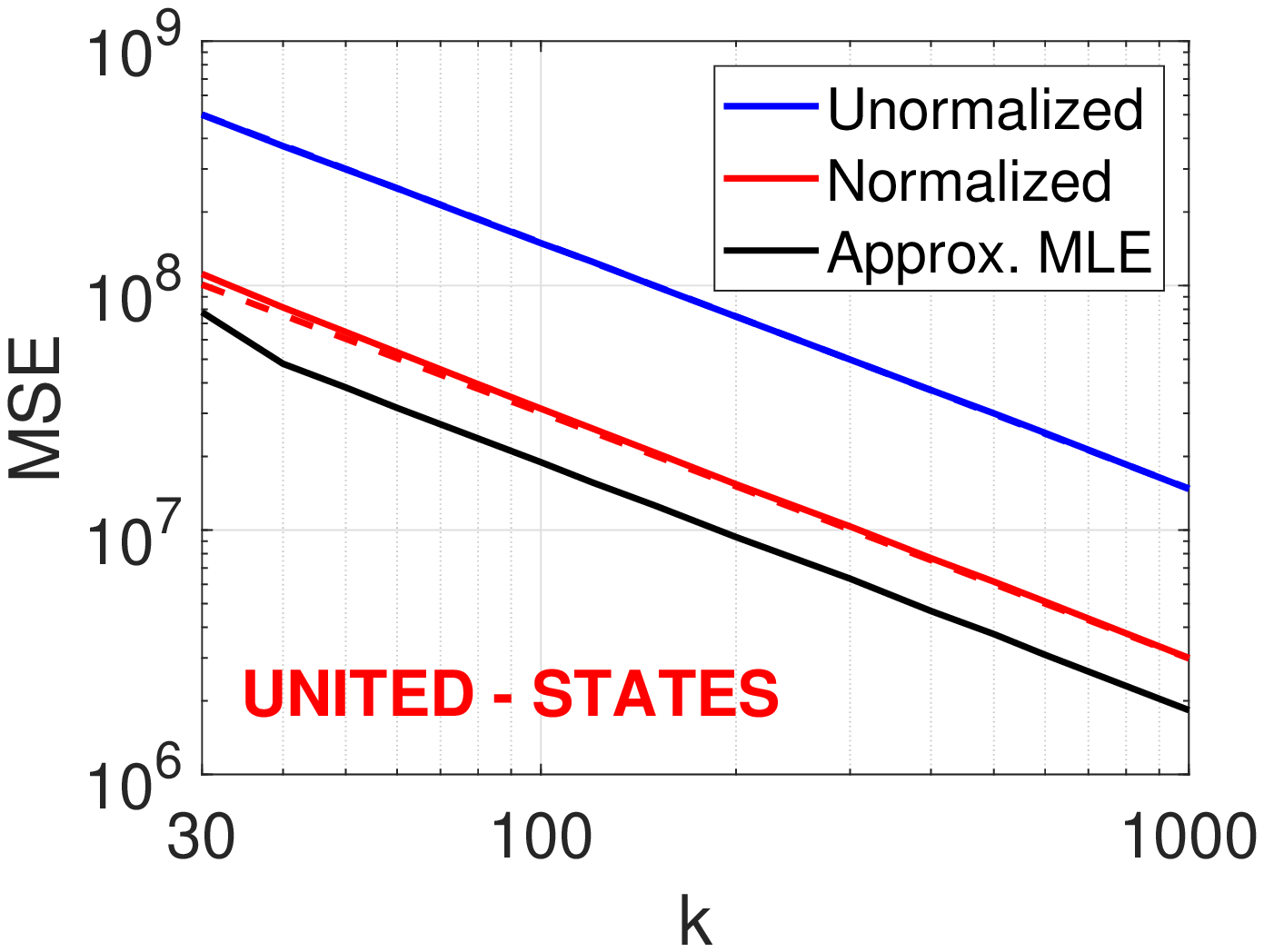}\hspace{-0.15in} 
    \includegraphics[width=2.4in]{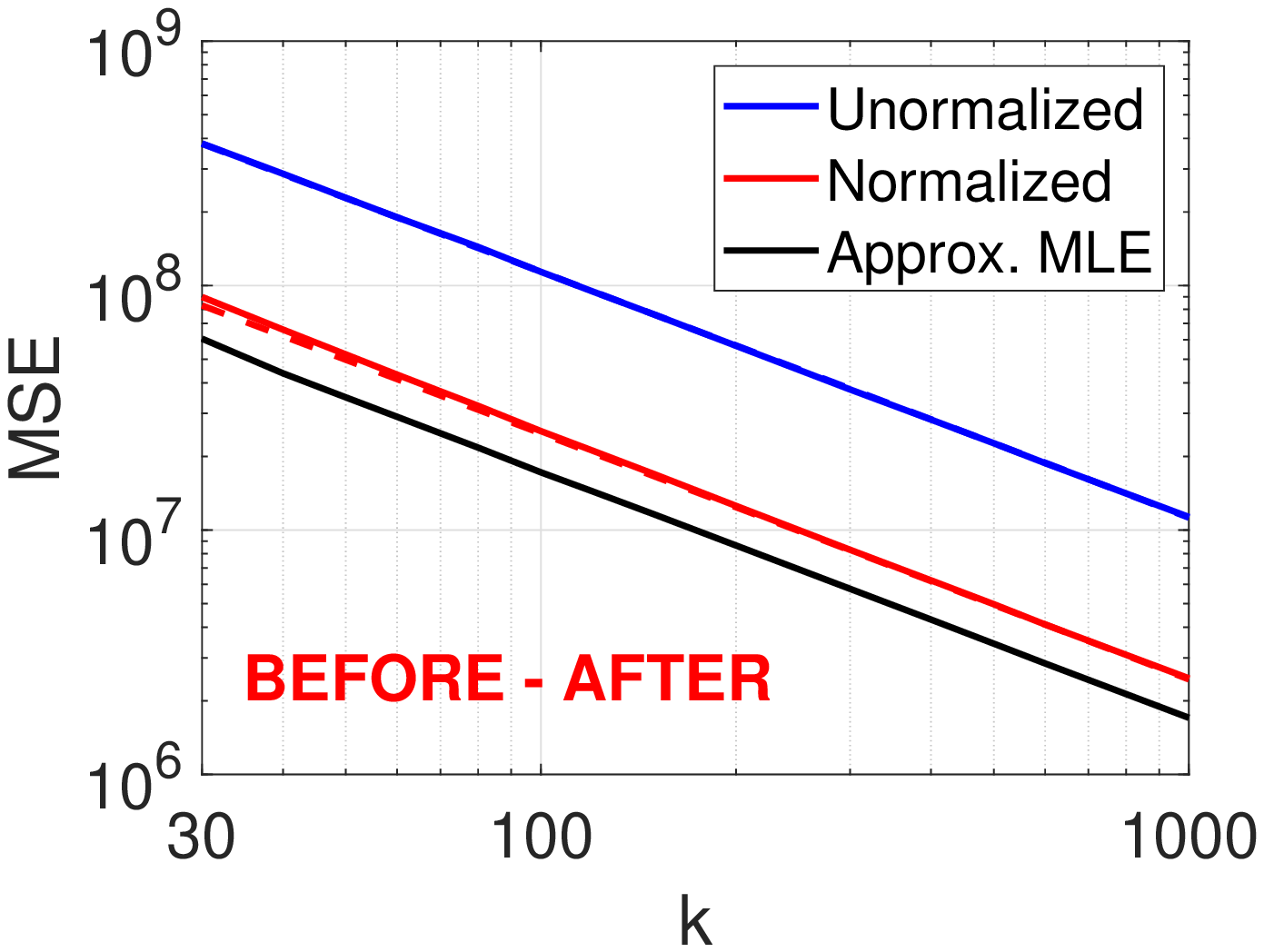}\hspace{-0.15in} 
    \includegraphics[width=2.4in]{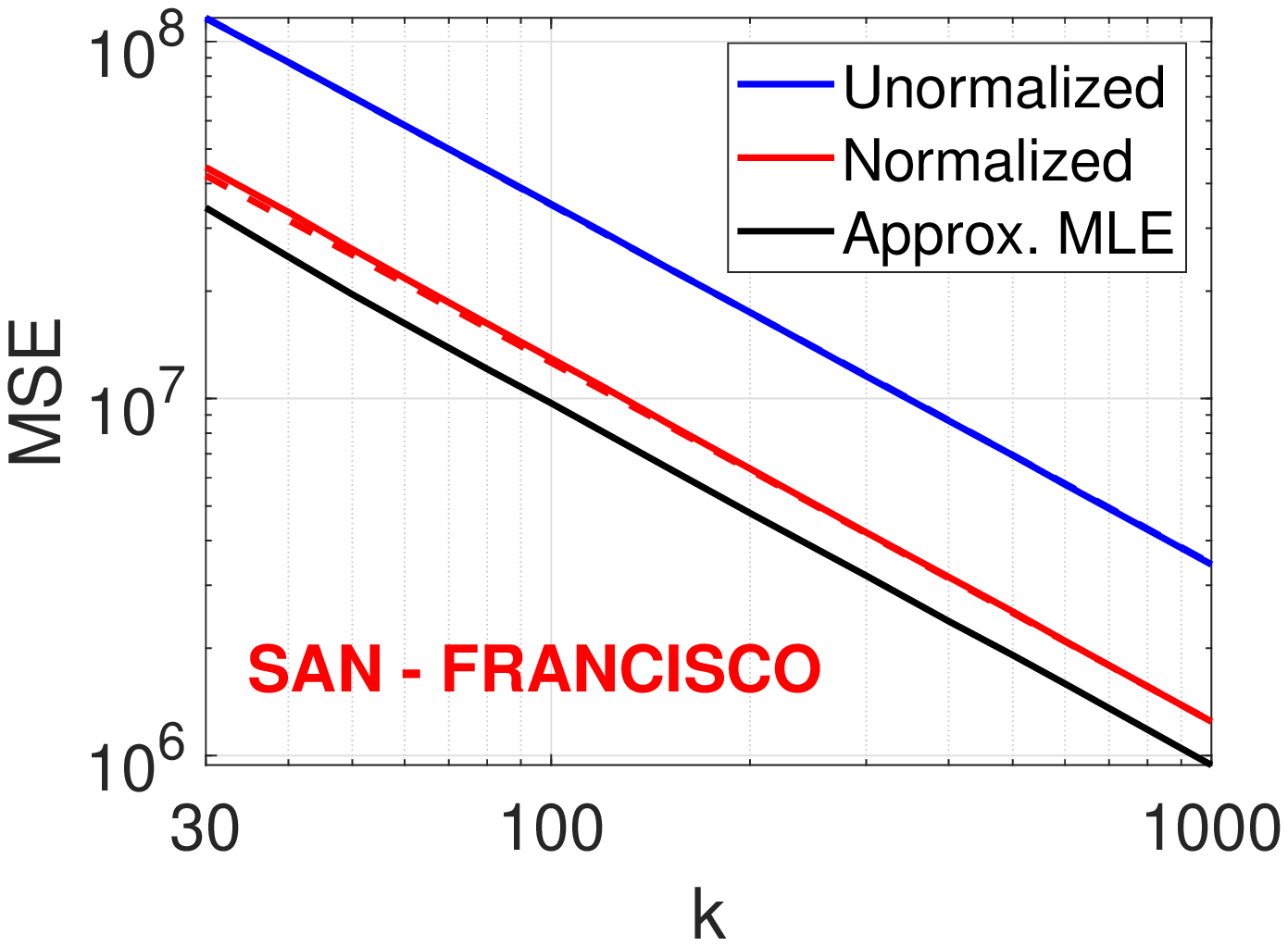}
    }    
\mbox{\hspace{-0.2in} 
    \includegraphics[width=2.4in]{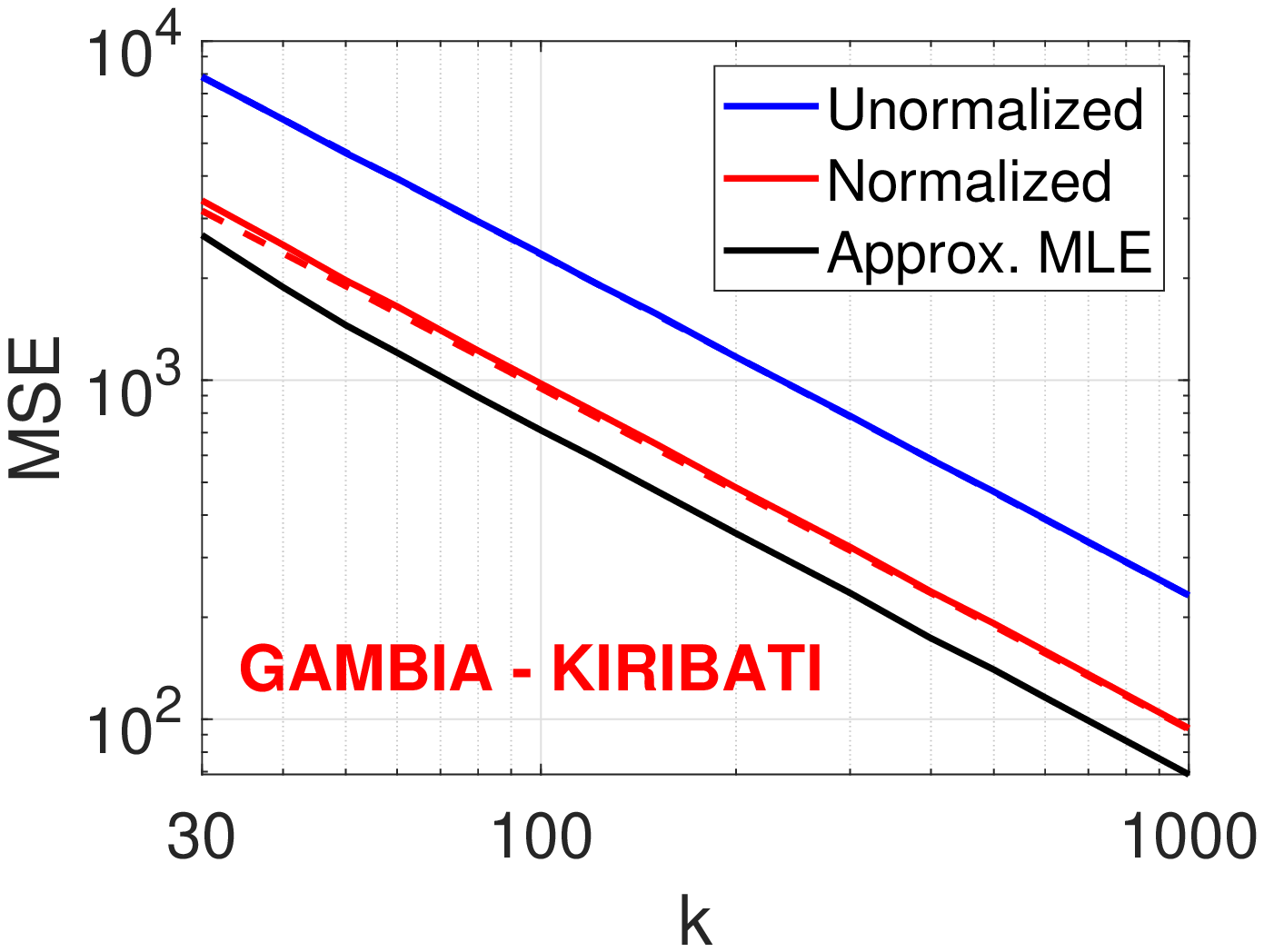}\hspace{-0.15in} 
    \includegraphics[width=2.4in]{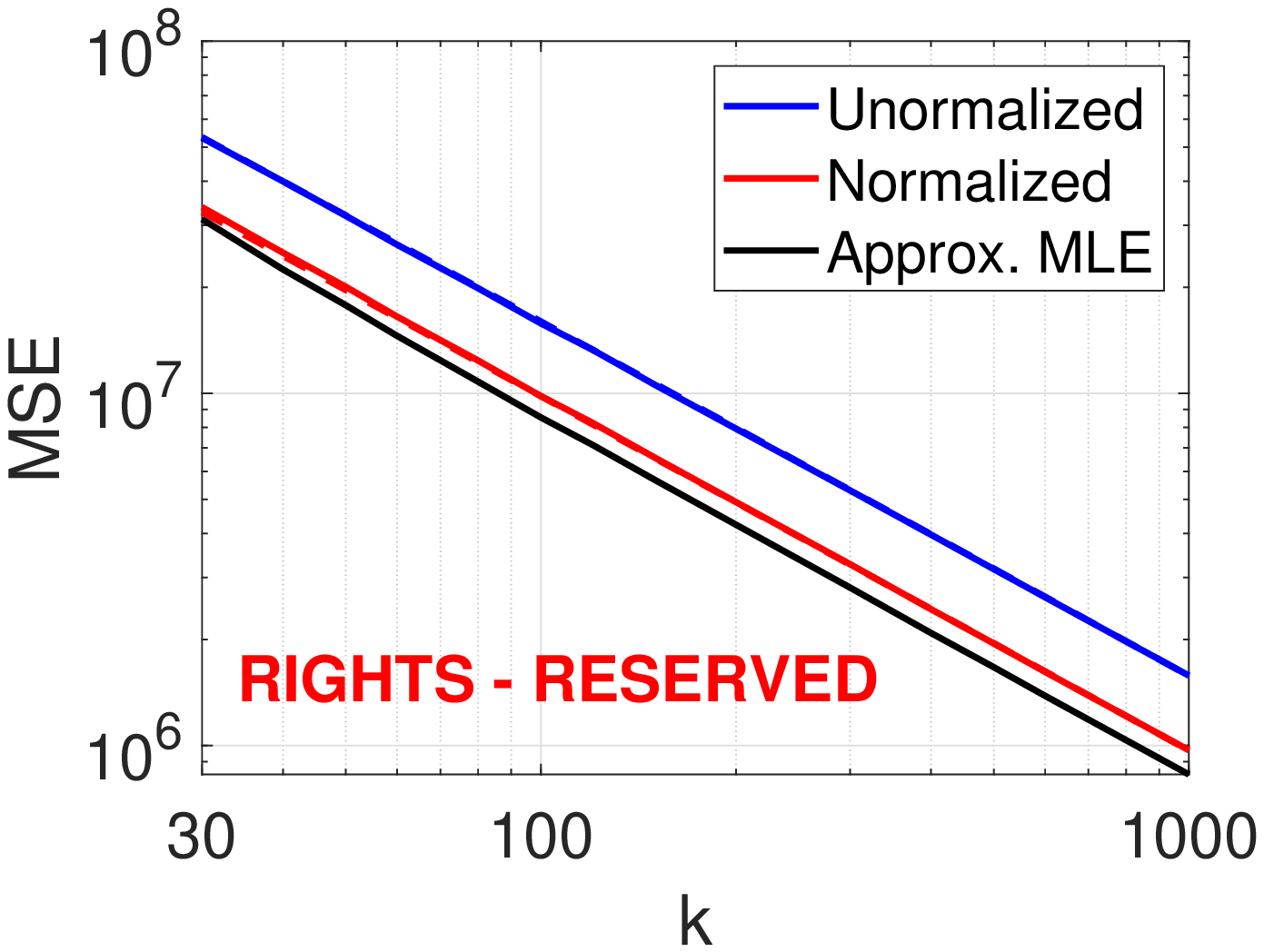}\hspace{-0.15in} 
    \includegraphics[width=2.4in]{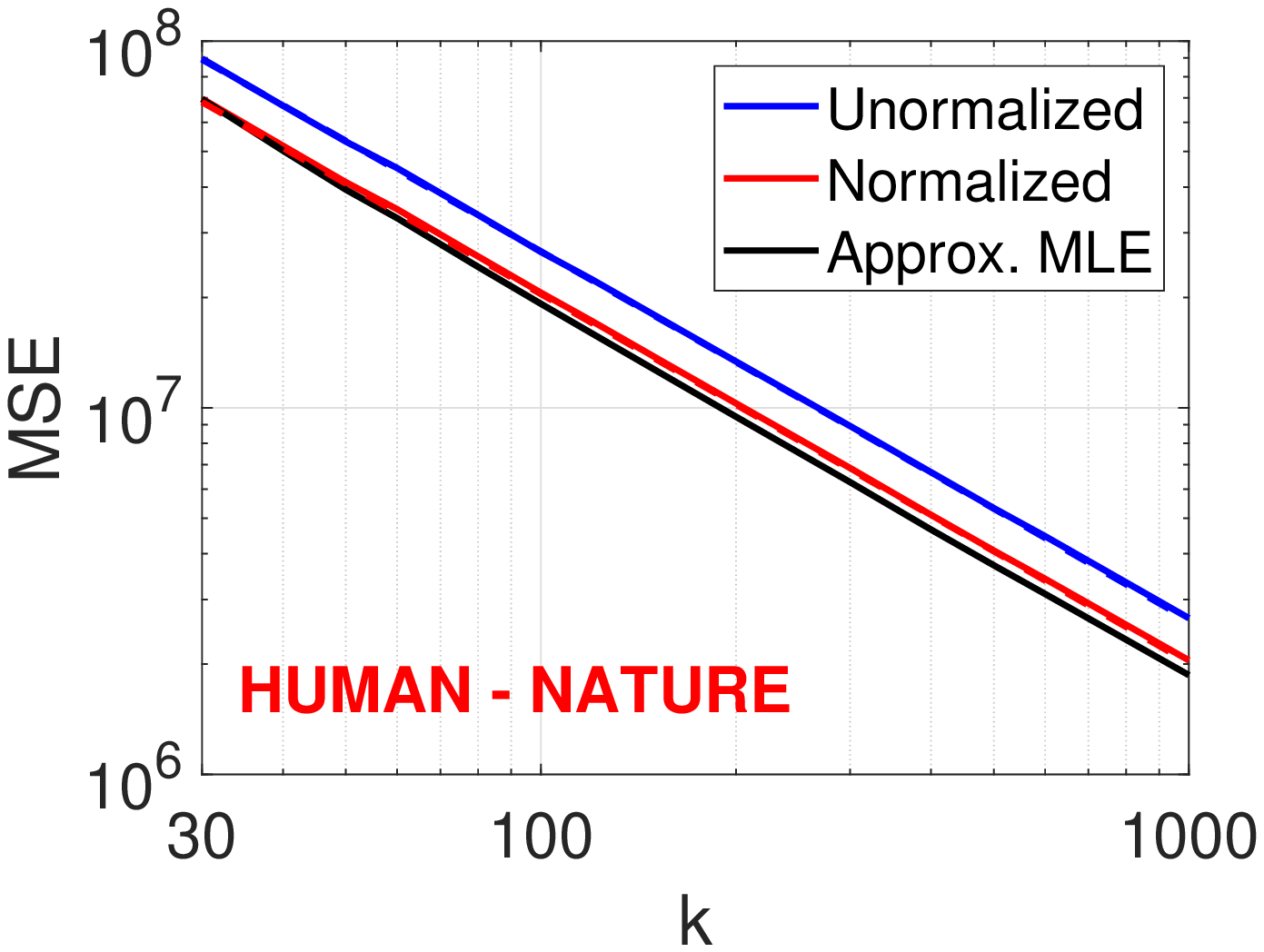}
    }    

\vspace{-0.15in}

    \caption{We estimate the inner products of the 9 pairs of words in Table~\ref{tab:words}, using the un-normalized estimator $\hat{a}$, the normalized estimator $\hat{a}_n$, as well as the approximate MLE estimator $\hat{a}_m$. We also plot, as dashed curves, the theoretical variances for $\hat{a}$ and $\hat{a}_n$. As expected, for $\hat{a}$, the empirical MSEs overlap with the theoretical variances. The normalized estimator $\hat{a}_n$ is considerably more accurate than $\hat{a}$, especially for word pairs with higher similarities. Also, for $\hat{a}_n$, the empirical MSEs do not differ much from the theoretical asymptotic variances. The ``approximate MLE'' $\hat{a}_m$ is still more accurate than the normalized estimator $\hat{a}_n$, although the differences are quite small.    }
    \label{fig:my_label}\label{fig:mle}
\end{figure}

Table~\ref{tab:words} lists 9 word-pairs from the ``Words'' dataset~\citep{li2005using}. Basically, each word represents a vector of length $2^{16}$ and each entry of the vector records the number of documents that word appears in a collection of $2^{16}$ documents. The selected 9 pairs cover a variety of scenarios (high sparsity versus low similarity, high similarity versus low similarity, etc).

Next, we compare the two inner product estimators $\hat{a}$ and $\hat{a}_n$ for these 9 pairs of words. In order to provide a more complete picture, we also add another estimator based on the (approximate) maximum likelihood estimation (MLE). Because characterizing the exact joint distribution of $(x_j, y_j), j= 1, 2, ..., k$ would be too complicated, we resort to the MLE for the standard Gaussian random projections, as studied in~\citet{li2006improving}. Basically, they show that the estimator $\hat{a}_m$, which is the solution to the following  cubic equation:
\begin{align}\notag
\hat{a}_m^3 -\hat{a}_m^2\sum_{j=1}^k x_jy_j+\hat{a}_m\left(-\sum_{i=1}^Du_i^2\sum_{i=1}^Dv_i^2+\sum_{i=1}^Du_i^2\sum_{j=1}^ky_j^2 + \sum_{i=1}^Dv_i^2\sum_{j=1}^kx_j^2 \right) - \sum_{i=1}^Du_i^2\sum_{i=1}^Dv_i^2\sum_{j=1}^k x_jy_j  = 0.
\end{align}

\newpage

The MLE has the smallest estimation variance if the margins $\sum_{i=1}^D u_i^2$ and $\sum_{i=1}^Dv_i^2$ are known. Obviously, the estimator $\hat{a}_m$ can no longer be written as an inner product (i.e., $\hat{a}_m$ is not a valid kernel for machine learning), unlike our $\hat{a}$ or $\hat{\rho}$ or $\hat{a}_n$.  Nevertheless, we can still use the MLE to assess the accuracy of estimators to see how close they are to be optimal. 

\vspace{0.1in}

Although we do not know the exact MLE for OPORP, we still use the above cubic equation as the ``surrogate'' for the MLE equation of OPORP and plot the empirical MSEs together  with the MSEs of $\hat{a}$ and $\hat{a}_n$ in Figure~\ref{fig:mle}, for estimating the inner products of the 9 pairs of words in Table~\ref{tab:words}.

\vspace{0.1in}

In each panel of Figure~\ref{fig:mle}, we present 5 curves: the empirical MSEs for $\hat{a}$, $\hat{a}_n$, and $\hat{a}_m$, and the theoretical variances for $\hat{a}$ and $\hat{a}_n$. As expected, for $\hat{a}$ (the un-normalized estimator), the empirical MSEs overlap with the theoretical variances. The normalized estimator $\hat{a}_n$ is considerably more accurate than the un-normalized estimator $\hat{a}$, especially for word pairs with higher similarities. Also, for $\hat{a}_n$, the empirical MSEs do not differ much from the theoretical asymptotic variances, although they do not fully overlap. Interestingly, the ``approximate MLE'' $\hat{a}_m$ is still more accurate than the normalized estimator $\hat{a}_n$, although the differences are quite small. 

\vspace{0.2in}

\begin{figure}[h]

\centering 

\mbox{
    \includegraphics[width=3in]{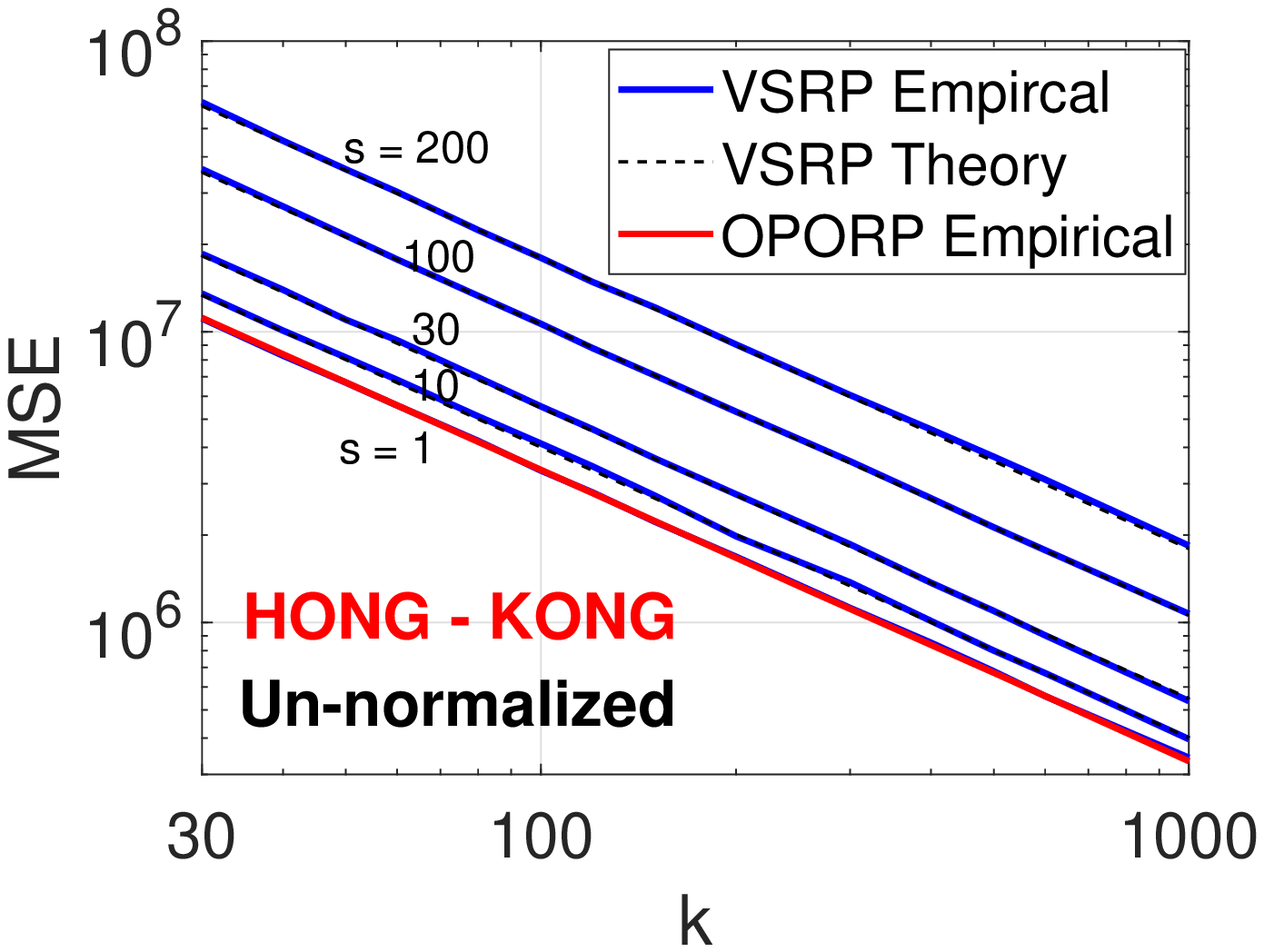}  
    \includegraphics[width=3in]{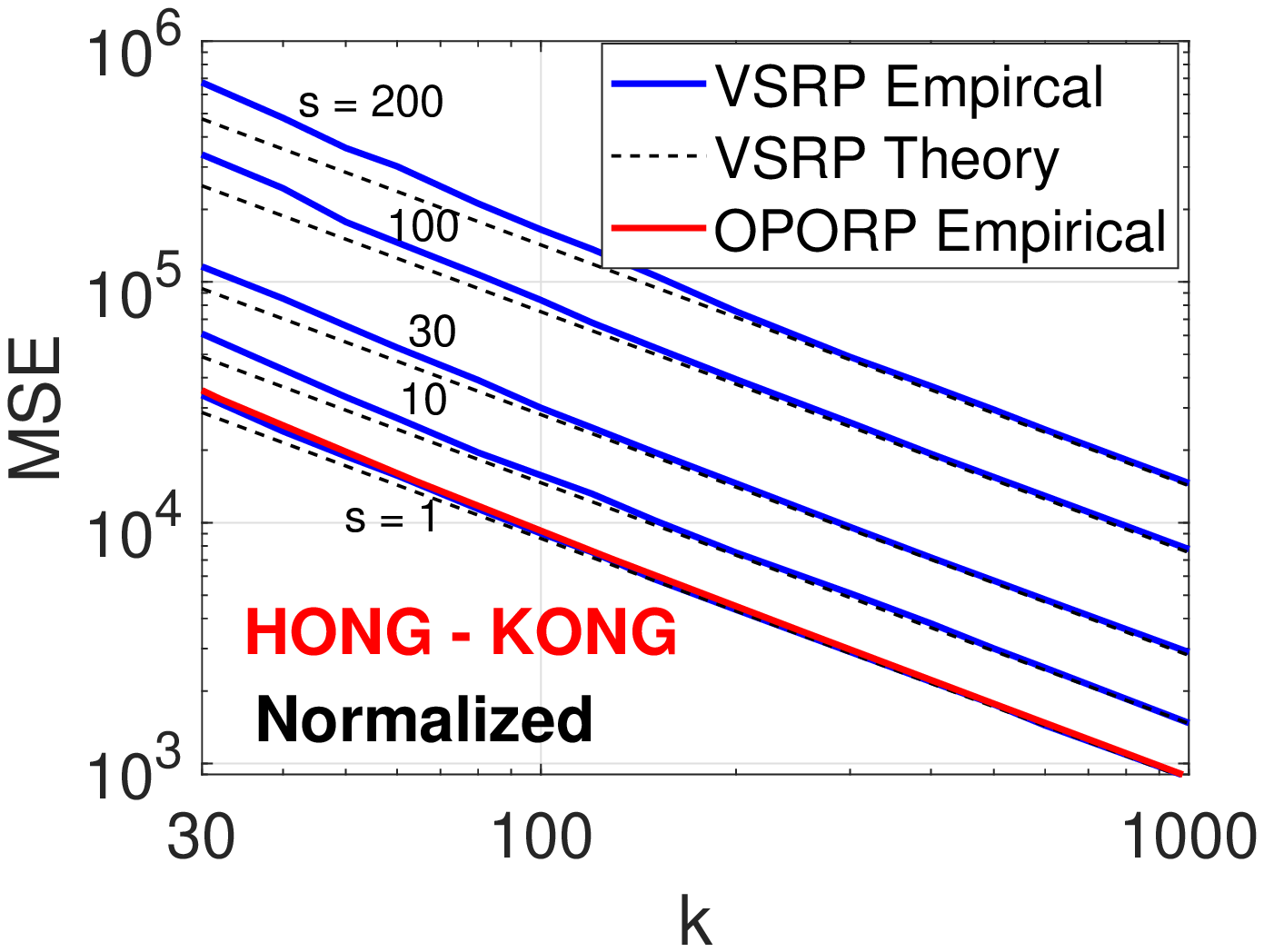}
}

\vspace{-0.1in}

    \caption{Comparing VSRP (with $s\in\{1,10,30,100,200\}$) with OPORP, in terms of their empirical MSEs, for both the un-normalized (left) and normalized (right) estimators, for the ``HONG-KONG'' word pair. As predicted by the theory, VSRP with $s=1$ essentially has the same accuracy as OPORP. Clearly, the normalized estimators are substantially more accurate than the un-normalized estimators. For the un-normalized VSRP estimator, the theoretical variance curves (dashed) overlap the solid MSE curves (solid). For the normalized VSRP estimator, the empirical MSEs slightly deviate from the theoretical variances (in Theorem~\ref{thm:vsrp}) when $k$ is small.   }
    \label{fig:vsrp_oporp}
\end{figure}

Finally, Figure~\ref{fig:vsrp_oporp} compares VSRP (for its $s\in\{1,10,30,100,200\}$) with OPORP, for both the normalized and un-normalize estimator, using the ``HONG-KONG'' word pair. The plots confirm the theoretical result in Theorem~\ref{thm:vsrp}. In this example, VSRP with $s=1$ has essentially the same MSEs as OPORP, as the theory predicts. Note that in this case $\frac{D-k}{D-1}$ is too small to be able to help OPORP to reduce the variance. As we increase $s$ for VSRP, the accuracy degrades quite substantially, again as predicted by the theory. We will observe the similar pattern in the experimental study in Section~\ref{sec:experiment}.

\section{Experiments}\label{sec:experiment}

We conduct experiments on two standard datasets: the MNIST dataset with 60000 training samples and 10000 testing samples, and the ZIP dataset (zipcode) with 7291 training samples and 2007 testing samples. The data vectors are normalized to have the unit $l_2$ norm. The MNIST dataset has 784 features and the ZIP dataset has 256 features. These dimensions well correspond  with typical EBR embedding vector sizes (i.e., $D=256\sim 1024$). 

\subsection{Retrieval}

In this experiment, we do not use the class labels. We treat the data vectors in the test sets as query vectors. For each query vector, we compute/estimate the cosine similarities with all the data vectors in the training set. For each estimation method, we rank the retrieved data vectors according to the estimated cosine similarities. In other words, there will be two ranked lists, one using the true cosines and the other using estimated cosines. By walking down the lists, we can compute the precision and recall curves. This allows us to compare OPORP with VSRP and their various estimators. Again, since the original data are already normalized, the inner product estimators are also cosine estimators. This makes it convenient to present the comparisons. 

\vspace{0.1in}

\begin{figure}[t]

    \centering
\mbox{
   \includegraphics[width=2.7in]{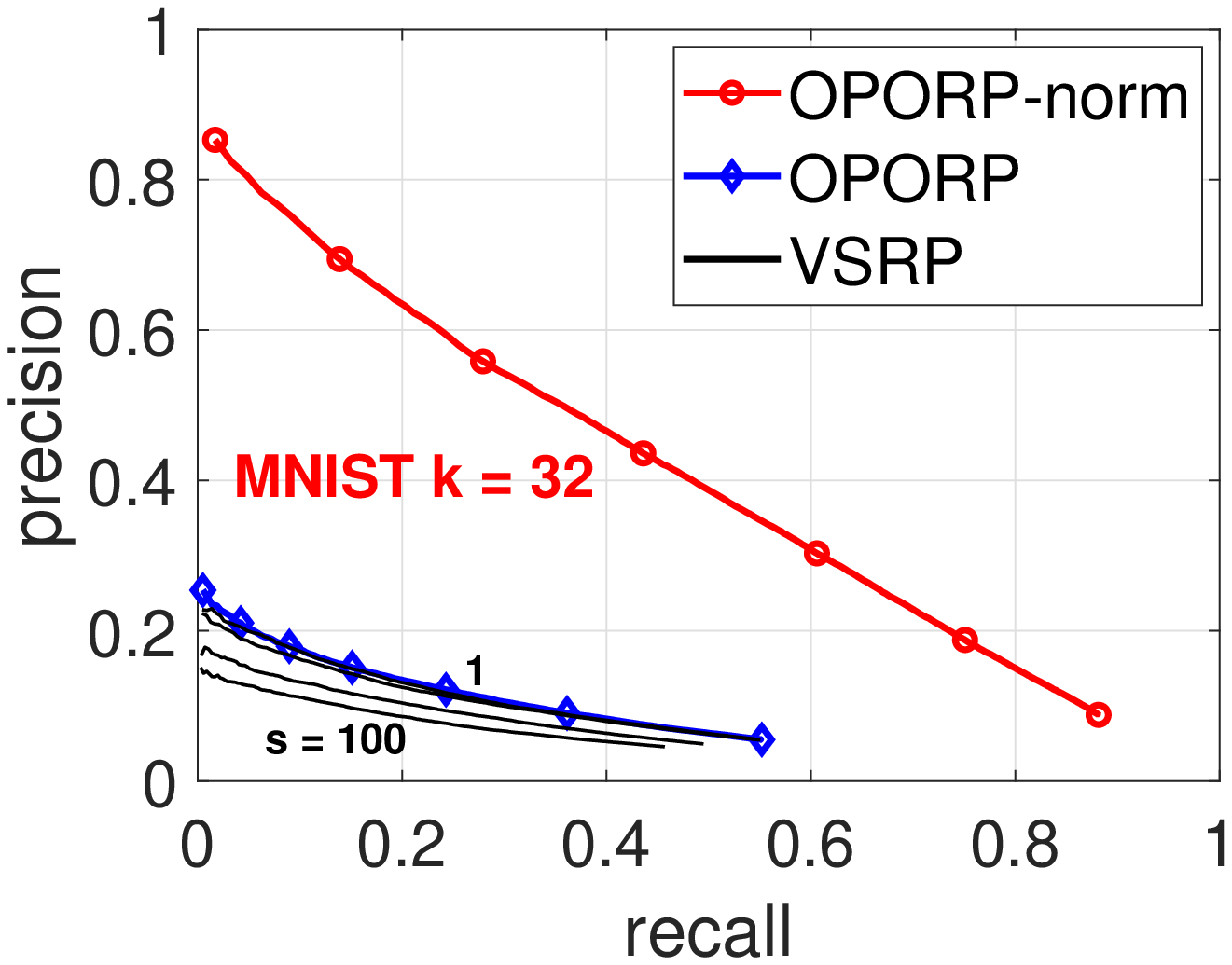}
   \includegraphics[width=2.7in]{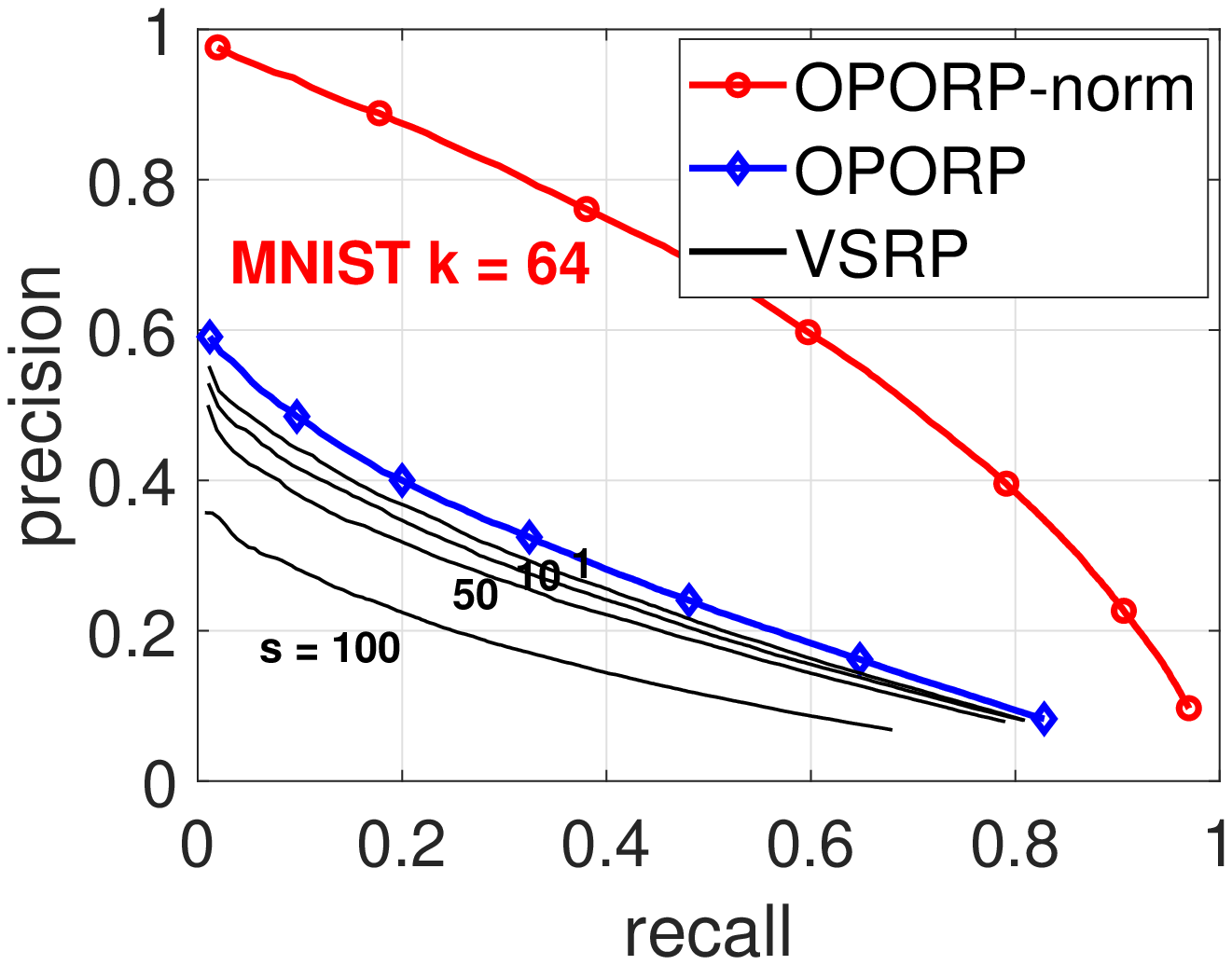}
}

\mbox{
   \includegraphics[width=2.7in]{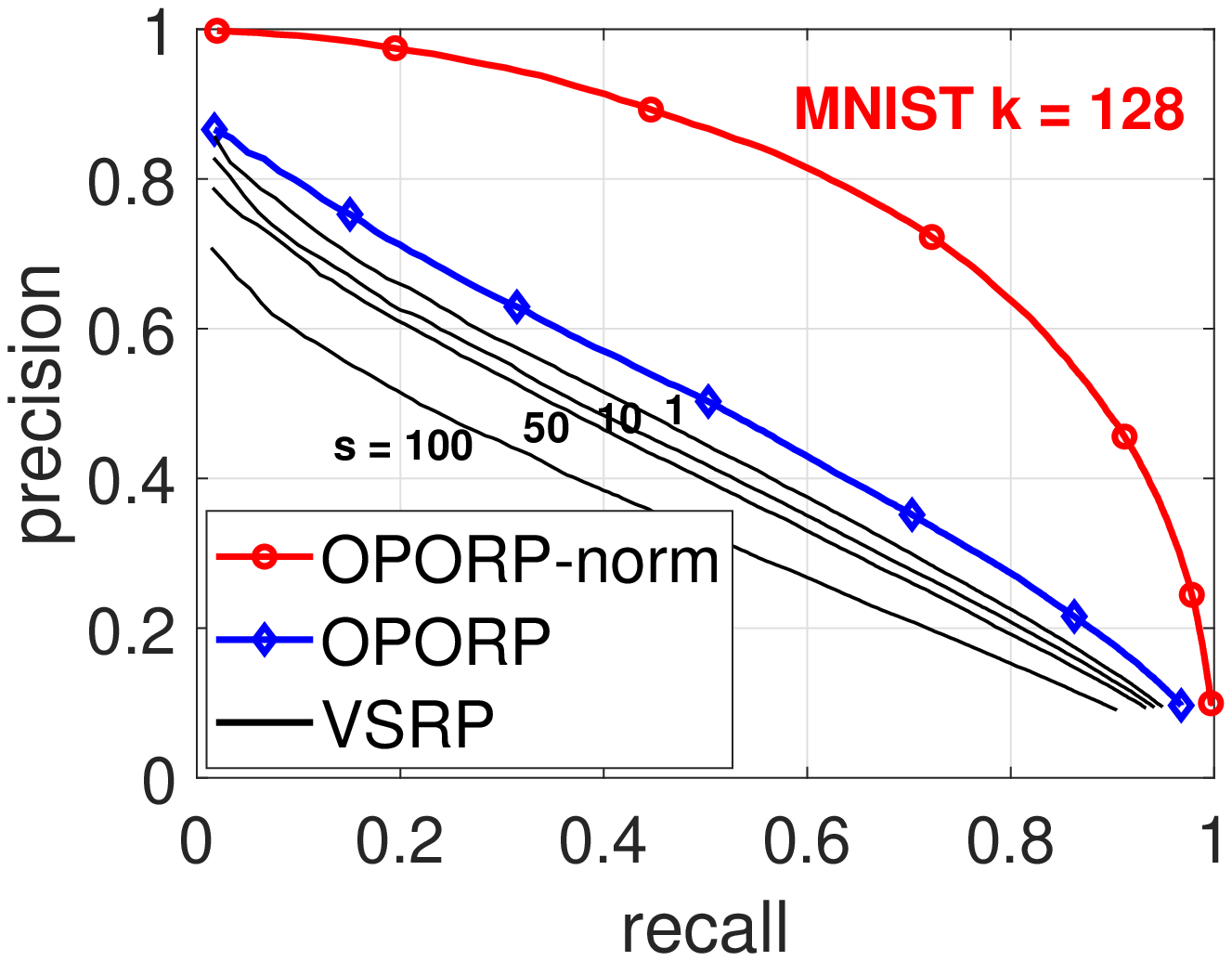}
   \includegraphics[width=2.7in]{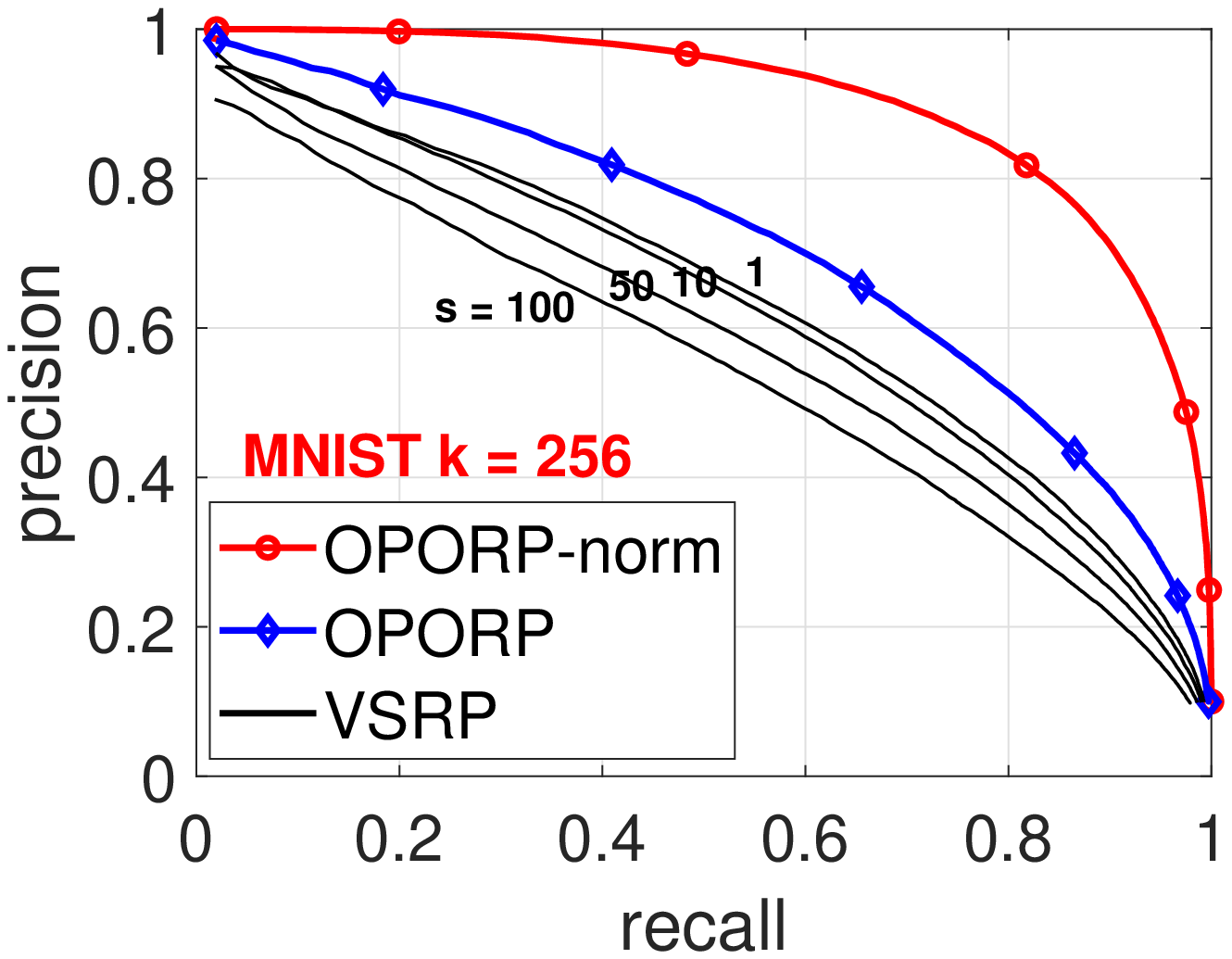}
}

\vspace{-0.2in}

   \caption{Precision-recall curves for MNIST (top-50) retrieval, using estimated cosines from the OPORP normalized estimator $\hat{\rho}$, the OPORP un-normalized estimator $\hat{a}$ (note that the original data are normalized), and the VSRP (parameterized by $s$) inner product estimator for $s\in\{1,10,50,100\}$.  }
    \label{fig:MNIST_pr}
\end{figure}

Figure~\ref{fig:MNIST_pr} presents the precision-recall curves for retrieving the top-50 candidates on MNIST. The curves for top-10 are pretty similar. As expected, the OPORP normalized estimator performs much better than the un-normalized inner product estimator of OPORP, for all $k\in\{32, 64, 128, 256\}$. The comparisons with VSRP (parameterized by $s$) are very interesting. Recall that VSRP with $s=1$ has the same variance as the un-normalized estimator of OPORP except for the $\frac{D-k}{D-1}$ term. In Figure~\ref{fig:MNIST_pr}, it is clear that the un-normalized OPORP estimator performs better than VSRP, which is due to the $\frac{D-k}{D-1}$ term. This effect is especially obvious for $k=256$ and $k=128$. By increasing $s$ for VSRP, we can observe deteriorating performances. In particular, when $s=100$ (i.e., the projection matrix of VSRP is extremely sparse), the loss of accuracy might be unacceptable. 

\begin{figure}[h!]
    \centering
\mbox{
   \includegraphics[width=2.7in]{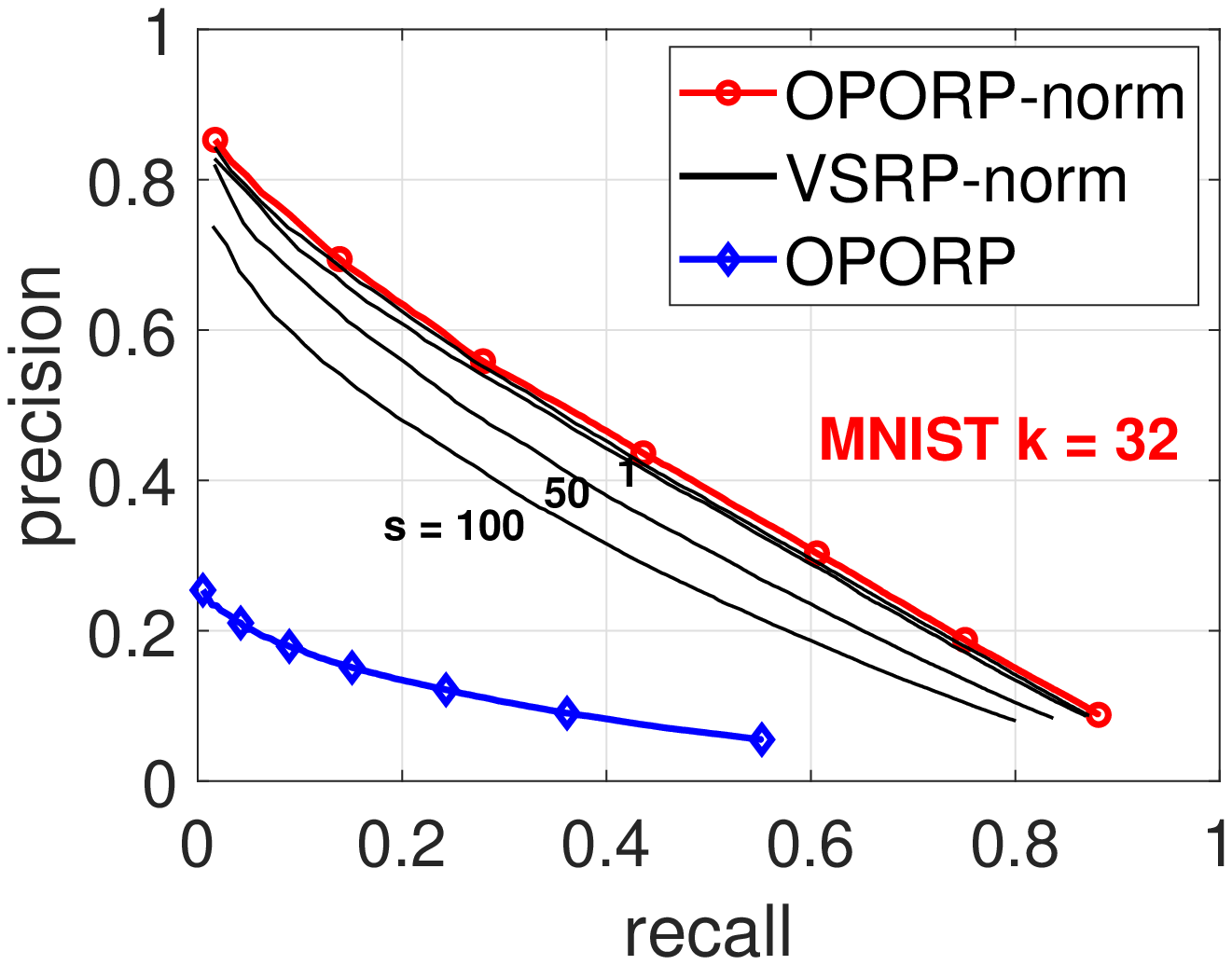}
   \includegraphics[width=2.7in]{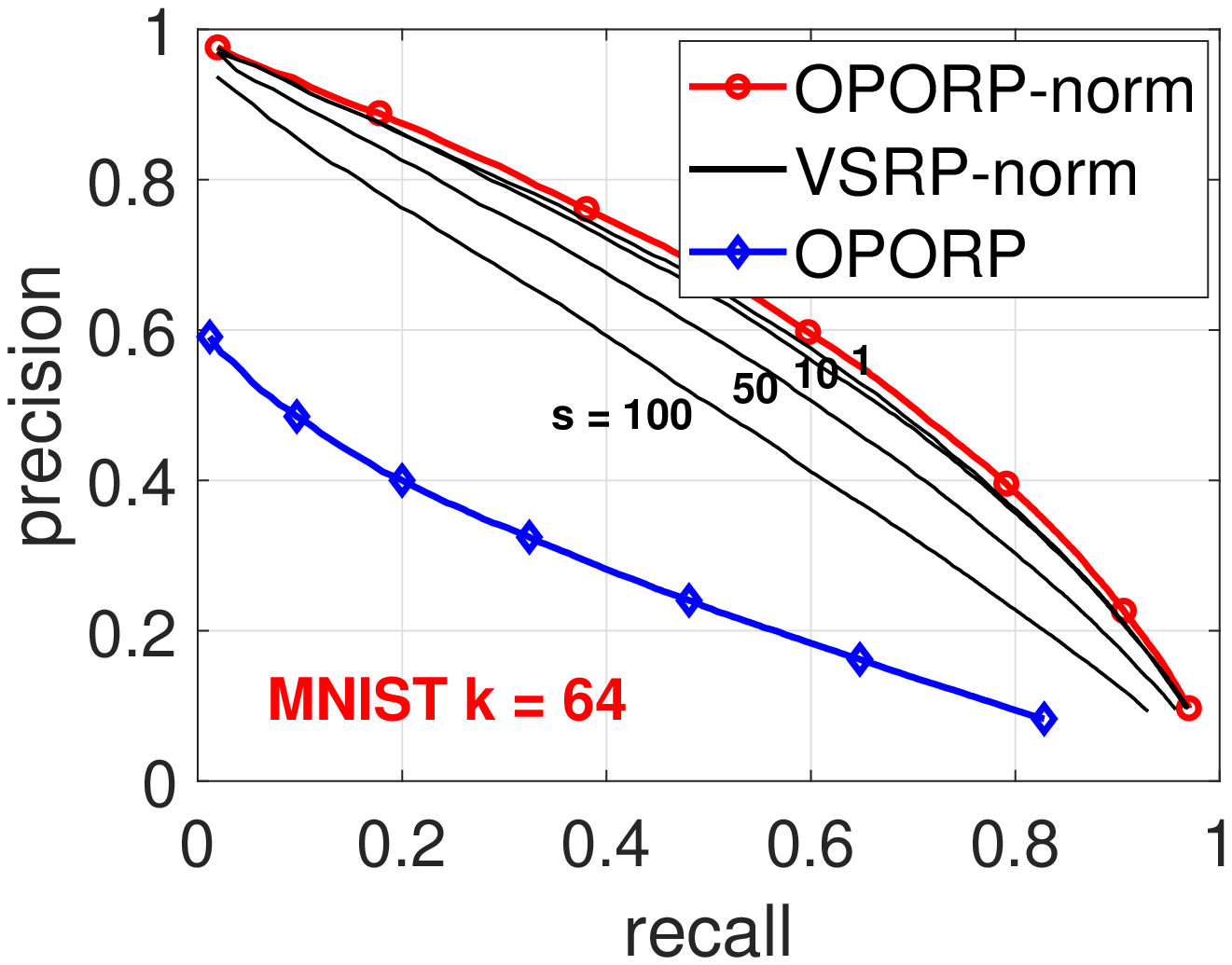}
}

\mbox{
   \includegraphics[width=2.7in]{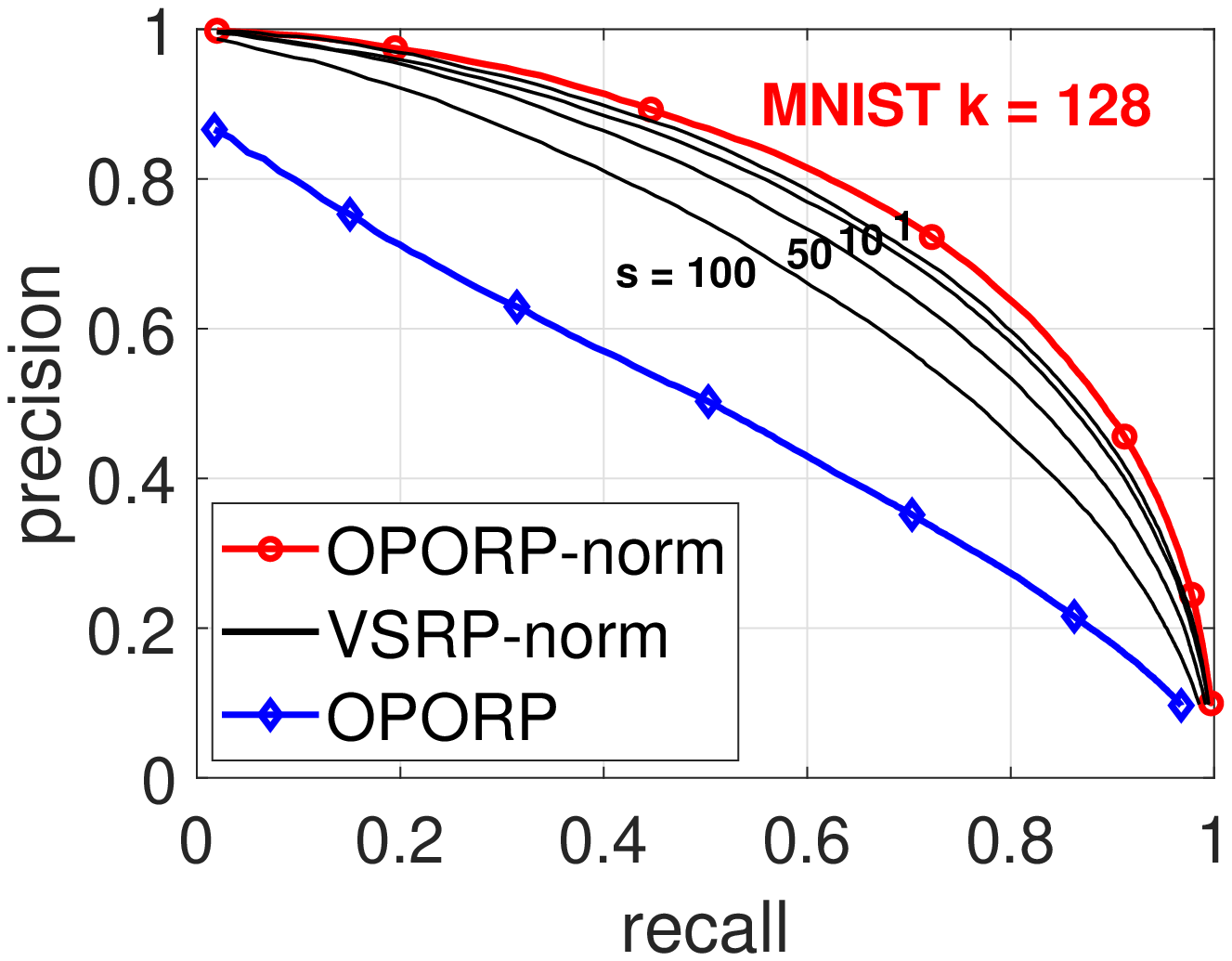}
   \includegraphics[width=2.7in]{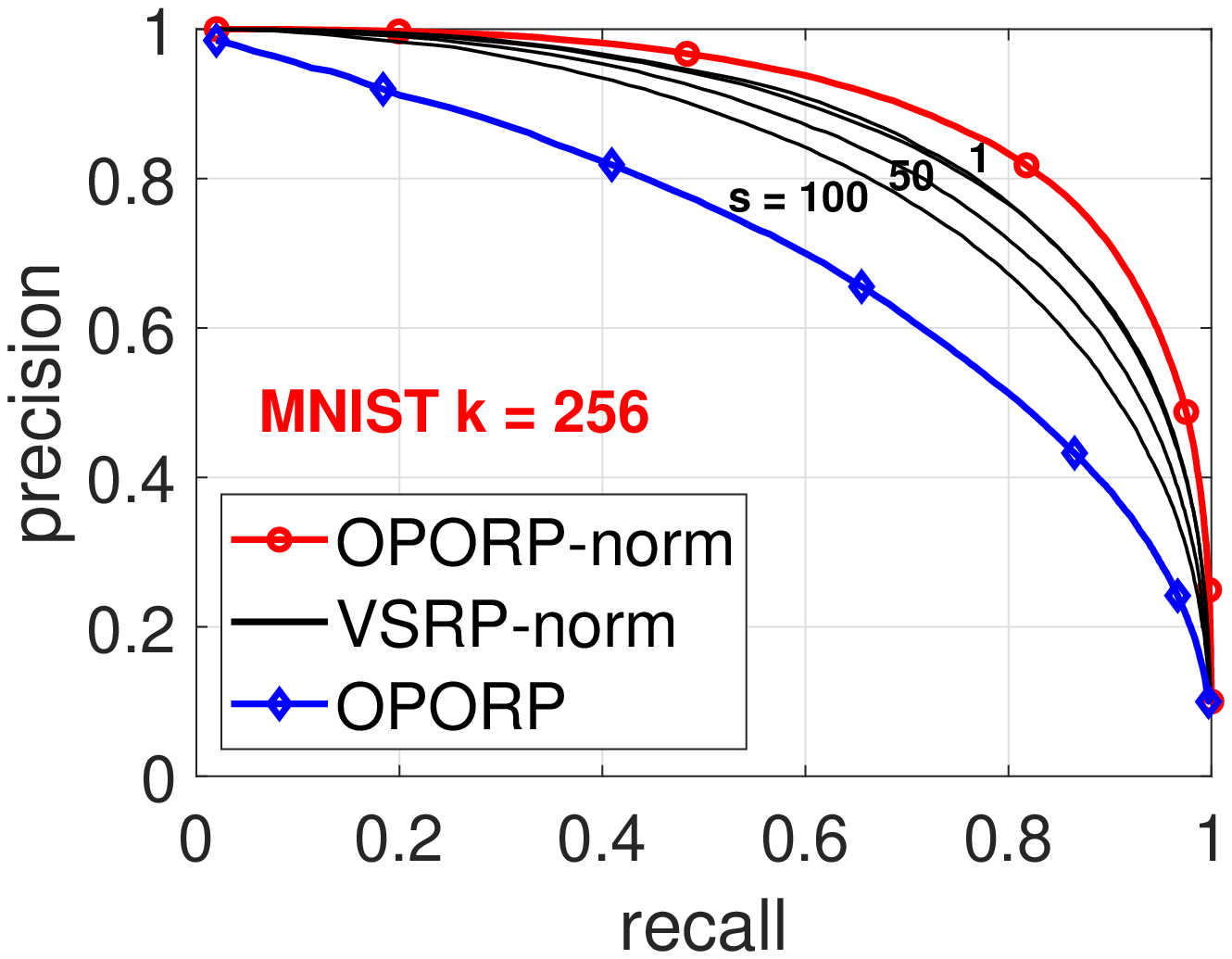}
}

\vspace{-0.15in}

   \caption{The content is pretty similar to that of Figure~\ref{fig:MNIST_pr}, but this time we normalize the estimator of VSRP (parameterized by $s\in\{1,10,50,100\}$).}
    \label{fig:MNIST_pr2}
\end{figure}

\newpage

Figure~\ref{fig:MNIST_pr2} is quite similar to Figure~\ref{fig:MNIST_pr} except that Figure~\ref{fig:MNIST_pr2} presents the normalized inner product estimator of VSRP, again for  $s\in\{1,10,50,100\}$. Indeed, as already shown by theory, the normalized estimator of VSRP improves the accuracy considerably. On the other hand, we still observe that, when $s=1$ for VSRP, its accuracy is slightly worse than OPORP (due to the $\frac{D-k}{D-1}$ factor); and when $s=100$, there is a severe deterioration of performance. Figure~\ref{fig:MNIST_pr2} once again confirms that the normalization trick is an excellent tool, which ought to be taken advantage of.

\begin{figure}[h]


\centering

\mbox{\hspace{-0.15in}
    \includegraphics[width=2.7in]{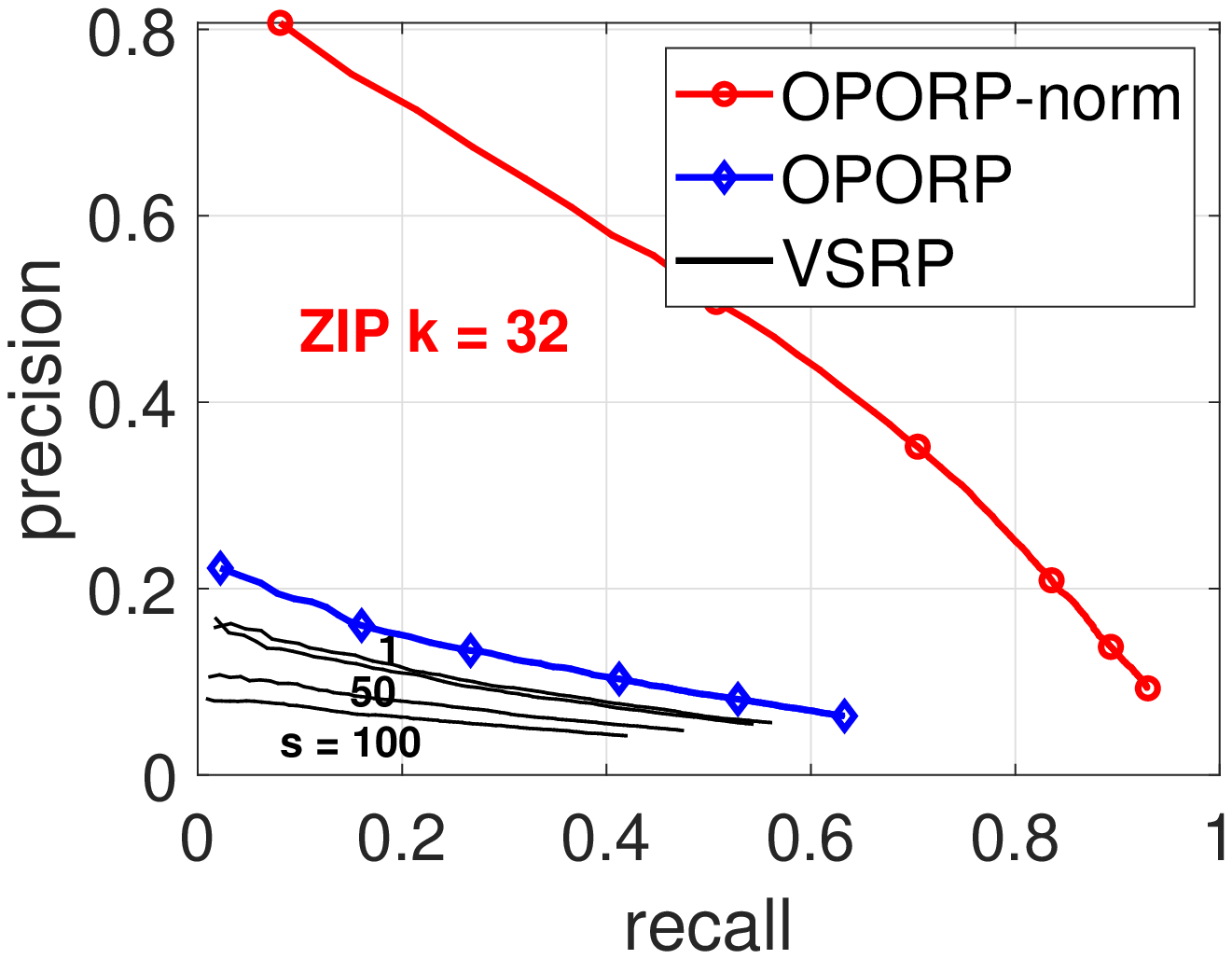}
    \includegraphics[width=2.7in]{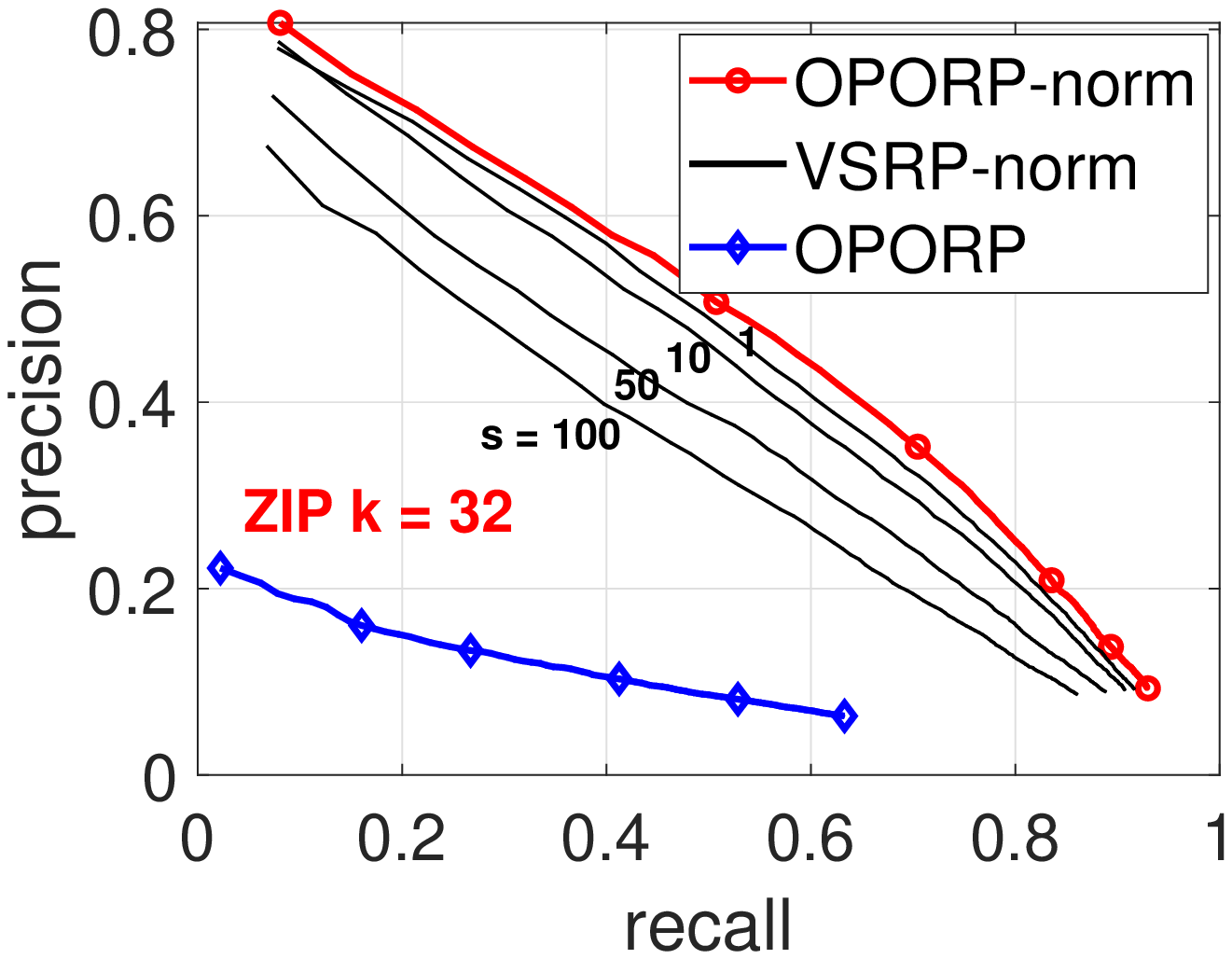}
    }
\mbox{\hspace{-0.15in}
   \includegraphics[width=2.7in]{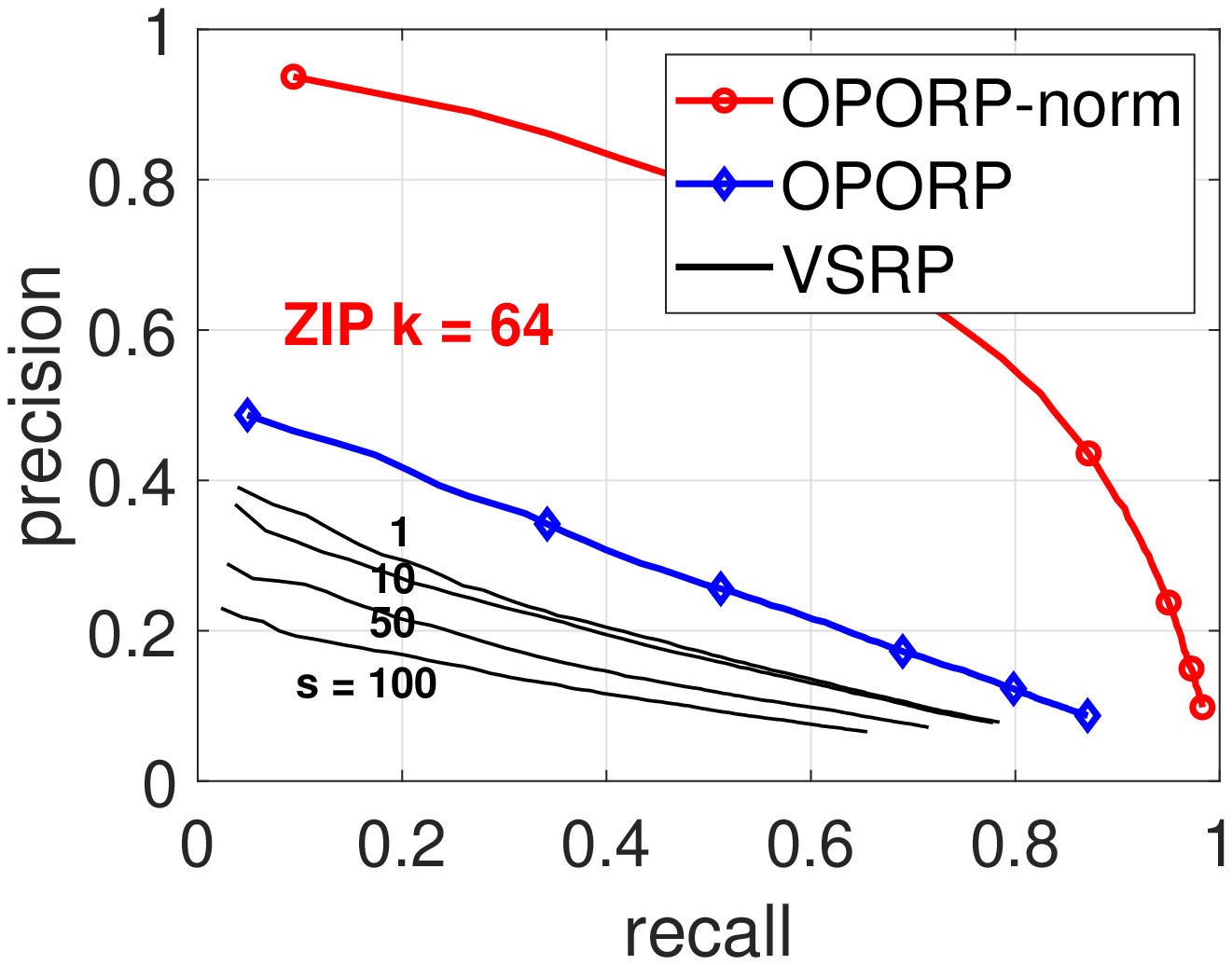}
   \includegraphics[width=2.7in]{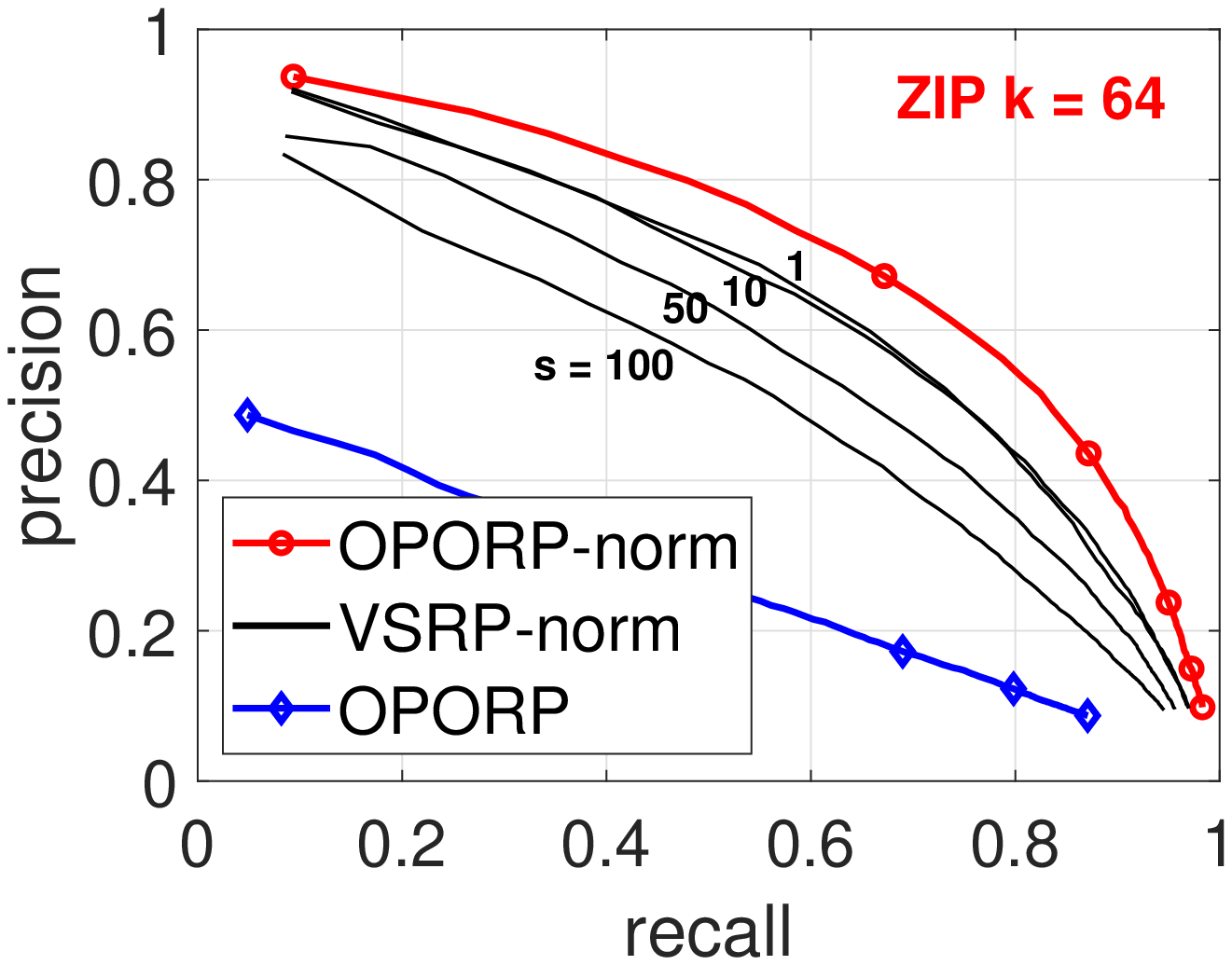}
   }
\mbox{\hspace{-0.15in}
   \includegraphics[width=2.7in]{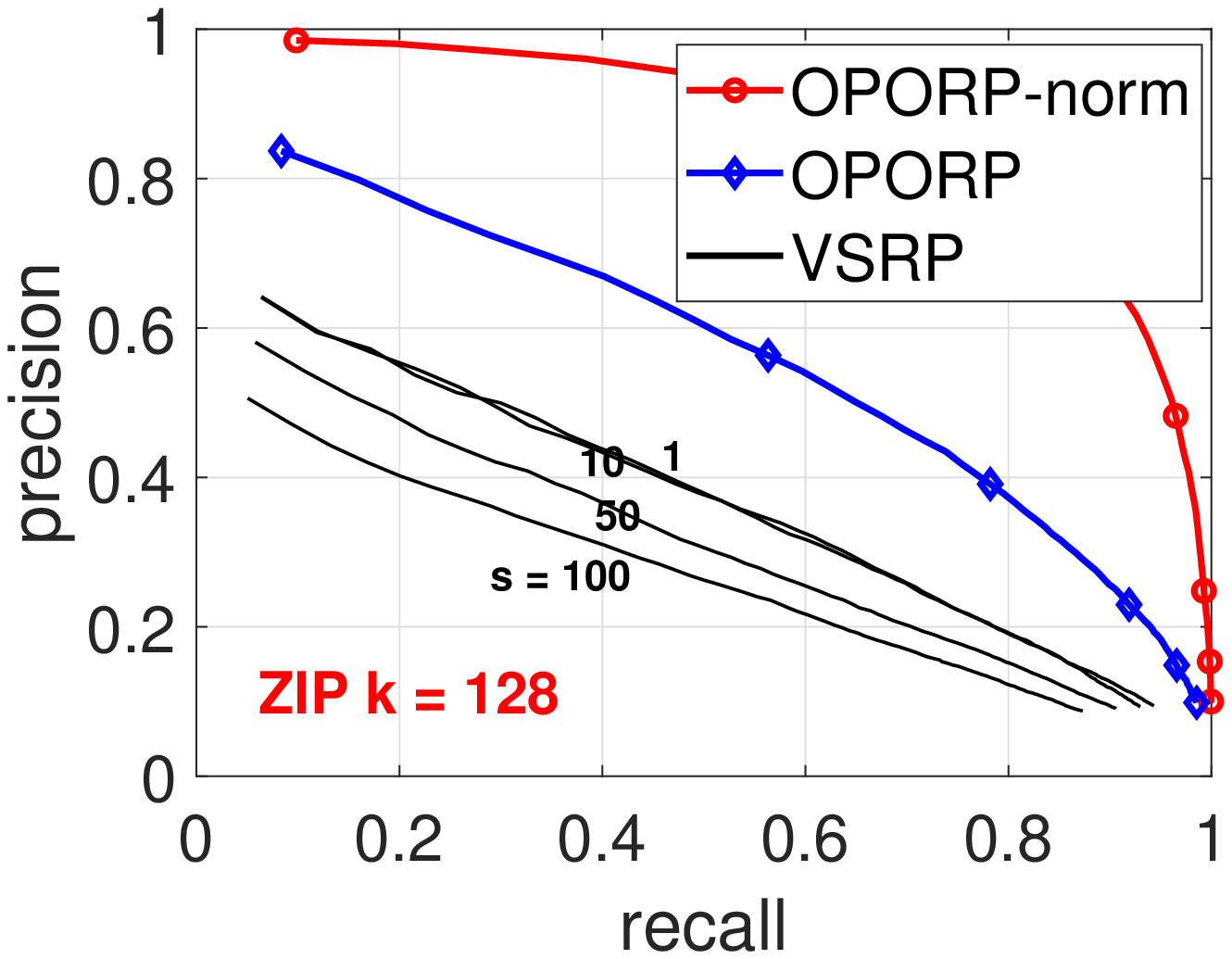}
   \includegraphics[width=2.7in]{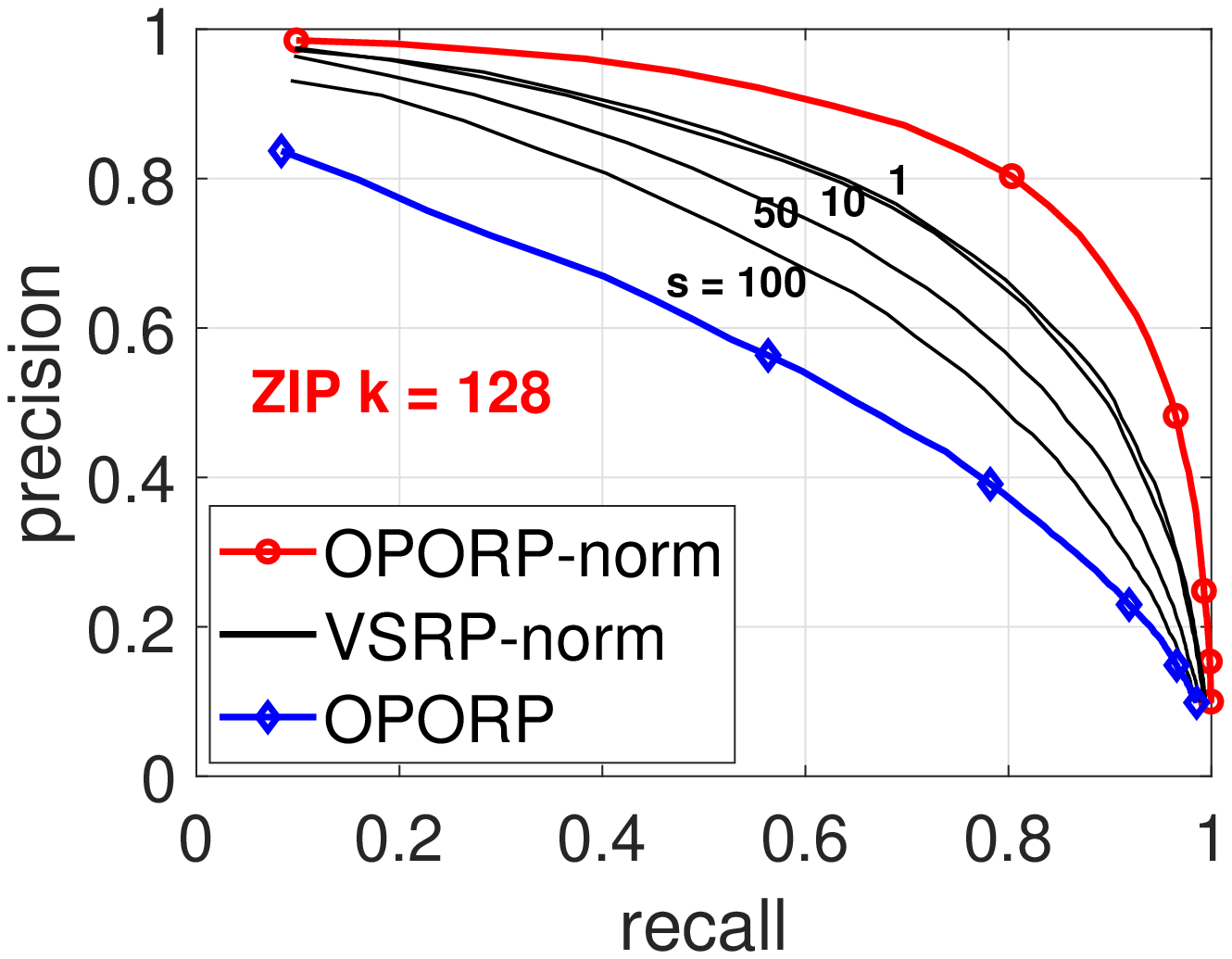}
   }
   
\vspace{-0.05in}

   \caption{Precision-recall curves for ZIP (top-10) retrieval. The left panels are analogous to Figure~\ref{fig:MNIST_pr} and the right panels are analogous to Figure~\ref{fig:MNIST_pr2} for MNIST retrieval. }
    \label{fig:ZIP_pr}\vspace{-0.1in}
\end{figure}

Figure~\ref{fig:ZIP_pr} presents the (top-10) retrieval experiments on the ZIP dataset. The plots are analogous to the plots in Figure~\ref{fig:MNIST_pr} and Figure~\ref{fig:MNIST_pr2}, with essentially the same conclusion. 

\vspace{0.2in}

\subsection{KNN classification}

\begin{figure}[h!]


    \centering
\mbox{
   \includegraphics[width=2.7in]{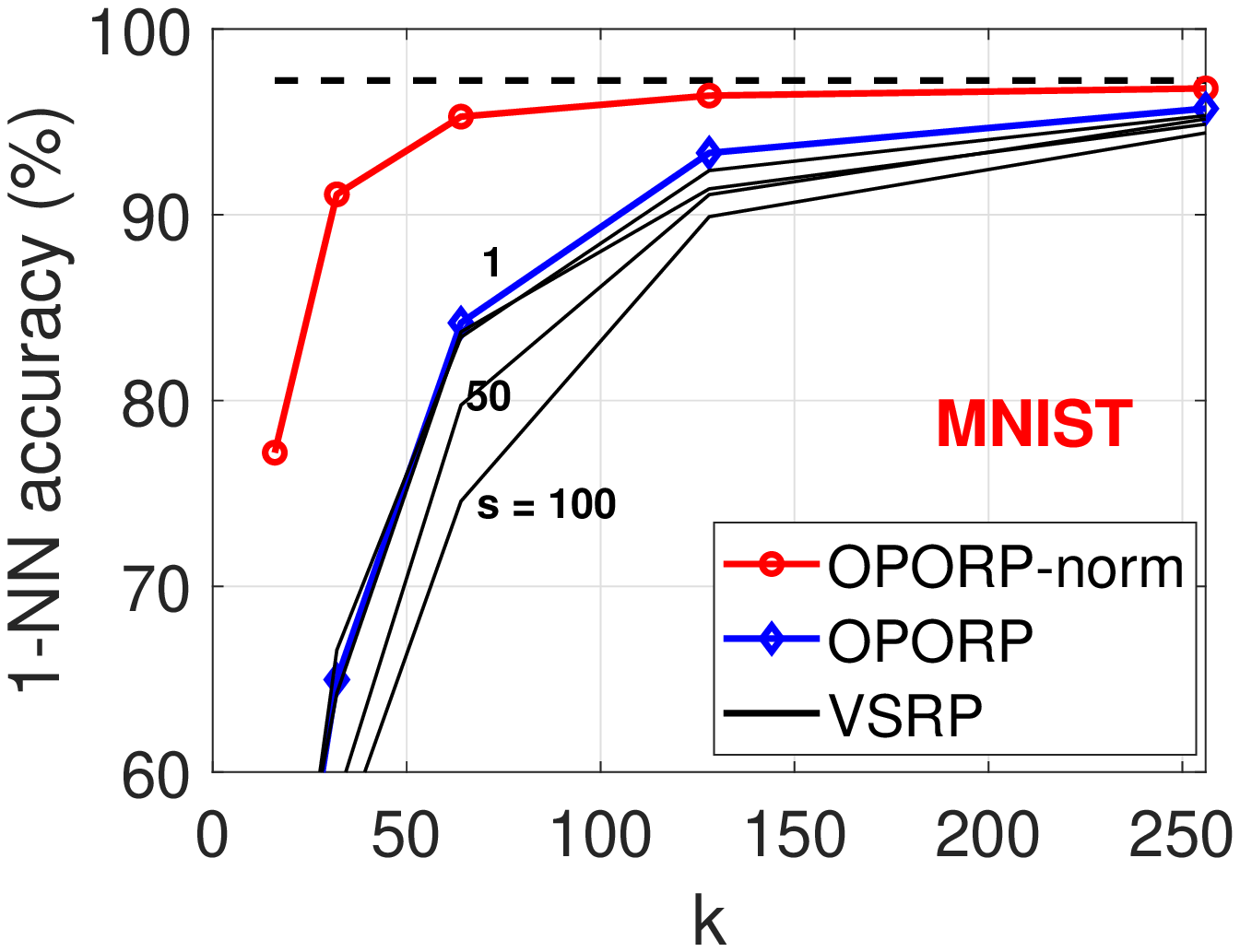}
   \includegraphics[width=2.7in]{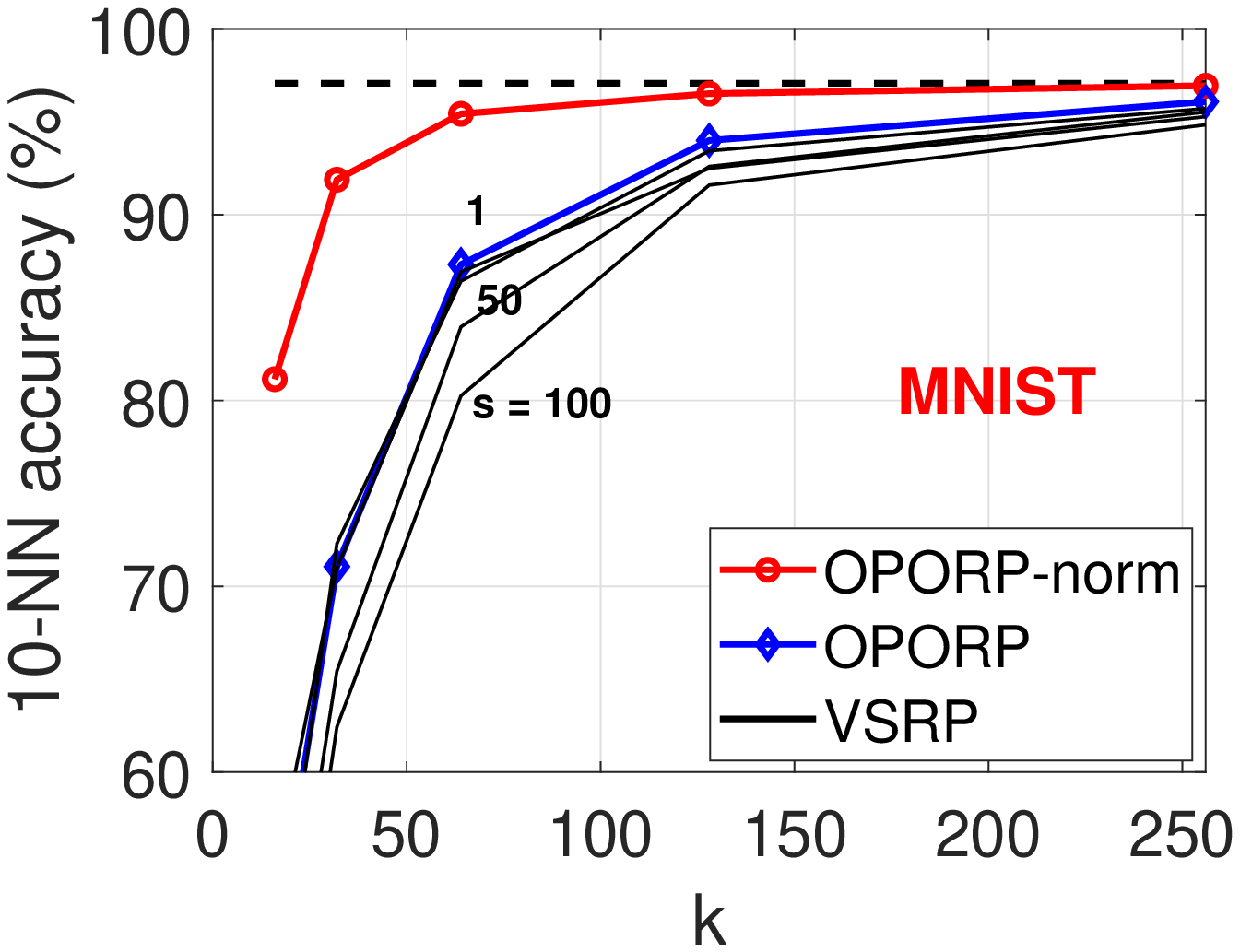}
}

\mbox{
   \includegraphics[width=2.7in]{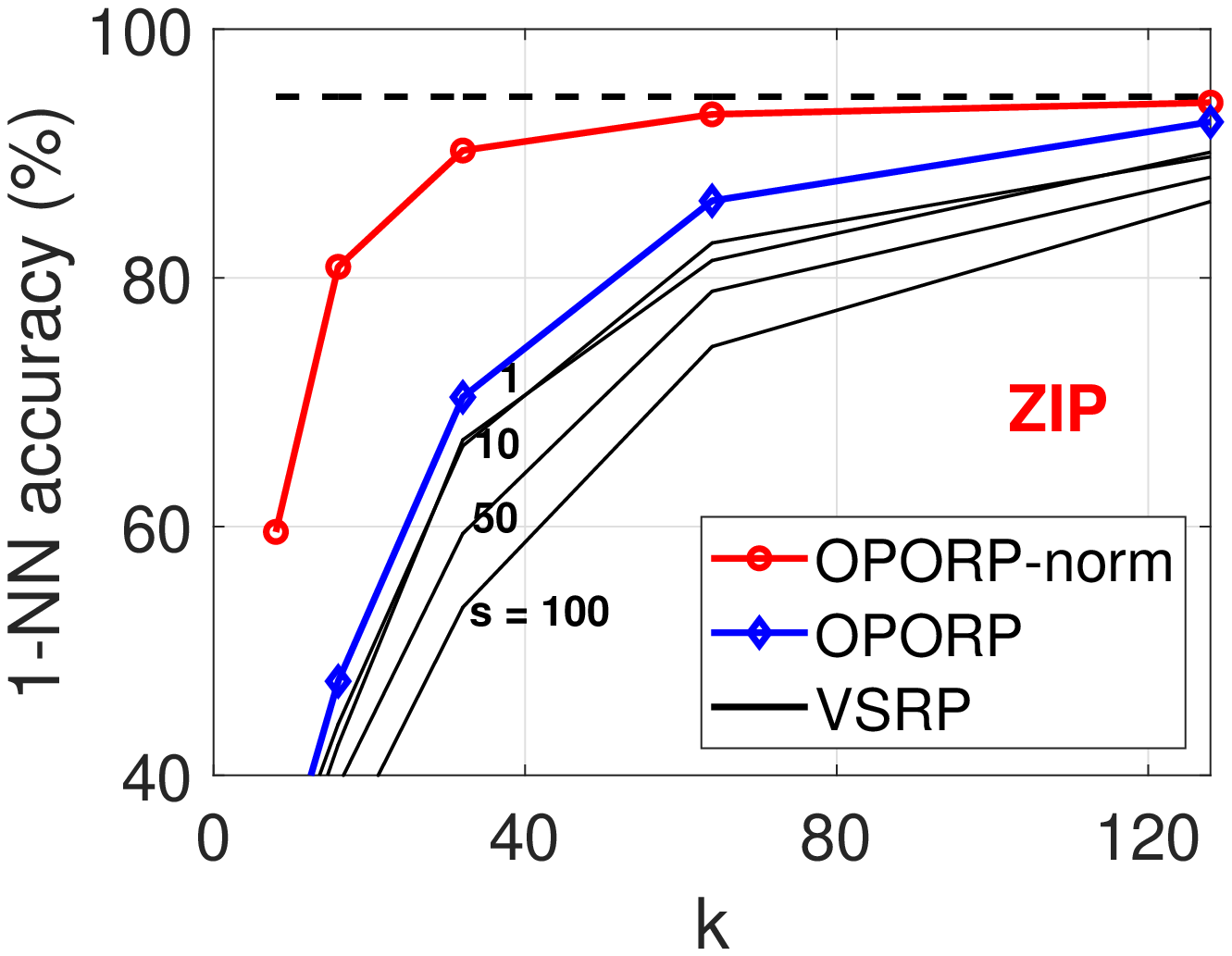}
   \includegraphics[width=2.7in]{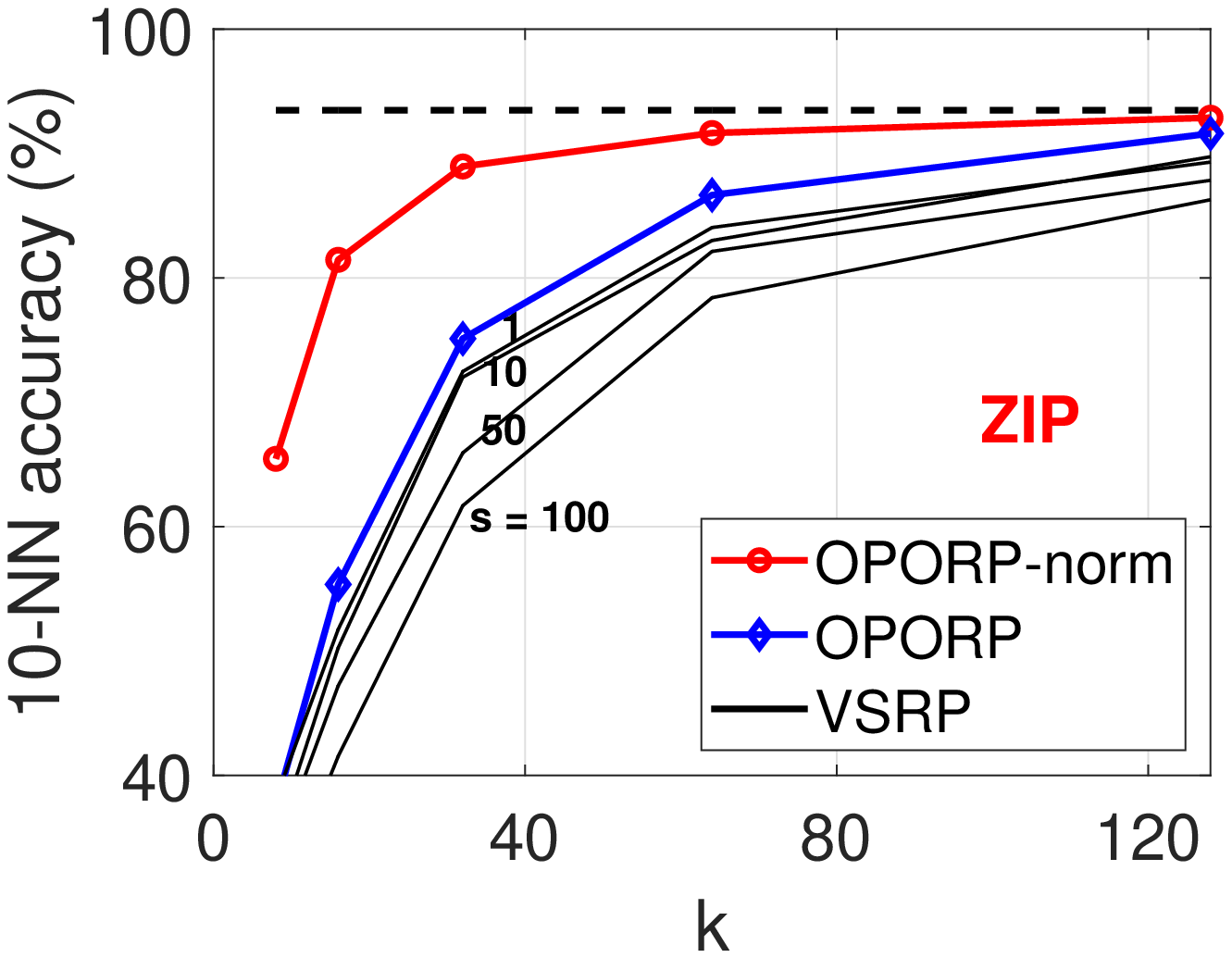}
}

\vspace{-0.15in}

   \caption{1-NN and 10-NN classification results using cosines. The horizontal dashed lines represent the results using the true cosines. The general trends are pretty much the same as  observed in the retrieval experiments in Figure~\ref{fig:MNIST_pr}. The vertical axis is the test classification accuracy. OPORP normalized estimator considerably improves OPORP un-normalized estimator. OPORP  improves  VSRP with $s=1$ (due to $\frac{D-k}{D-1}$). Again, using VSRP with large $s$ values leads to poor performance.}
    \label{fig:KNN}\vspace{-0.15in}
\end{figure}

Figure~\ref{fig:KNN} presents the experiments on KNN (K nearest neighbors) classification, in particular 1-NN and 10-NN, for both MNIST and ZIP datasets. We need the class labels for this set of experiments. In each panel, the vertical axis represents the test classification accuracy (in $\%$). The original classification accuracy (the dashed horizontal curve) is pretty high, but we can approach the same accuracy with OPORP using the normalized estimator (with e.g., $k\geq 128$ for MNIST and $k\geq 64$ for ZIP). The performance of the un-normalized estimator of OPORP is considerably worse. Also, OPORP  improves  VSRP with $s=1$ owing to the $\frac{D-k}{D-1}$ factor. Again, using VSRP with large $s$ values leads to poor performance.

\newpage\clearpage

\section{Differential Privacy: DP-OPORP and DP-SignOPORP}

In a recent work~\citep{li2023differential}, OPORP has been used as a basic building block for differential privacy (DP)~\citep{dwork2006calibrating} for RP-type data compression. DP is a standard privacy definition defined as below, whose mathematical intuition is that the distribution of the algorithm output should stay close when the dataset is changed by a little.

\begin{definition}[Differential Privacy~\citep{dwork2006calibrating}] \label{def:DP}
For a randomized algorithm $\mathcal M:\mathcal U\mapsto Range(\mathcal M)$, if for any two adjacent datasets $U$ and $U'$, it holds that
\begin{equation} \label{eq:DP-def}
    Pr[\mathcal M(U)\in O] \leq e^\epsilon Pr[\mathcal M(U')\in O]+\delta
\end{equation}
for $\forall O\subset Range(\mathcal M)$ and some $\epsilon,\delta\geq 0$,
then algorithm $\mathcal M$ is called $(\epsilon,\delta)$-differentially private. If $\delta=0$, $\mathcal M$ is called $\epsilon$-differentially private.
\end{definition}

The definition of ``neighboring'' is given as follows.

\begin{definition}[$\beta$-adjacency] \label{def:neighbor}
Let $u\in [-1,1]^p$ be a data vector. A vector $u'\in [-1,1]^p$ is said to be $\beta$-adjacent to $u$ if $u'$ and $u$ differ in one dimension $i$, and $|u_i-u_i'|\leq \beta$. 
\end{definition}

The privacy parameter $\beta$ is flexible depending on the application scenarios. \citet{li2023differential} proposed two variants of OPORP under DP: the full-precision DP-OPORP and the 1-bit DP-SignOPORP for even higher data compression rate. We briefly introduce these algorithms as follows.

\begin{figure}[t]
\centering

    \mbox{
    \includegraphics[width=2.7in]{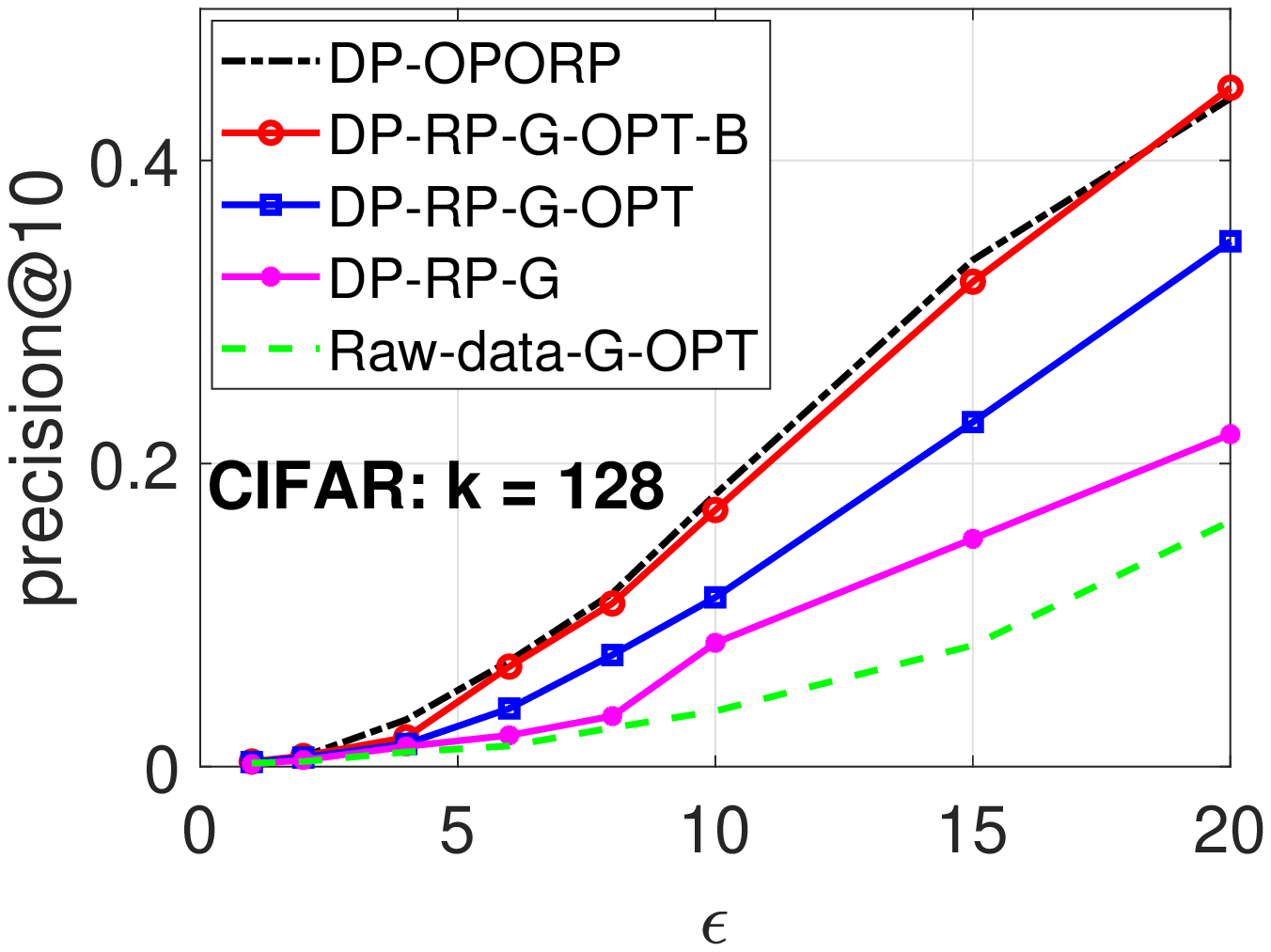}
    \includegraphics[width=2.7in]{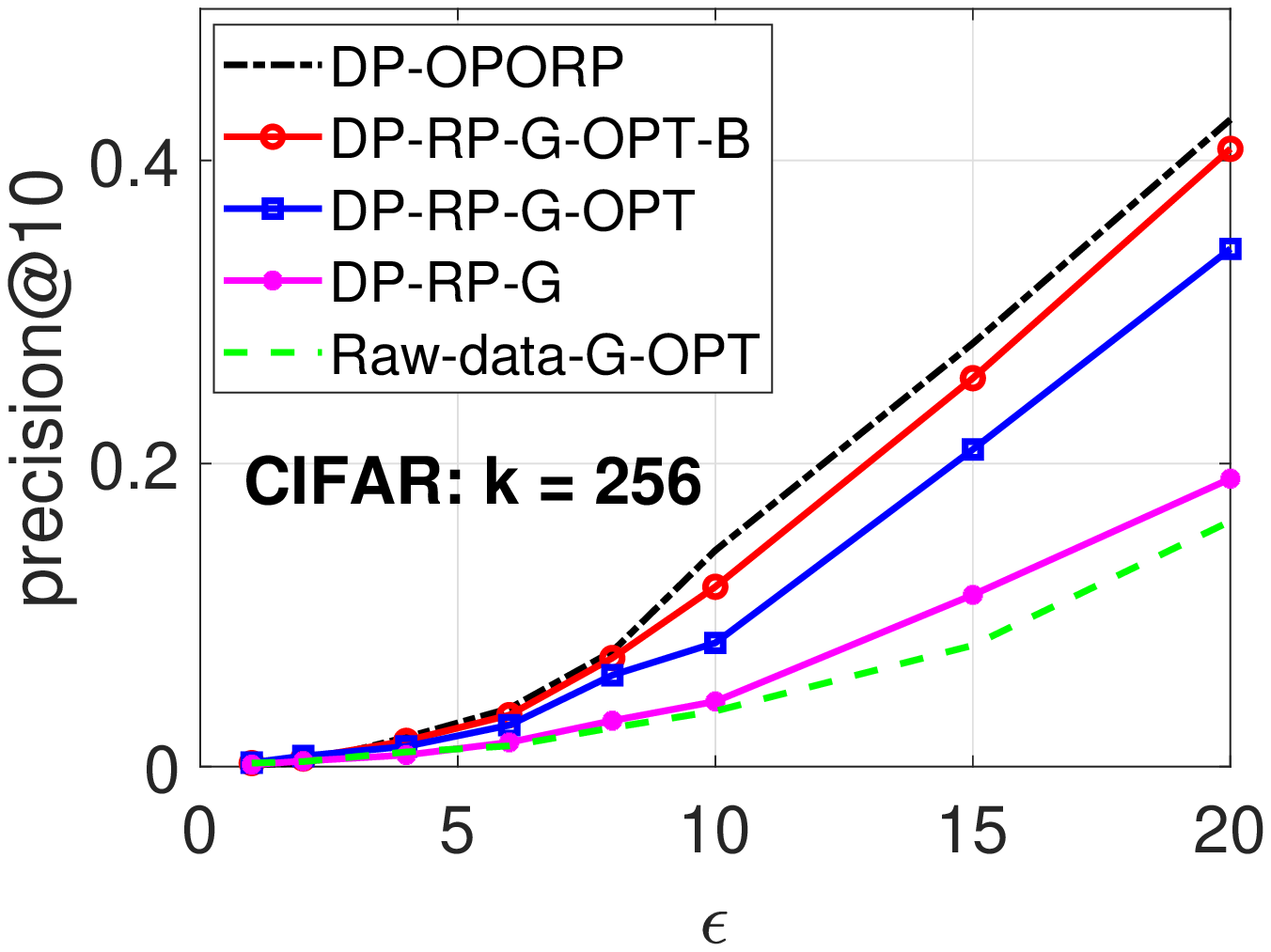}
    }

    \mbox{
    \includegraphics[width=2.7in]{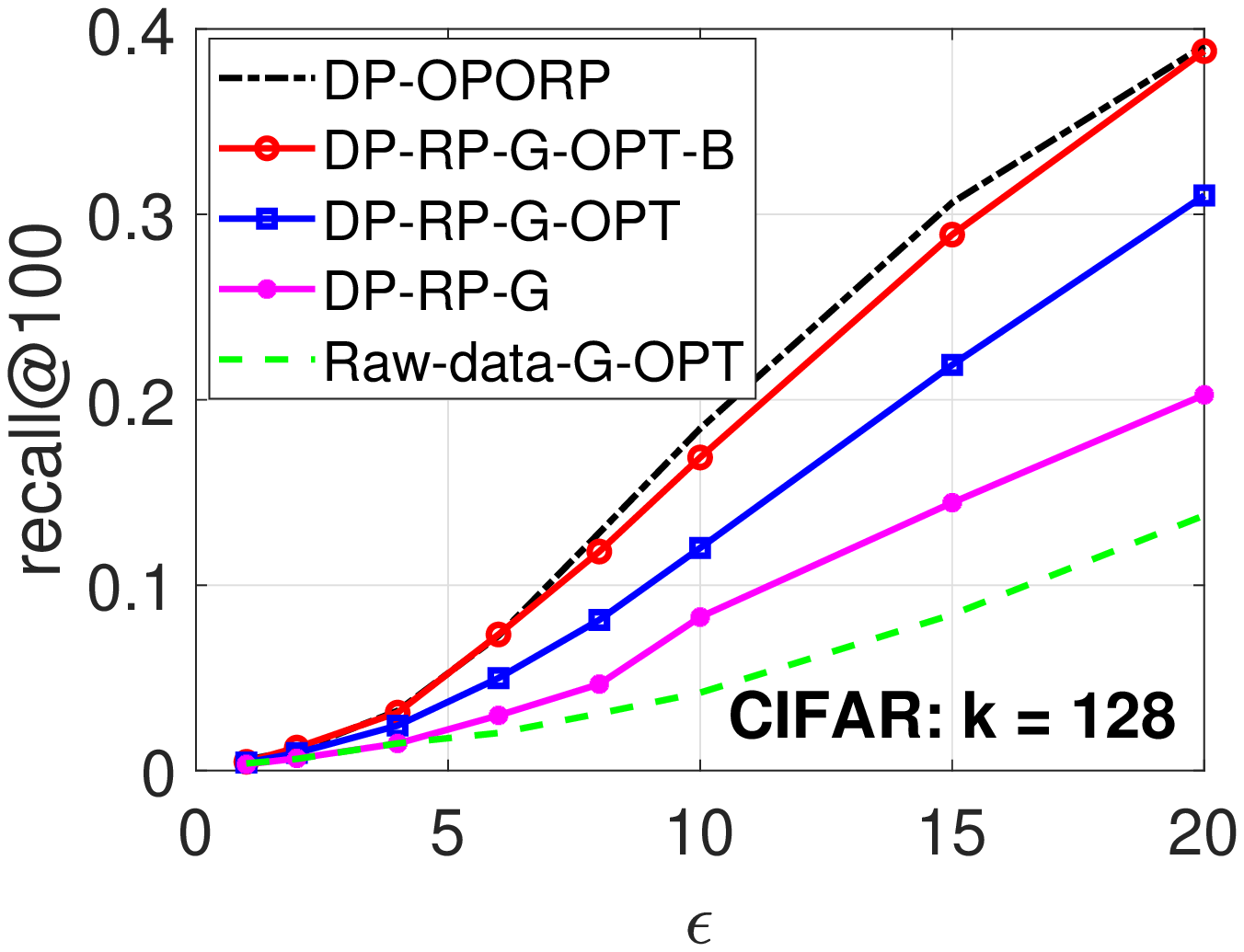}
    \includegraphics[width=2.7in]{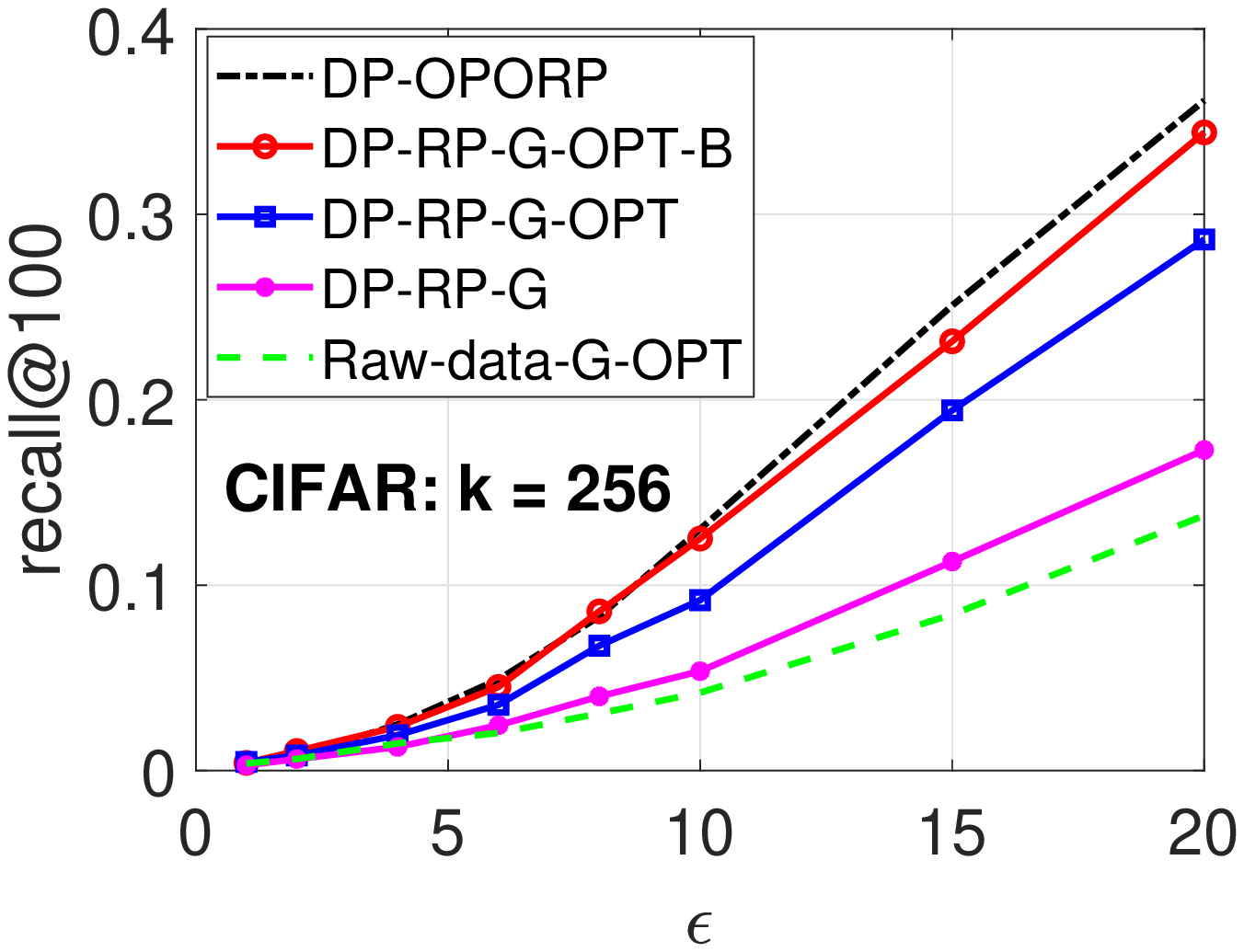}  
    }

\vspace{-0.15in}

\caption{\citep{li2023differential} Recall and precision on CIFAR, $\beta=1$, $\delta=10^{-6}$.}
\label{fig:CIFAR_vs_eps_DP-RP}
\end{figure}

\vspace{0.1in}
\noindent\textbf{DP-OPORP.} \hspace{0.1in} The DP-OPORP method is summarized in Algorithm~\ref{alg:DP-OPORP}. We first produce the (non-DP) OPORP samples (in $\mathbb R^k$), and then add a random Gaussian noise vector to the OPORP. At Line 5, the Gaussian noise magnitude $\sigma$ is computed by solving the equation for $\sigma^*$:
\begin{align}
    \Phi\left(\frac{\triangle_2}{2\sigma}-\frac{\epsilon \sigma}{\triangle_2}\right) - e^\epsilon \Phi\left(-\frac{\triangle_2}{2\sigma}-\frac{\epsilon \sigma}{\triangle_2}\right)= \delta,  \label{eqn:optimal Gaussian}
\end{align}
which is the ``optimal Gaussian mechanism''~\citep{balle2018improving}, where $\triangle_2$ is the ``$l_2$-sensitivity''. Algorithm~\ref{alg:DP-OPORP} achieves $(\epsilon,\delta)$-DP, and is superior to the best DP variant for random projections (DP-RP) both theoretically (in terms of inner product estimation variance) and empirically (on similarity search and SVM classification). Figure~\ref{fig:CIFAR_vs_eps_DP-RP} 
 (from~\citet{li2023differential}) compares several DP algorithms based  on random projections to confirm that DP-OPORP achieves the overall best performance.

\begin{algorithm}[h]
	{
		\textbf{Input:} Data $u\in[-1,1]^p$, privacy parameters $\epsilon>0$, $\delta\in (0,1)$, number of projections~$k$
		
		\textbf{Output:}   Differentially private OPORP
		
		Generate the OPORP of $u$ as $x=OPORP(u)$
		
		Set sensitivity $\triangle_2=\beta$
		
		Generate iid random vector $G\in\mathbb R^k$ following $N(0,\sigma^2)$ where $\sigma$ is computed by (\ref{eqn:optimal Gaussian})

		Return  $\tilde x =x+G$
	}
	\caption{DP-OPORP~\citep{li2023differential}}
	\label{alg:DP-OPORP}
\end{algorithm}

\begin{algorithm}[t]
{
    \vspace{0.05in}
    \textbf{Input:} Data $u\in[-1,1]^p$; $\epsilon>0$; Number of projections~$k$
     
    \vspace{0.05in}
    
    \textbf{Output:}   Differentially private sign OPORP

    \vspace{0.05in}
    
    Generate the OPORP of $u$ as $x=OPORP(u)$
    
    \vspace{0.05in}
    
    \nonl\textbf{DP-SignOPORP-RR:}
    
    Compute $\tilde s_j=\begin{cases}
    sign(x_j), & \text{with prob.}\ \frac{e^{\epsilon}}{e^{\epsilon}+1}\\
    -sign(x_j), & \text{with prob.}\ \frac{1}{e^{\epsilon}+1}
    \end{cases}$ for $j=1,...,k$
    
    \nonl\textbf{DP-SignOPORP-RR-smooth:} 
    
    Compute $L_j=\lceil \frac{|x_j|}{\beta} \rceil$ for $j=1,...,k$

    Compute $\tilde s_j=\begin{cases}
    sign(x_j), & \text{with prob.}\ \frac{e^{\epsilon_j'}}{e^{\epsilon_j'}+1}\\
    -sign(x_j), & \text{with prob.}\ \frac{1}{e^{\epsilon_j'}+1}
    \end{cases}$ for $j=1,...,k$, with $\epsilon_j'=L_j\epsilon$
    
    For $\tilde s_j=0$, assign a random coin in $\{-1,1\}$
    
    Return $\tilde s$ as the DP-SignRP of $u$
    }
    \caption{DP-SignOPORP-RR and DP-SignOPORP-RR-smooth~\citep{li2023differential}}
    \label{alg:DP-signOPORP-RR}
\end{algorithm}

\begin{figure}[b!]
\centering

    \mbox{
    \includegraphics[width=2.7in]{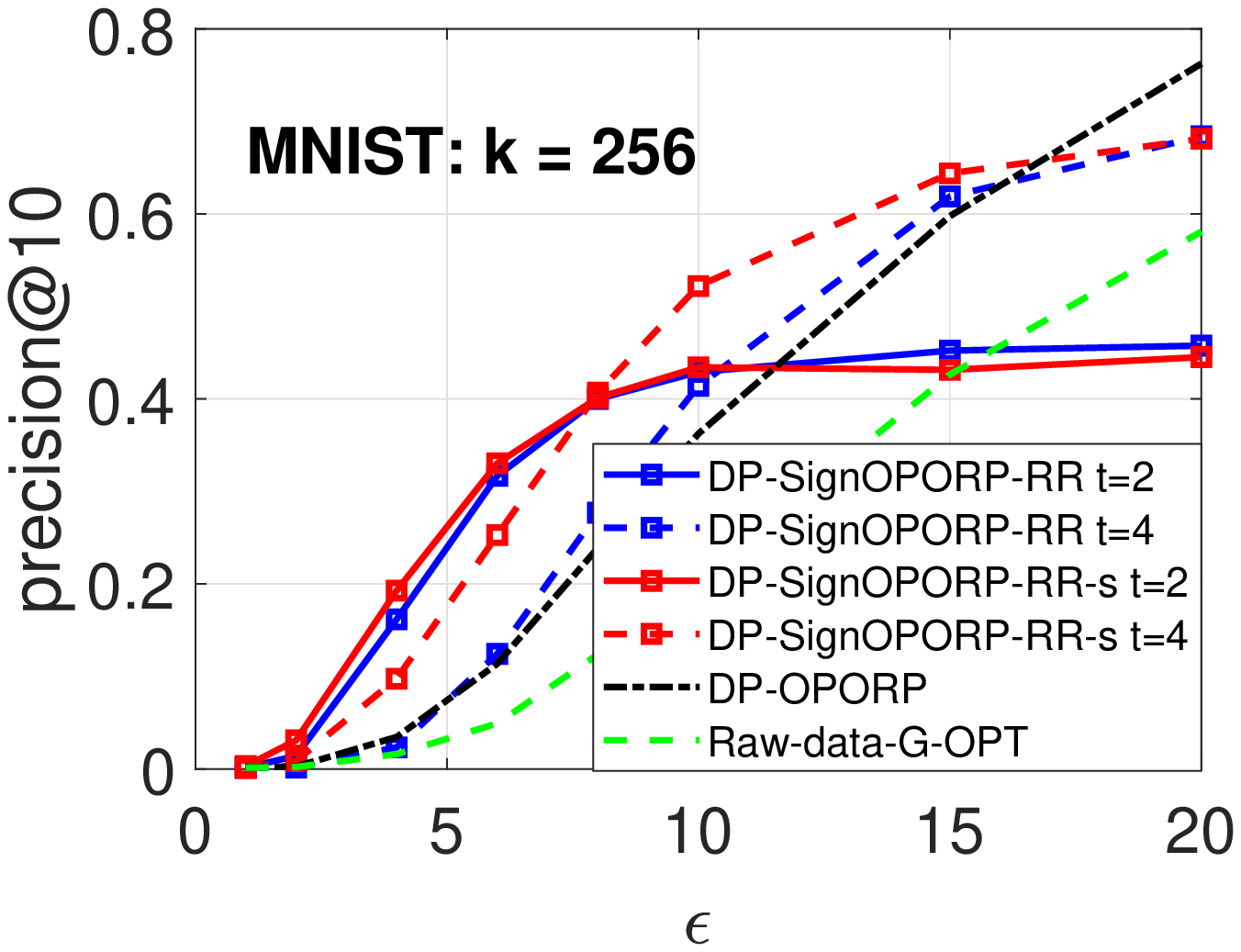} 
    \hspace{-0.15in}
    \includegraphics[width=2.7in]{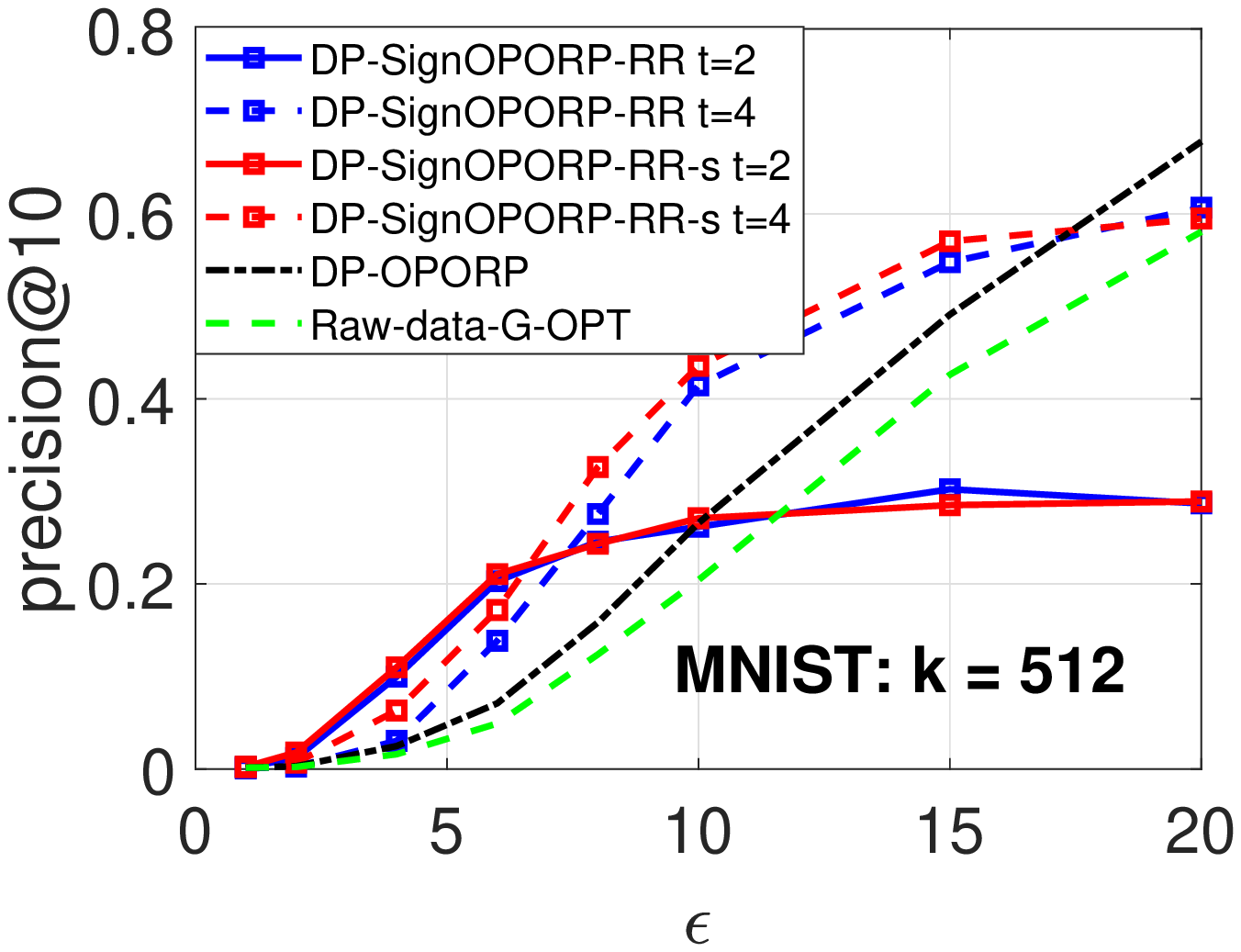}
    }

    \mbox{
    \includegraphics[width=2.7in]{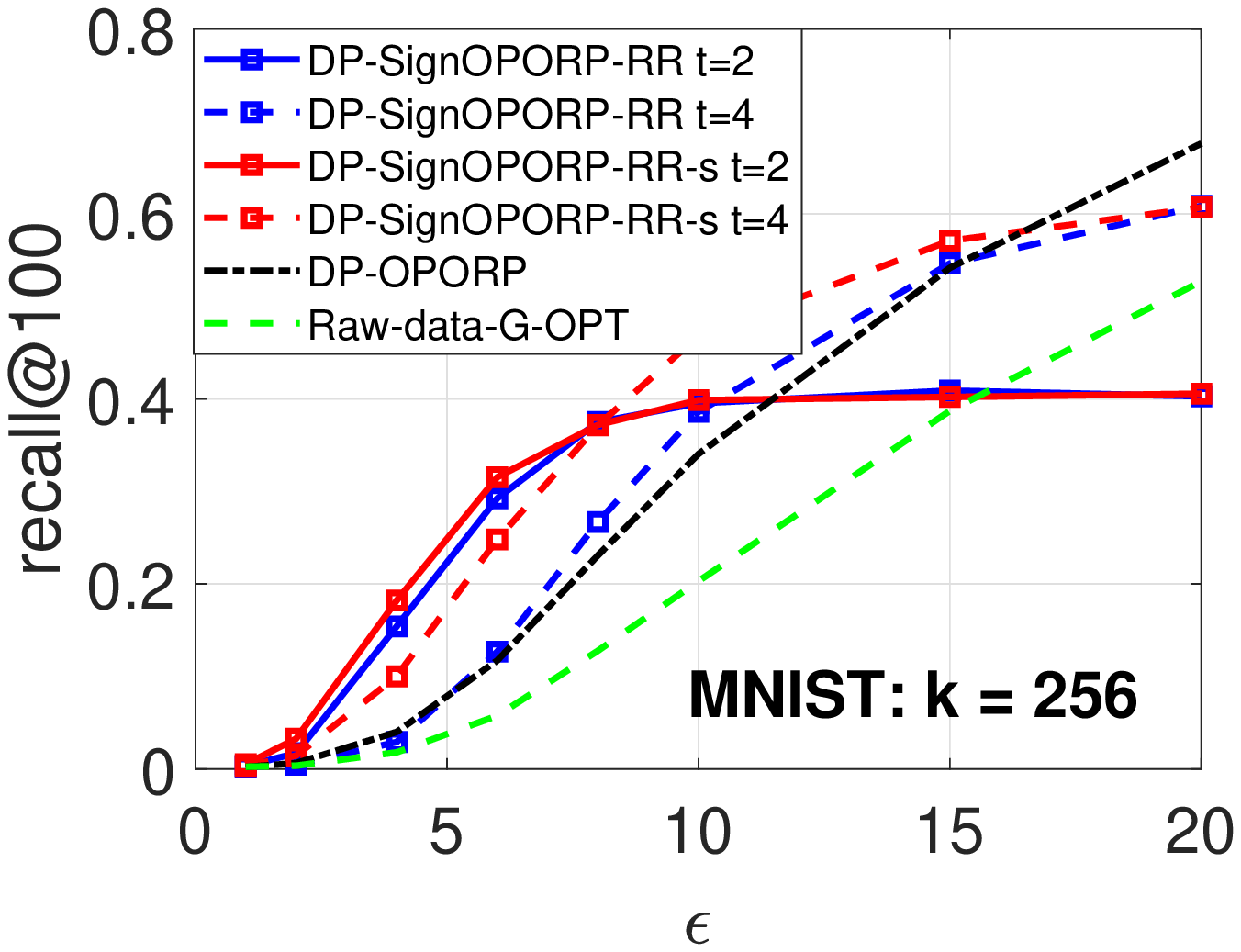} 
    \hspace{-0.15in}
    \includegraphics[width=2.7in]{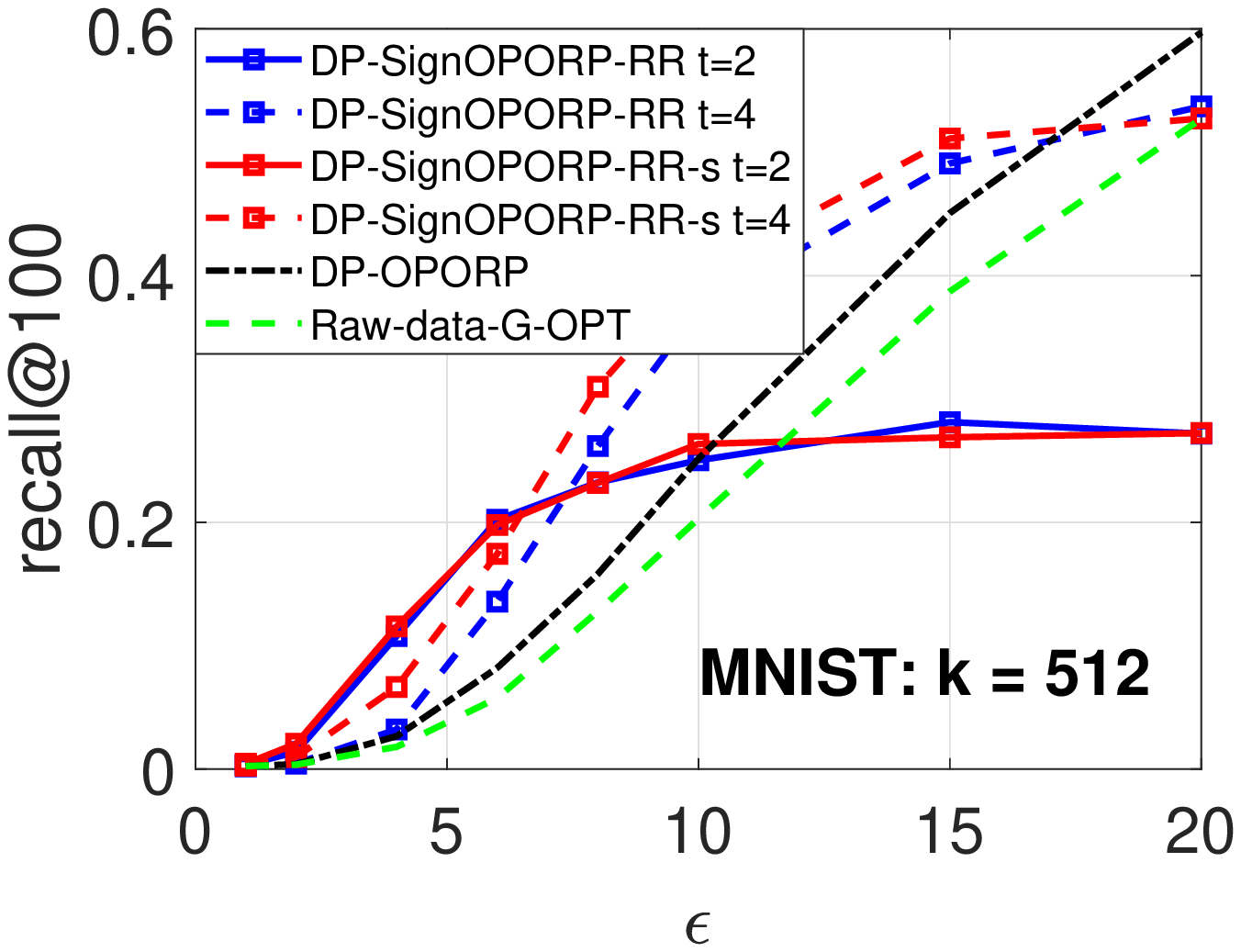}
    }

\vspace{-0.1in}

\caption{\citep{li2023differential} Retrieval on MNIST with DP-SignOPORP-RR and DP-SignOPORP-RR-smooth (in the caption, ``-s'' stands for ``-smooth''. }
\label{fig:MNIST_vs_eps_sign_oporp}
\end{figure}

\newpage\clearpage

\noindent\textbf{DP-SignOPORP.} \hspace{0.1in} The algorithm is summarized in Algorithm~\ref{alg:DP-signOPORP-RR}. We first generate the OPORP $x=OPORP(u)$, then takes the sign of $x$ to get a length-$k$ bit vector. Then, we have  two options. The first ``DP-SignOPORP-RR'' strategy is to apply the standard/classic randomized response (RR) technique to flip each bit with probability $\frac{1}{e^{\epsilon}+1}$ and keep the true sign otherwise. The second (improved) solution, called ``DP-SignOPORP-RR-smooth'' is based on the concept proposed by \citet{li2023differential} called ``smooth flipping probability'', inspired by the work of~\cite{nissim2007smooth}. That is, the flipping probability for each bit is $\frac{1}{e^{\epsilon_j'}+1}$ with $\epsilon_j'=L_j\epsilon$ and $L_j=\lceil \frac{|x_j|}{\beta} \rceil$. The intuitive understanding is that, when the OPORP value is farther from $0$, the flipping probability of the sign of OPORP can be smaller. This is a consequence of the fact that the ``aggregate-and-sign'' operation of RP-type algorithms brings robustness against small changes in the data. It is proved that both approaches are $\epsilon$-DP. The DP-SignOPORP-RR-smooth has smaller flipping probability than DP-SignOPORP-RR and thus better utility. \citet{li2023differential} also shown that applying multiple $t$ repetitions of DP-SignOPORP (each time with $k/t$ projections and privacy budget $\epsilon/t$) may help improve the performance of DP-SignOPORP. Figure~\ref{fig:MNIST_vs_eps_sign_oporp}  (also from~\citet{li2023differential}) confirms that DP-SignOPORP achieves good better performance than DP-OPORP especially when $\epsilon$ is not large.

\vspace{0.1in}
Finally, We should mention that~\citet{li2023differential} also developed algorithms (e.g., iDP-SignRP) for ``individual differential privacy'' (iDP)~\citep{comas2017individual} which are able to achieve remarkable utility performance even at very small $\epsilon$ (e.g., $\epsilon<0.5$).

\section{Conclusion}

Computing or estimating the inner products  (or cosines) is the routine operation in numerous applications, not limited to machine learning. Reducing the storage/memory cost and speeding up the computations for computing/estimating the inner products or cosines can be crucial especially in many industrial applications such as embedding-based retrieval (EBR) for search and advertising.

\vspace{0.1in}

\noindent The  ``one permutation + one random projection'' (OPORP)  is a variant of count-sketch and is closely related to ``very sparse random projections'' (VSRP). Compared to the standard random projections, OPORP is substantially more efficient (as it involves only one projection) and also more accurate. It differs from the  standard count-sketch in that OPORP utilizes \textbf{(i) the fixed-length binning} scheme; \textbf{(ii) the normalized estimator} of  cosine and inner product. We have conducted thorough variance analysis for OPORP (as well as VSRP) for both un-normalized and normalized estimators.

\vspace{0.1in}

\noindent Among many applications (e.g., AI model compression), this work can be used as a key component in modern ANN (approximate near neighbor search) systems. For example, \citet{zhao2020song} developed the GPU graph-based ANN algorithm and  used random projections to reduce  memory cost when  data do not in the memory. For large-scale graph-based ANN methods~\citep{zhou2019mobius,malkov2020efficient}, the main  cost is to compute similarities on the fly. We can effectively compress the  vectors using OPORP to facilitate the distance computations at reduced storage. 

\vspace{0.15in}

\noindent As elaborated in the paper, OPORP and VSRP (very sparse random projections)~\citep{li2006very} can be viewed as two extreme examples of sparse random projections. Our work on OPORP naturally recovers the estimator and theory of VSRP. In fact, as a by-product, we also develop the normalized estimator for VSRP and derive its variance. To compare VSRP with OPORP, the general conclusion is that both have the same variance (neglecting the beneficial $\frac{D-k}{D-1}$ factor for OPORP) if VSRP uses $s=1$ (i.e., a fully dense projection matrix); but VSRP can severely lose the accuracy if VSRP uses a large $s$ value in order to achieve the same level of sparsity as OPORP.

\newpage\clearpage 

\appendix

\section{Proof of Theorem~\ref{thm:a}}\label{proof:thm:a}
For two data vectors $u,v\in \mathbb R^D$, recall the notations of OPORP:
\begin{align}\notag
\hat{a} = \sum_{j=1}^k x_j y_j, \hspace{0.2in} x_j = \sum_{i=1}^D u_i r_i I_{ij}, \hspace{0.2in} y_j = \sum_{i=1}^D v_i r_i I_{ij}.
\end{align}
Assume the random variable $r$ admits
\begin{align}\notag
E(r_i) = 0,\hspace{0.3in}  E(r_i^2) = 1,\hspace{0.3in}  E(r_i^3) = 0,\hspace{0.3in}    E(r_i^4) = s.
\end{align}
Our goal is to show
\begin{align*}
&E(\hat{a}) = a, \\
&Var(\hat{a}_1)
=\left(s-1\right) \sum_{i=1}^D u_i^2 v_i^2 + \frac{1}{k}\left[a^2+ \sum_{i=1}^Du_i^2 \sum_{i=1}^Dv_i^2 -2\sum_{i=1}^D u_i^2v_i^2\right]\frac{D-k}{D-1}, \\
&Var(\hat{a}_2)
=\left(s-1\right) \sum_{i=1}^D u_i^2 v_i^2 +\frac{1}{k}\left[a^2+ \sum_{i=1}^Du_i^2 \sum_{i=1}^Dv_i^2 -2\sum_{i=1}^D u_i^2v_i^2\right].
\end{align*}
Firstly, for the mean, we have
\begin{align}\notag
E\left(\hat{a}\right) =& E\left(\sum_{j=1}^k x_j y_j \right) = 
E\left(\sum_{j=1}^k \sum_{i=1}^{D} u_i r_i I_{ij} \sum_{i=1}^{D} v_i r_i I_{ij} \right)\\\notag
=&
E\left(\sum_{j=1}^k \sum_{i=1}^{D} u_iv_i r_i^2 I_{ij}^2+ \sum_{i\neq i^\prime} u_iv_{i^\prime} r_ir_{i^\prime} I_{ij} I_{i^\prime j} \right)\\\notag
=&E\left(\sum_{j=1}^k \sum_{i=1}^{D} u_iv_i  \frac{1}{k}\right) + 0\\\notag
=&\sum_{i=1}^Du_iv_i = a,
\end{align}
which shows that $\hat a$ is an unbiased estimator of $a$. 

We can compute the second moment of $\hat a$ as
\begin{align}\notag
E\left(\hat{a}^2\right) =& E\left(\sum_{j=1}^k x_j y_j \right)^2 = 
E\left(\sum_{j=1}^k \sum_{i=1}^{D} u_i r_i I_{ij} \sum_{i=1}^{D} v_i r_i I_{ij} \right)^2\\\notag
=&E\left(\sum_{j=1}^k \sum_{i=1}^{D} u_iv_i r_i^2 I_{ij}^2+ \sum_{i\neq i^\prime} u_iv_{i^\prime} r_ir_{i^\prime} I_{ij} I_{i^\prime j} \right)^2\\\notag
=&E\left(\sum_{j=1}^k \left(\sum_{i=1}^{D} u_iv_i r_i^2 I_{ij}^2+ \sum_{i\neq i^\prime} u_iv_{i^\prime} r_ir_{i^\prime} I_{ij} I_{i^\prime j} \right)^2\right)\\\notag
+&E\left(\sum_{j\neq j^\prime} \left(\sum_{i=1}^{D} u_iv_i r_i^2 I_{ij}^2+ \sum_{i\neq i^\prime} u_iv_{i^\prime} r_ir_{i^\prime} I_{ij} I_{i^\prime j} \right)  \left(\sum_{i=1}^{D} u_iv_i r_i^2 I_{ij^\prime}^2+ \sum_{i\neq i^\prime} u_iv_{i^\prime} r_ir_{i^\prime} I_{ij^\prime} I_{i^\prime j^\prime} \right)\right)\\\notag
=&\sum_{j=1}^k E\left(\sum_{i=1}^{D} u_iv_i r_i^2 I_{ij}^2+ \sum_{i\neq i^\prime} u_iv_{i^\prime} r_ir_{i^\prime} I_{ij} I_{i^\prime j} \right)^2\\ \label{eqn:a1}
+&\sum_{j\neq j^\prime} E\left(\sum_{i=1}^{D} u_iv_i r_i^2 I_{ij}^2+ \sum_{i\neq i^\prime} u_iv_{i^\prime} r_ir_{i^\prime} I_{ij} I_{i^\prime j} \right)  \left(\sum_{i=1}^{D} u_iv_i r_i^2 I_{ij^\prime}^2+ \sum_{i\neq i^\prime} u_iv_{i^\prime} r_ir_{i^\prime} I_{ij^\prime} I_{i^\prime j^\prime} \right).
\end{align}
We now compute the two terms separately. For the first term, we have 
\begin{align}\notag
&E\left(\sum_{i=1}^{D} u_iv_i r_i^2 I_{ij}^2+ \sum_{i\neq i^\prime} u_iv_{i^\prime} r_ir_{i^\prime} I_{ij} I_{i^\prime j} \right)^2\\\notag
=&E\left(\sum_{i=1}^{D} u_iv_i r_i^2 I_{ij}^2\right)^2+ E\left(\sum_{i\neq i^\prime} u_iv_{i^\prime} r_ir_{i^\prime} I_{ij} I_{i^\prime j} \right)^2
+2E\left(\sum_{i=1}^{D} u_iv_i r_i^2 I_{ij}^2\right)\left(\sum_{i\neq i^\prime} u_iv_{i^\prime} r_ir_{i^\prime} I_{ij} I_{i^\prime j} \right)\\\notag
=&s\frac{1}{k}\sum_{i=1}^D u_i^2 v_i^2 + \sum_{i\neq i^\prime} u_iv_i   u_{i^\prime}v_{i^\prime} E\left( I_{ij} I_{i^\prime j}\right) + \sum_{i\neq i^\prime} u_i^2v_{i^\prime}^2 E( I_{ij} I_{i^\prime j})+ \sum_{i\neq i^\prime} u_iv_i   u_{i^\prime}v_{i^\prime} E\left( I_{ij} I_{i^\prime j}\right)\\\notag
=&s\frac{1}{k}\sum_{i=1}^D u_i^2 v_i^2 + \sum_{i\neq i^\prime} \left(u_i^2v_{i^\prime}^2+2u_iv_i   u_{i^\prime}v_{i^\prime} \right)E\left( I_{ij} I_{i^\prime j}\right) 
\end{align}
To see the above calculations, we can calculate the three terms respectively.
\begin{align}\notag
E\left(\sum_{i=1}^{D} u_iv_i r_i^2 I_{ij}^2\right)^2 =& E\left(\sum_{i=1}^{D} u_i^2v_i^2 r_i^4 I_{ij}^4\right) +  E\left(\sum_{i\neq i^\prime} u_iv_i r_i^2 I_{ij}^2 u_{i^\prime} v_{i^\prime} r_{i^\prime}^2 I_{i^\prime j}^2\right)\\\notag
=&s\frac{1}{k}\sum_{i=1}^D u_i^2 v_i^2 + \sum_{i\neq i^\prime} u_iv_i   u_{i^\prime} v_{i^\prime} E\left( I_{ij} I_{i^\prime j}\right),
\end{align}
\begin{align}\notag
&E\left(\sum_{i\neq i^\prime} u_iv_{i^\prime} r_ir_{i^\prime} I_{ij} I_{i^\prime j} \right)^2\\\notag
=& E\left(\sum_{i< i^\prime} u_iv_{i^\prime} r_ir_{i^\prime} I_{ij} I_{i^\prime j} + \sum_{i> i^\prime} u_iv_{i^\prime} r_ir_{i^\prime} I_{ij} I_{i^\prime j}\right)^2\\\notag
=&E\left(\sum_{i\neq i^\prime} u_i^2v_{i^\prime}^2r_i^2r_{i^\prime}^2 I_{ij}^2 I_{i^\prime j}^2 \right)+2E\left(\sum_{i< i^\prime} u_iv_{i^\prime} r_ir_{i^\prime} I_{ij} I_{i^\prime j}\right) \left(\sum_{i< i^\prime} u_{i^\prime}v_{i} r_ir_{i^\prime} I_{ij} I_{i^\prime j}\right)
\\\notag
=&\sum_{i\neq i^\prime} u_i^2v_{i^\prime}^2 E( I_{ij}I_{i^\prime j})
+2\sum_{i< i^\prime} u_iu_{i^\prime}v_iv_{i^\prime} E\left( I_{ij} I_{i^\prime j}\right)\\\notag
=&\sum_{i\neq i^\prime} u_i^2v_{i^\prime}^2 E( I_{ij}I_{i^\prime j})
+\sum_{i\neq i^\prime} u_iu_{i^\prime}v_iv_{i^\prime} E\left( I_{ij} I_{i^\prime j}\right),
\end{align}
and
\begin{align}\notag
E\left(\sum_{i=1}^{D} u_iv_i r_i^2 I_{ij}^2\right)\left(\sum_{i\neq i^\prime} u_iv_{i^\prime} r_ir_{i^\prime} I_{ij} I_{i^\prime j} \right) = 0.
\end{align}
In the calculations, we can simplify the algebra by noting that $r_i$'s are i.i.d. and $E(r_i)=E(r_i^3)=0$. Next, we compute 
\begin{align}\notag
&E\left(\sum_{i=1}^{D} u_iv_i r_i^2 I_{ij}^2+ \sum_{i\neq i^\prime} u_iv_{i^\prime} r_ir_{i^\prime} I_{ij} I_{i^\prime j} \right)  \left(\sum_{i=1}^{D} u_iv_i r_i^2 I_{ij^\prime}^2+ \sum_{i\neq i^\prime} u_iv_{i^\prime} r_ir_{i^\prime} I_{ij^\prime} I_{i^\prime j^\prime} \right)\\\notag
=&E\left(\sum_{i=1}^{D} u_iv_i r_i^2 I_{ij}^2\right)\left(\sum_{i=1}^{D} u_iv_i r_i^2 I_{ij^\prime}^2\right)  +E\left(\sum_{i\neq i^\prime} u_iv_{i^\prime} r_ir_{i^\prime} I_{ij} I_{i^\prime j} \right) 
\left(\sum_{i\neq i^\prime} u_iv_{i^\prime} r_ir_{i^\prime} I_{ij^\prime} I_{i^\prime j^\prime} \right)\\\notag
=&s\sum_{i=1}^{D} u_i^2v_i^2  E\left(I_{ij}I_{ij^\prime}\right) + \sum_{i\neq i^\prime} u_iv_iu_{i^\prime}v_{i^\prime}  E\left(I_{ij}I_{i^\prime j^\prime}\right)  +\sum_{i\neq i^\prime} u_i^2v_{i^\prime}^2 E\left( I_{ij} I_{i^\prime j}  I_{ij^\prime} I_{i^\prime j^\prime} \right) +\sum_{i\neq i^\prime} u_iu_{i^\prime}v_iv_{i^\prime} E\left( I_{ij} I_{i^\prime j}  I_{ij^\prime} I_{i^\prime j^\prime} \right)\\\notag
=& \sum_{i\neq i^\prime} u_iv_iu_{i^\prime}v_{i^\prime}  E\left(I_{ij}I_{i^\prime j^\prime}\right),
\end{align}
where we have used the fact that $I_{ij}I_{ij^\prime} = 0$ always. Now turning back to (\ref{eqn:a1}), we obtain
\begin{align}\notag
E(\hat{a}^2)=&\sum_{j=1}^k\left[s\frac{1}{k}\sum_{i=1}^D u_i^2 v_i^2 + \sum_{i\neq i^\prime} \left(u_i^2v_{i^\prime}^2+2 u_iv_i   u_{i^\prime} v_{i^\prime} \right)E\left( I_{ij} I_{i^\prime j}\right)\right]+\sum_{j\neq j^\prime}\left[ \sum_{i\neq i^\prime} u_iv_iu_{i^\prime}v_{i^\prime}  E\left(I_{ij}I_{i^\prime j^\prime}\right) \right]\\\notag
=&s \sum_{i=1}^D u_i^2 v_i^2 + kE\left( I_{ij} I_{i^\prime j}\right) \sum_{i\neq i^\prime} \left(u_i^2v_{i^\prime}^2+2u_iv_i   u_{i^\prime} v_{i^\prime} \right)
+k(k-1)E\left(I_{ij}I_{i^\prime j^\prime}\right)\sum_{i\neq i^\prime} u_iv_iu_{i^\prime}v_{i^\prime}. 
\end{align}
Therefore, the variance can be expressed as
\begin{align}\notag
&Var(\hat{a})= E(\hat{a}^2) - a^2 \\\notag
=&s \sum_{i=1}^D u_i^2 v_i^2 + kE\left( I_{ij} I_{i^\prime j}\right) \sum_{i\neq i^\prime} \left(u_i^2v_{i^\prime}^2+2u_iv_i   u_{i^\prime} v_{i^\prime} \right)
+k(k-1)E\left(I_{ij}I_{i^\prime j^\prime}\right)\sum_{i\neq i^\prime} u_iv_iu_{i^\prime}v_{i^\prime}  - \left(\sum_{i=1}^D u_iv_i\right)^2\\\notag
=&\left(s-1\right) \sum_{i=1}^D u_i^2 v_i^2 + kE\left( I_{ij} I_{i^\prime j}\right) \sum_{i\neq i^\prime} u_i^2v_{i^\prime}^2+\left[2kE\left( I_{ij} I_{i^\prime j}\right) + k(k-1)E\left(I_{ij}I_{i^\prime j^\prime}\right)-1\right]\sum_{i\neq i^\prime} u_iv_iu_{i^\prime}v_{i^\prime}  \\\notag
=&\left(s-1\right) \sum_{i=1}^D u_i^2 v_i^2 + kE\left( I_{ij} I_{i^\prime j}\right) \sum_{i\neq i^\prime} u_i^2v_{i^\prime}^2+kE\left( I_{ij} I_{i^\prime j}\right) \sum_{i\neq i^\prime} u_iv_iu_{i^\prime}v_{i^\prime}  \\\notag
=&\left(s-1\right) \sum_{i=1}^D u_i^2 v_i^2 + kE\left( I_{ij} I_{i^\prime j}\right) \left[\sum_{i=1}^Du_i^2 \sum_{i=1}^Dv_i^2 - \sum_{i=1}^D u_i^2v_i^2+ \left(\sum_{i=1}^D u_iv_i\right)^2 -\sum_{i=1}^D u_i^2v_i^2
\right]\\\notag
=&\left(s-1\right) \sum_{i=1}^D u_i^2 v_i^2 + kE\left( I_{ij} I_{i^\prime j}\right) \left[a^2+\sum_{i=1}^Du_i^2 \sum_{i=1}^Dv_i^2  -2\sum_{i=1}^D u_i^2v_i^2
\right].
\end{align}
The remaining part is to compute $E(I_{ij}I_{i'j})$ for the two binning schemes respectively, which is finished by leveraging Lemma~\ref{lem:I}.  $\hfill\qed$

\newpage

\section{Proof of Theorem~\ref{thm:rho}}\label{proof:thm:rho}

Recall the notation in OPORP:
\begin{align}\notag
    x_j = \sum_{i=1}^{D} u_i r_i I_{ij},\hspace{0.3in}
    y_j = \sum_{i=1}^{D} v_i r_i I_{ij}, \hspace{0.2in} j = 1, 2,  ..., k.
\end{align}
To analyze the normalized cosine estimator: 
\begin{align}\notag
\hat{\rho} = \frac{\sum_{j=1}^k x_j y_j}{\sqrt{\sum_{j=1}^k x_j^2}{\sqrt{\sum_{j=1}^k y_j^2}} },
\end{align}
it suffices to assume  the original data are normalized to unit $l_2$ norms, i.e., $\sum_{i=1}^D u_i^2 = \sum_{i=1}^Dv_i^2=1$. When the data are normalized, the inner product and the cosine are the same, i.e., $a=\rho$. Thus,
\begin{align}\notag
E\left(\sum_{j=1}^k x_j y_j\right) = \rho, \hspace{0.2in} E\left(\sum_{j=1}^k x_j^2\right) = E\left(\sum_{j=1}^k y_j^2\right) = 1, \hspace{0.2in}
\end{align}
Via the Taylor expansion, we have
\begin{align}\notag
\hat{\rho}-\rho =& \frac{\sum_{j=1}^kx_jy_j -\rho}{\sqrt{\sum_{j=1}^k x_j^2}{\sqrt{\sum_{j=1}^k y_j^2}}}+\rho\frac{1-\sqrt{\sum_{j=1}^k x_j^2}{\sqrt{\sum_{j=1}^k y_j^2}}}{\sqrt{\sum_{j=1}^k x_j^2}{\sqrt{\sum_{j=1}^k y_j^2}}} \\ \notag
=& \sum_{j=1}^kx_jy_j -\rho +\rho\left(1-\sum_{j=1}^kx_j^2\right)/2+\rho\left(1-\sum_{j=1}^ky_j^2\right)/2+O_P\left(1/k\right)\\\notag
=&\sum_{j=1}^kx_jy_j -\rho/2\sum_{j=1}^kx_j^2- \rho/2\sum_{j=1}^ky_j^2+O_P\left(1/k\right),
\end{align}
where we use the approximation: for $a\approx 1$ and $b\approx 1$, $1-ab=(1-a)+(1-b)-(1-a)(1-b)$. It thus suffices to analyze the following term:
\begin{align}\notag
&\left(\sum_{j=1}^kx_jy_j -\rho/2\sum_{j=1}^kx_j^2- \rho/2\sum_{j=1}^ky_j^2\right)^2\\
=&\left(\sum_{j=1}^kx_jy_j\right)^2+\rho^2/4\left(\sum_{j=1}^kx_j^2+\sum_{j=1}^ky_j^2\right)^2-\rho\left(\sum_{j=1}^kx_jy_j\right)\left(\sum_{j=1}^kx_j^2+\sum_{j=1}^ky_j^2\right). \label{eqn:rho1}
\end{align}
By Theorem~\ref{thm:a}, we know that
\begin{align}\label{eqn:rho2}
&E\left(\sum_{j=1}^kx_jy_j\right)^2
=\left(s-1\right) \sum_{i=1}^D u_i^2 v_i^2 + kE\left( I_{ij} I_{i^\prime j}\right)\left[ 1+\rho^2 -2\sum_{i=1}^D u_i^2v_i^2\right]  + \rho^2,
\end{align}
and we can write
\begin{align}\notag
&E\left(\sum_{j=1}^kx_j^2 \sum_{j=1}^ky_j^2\right) = E\left(\sum_{j=1}^kx_j^2y_j^2 +  \sum_{j\neq j'}^kx_j^2y_{j^\prime}^2\right).
\end{align}
We now calculate each term. First, we have for $j=1,...,k$,
\begin{align} \notag
&E\left(x_j^2 y_j^2 \right) 
=s\frac{1}{k}\sum_{i=1}^D u_i^2 v_i^2 + E\left( I_{ij} I_{i^\prime j}\right)\sum_{i\neq i^\prime} \left(u_i^2v_{i^\prime}^2+2u_iv_i   u_{i^\prime}v_{i^\prime} \right).
\end{align}
Also, for $j\neq j'$,
\begin{align}\notag
&E(x_j^2y_{j^\prime}^2) = E\left(\sum_{i=1}^Du_ir_iI_{ij}\right)^2\left(\sum_{i=1}^Dv_ir_iI_{ij^\prime}\right)^2\\\notag
=&E\left(\sum_{i=1}^Du_i^2r_i^2I_{ij}+\sum_{i\neq i^\prime} u_iu_{i^\prime}r_ir_{i^\prime}r_{i^\prime j}I_{ij}I_{i^\prime j}\right)
\left(\sum_{i=1}^Dv_i^2r_i^2I_{ij^\prime}+\sum_{i\neq i^\prime} v_iv_{i^\prime}r_ir_{i^\prime}r_{i^\prime j}I_{ij^\prime}I_{i^\prime j^\prime}\right)\\\notag
=&E\left(\sum_{i=1}^Du_i^2r_i^2I_{ij}\right)\left(\sum_{i=1}^Dv_i^2r_i^2I_{ij^\prime}\right)+E\left(\sum_{i\neq i^\prime} u_iu_{i^\prime}r_ir_{i^\prime}r_{i^\prime j}I_{ij}I_{i^\prime j}\right)\left(\sum_{i\neq i^\prime} v_iv_{i^\prime}r_ir_{i^\prime}r_{i^\prime j}I_{ij^\prime}I_{i^\prime j^\prime}\right)\\\notag
=& E\left(I_{ij}I_{i^\prime j^\prime}\right) \sum_{i\neq i^\prime} u_i^2v_{i^\prime}^2.
\end{align}
Therefore, we obtain that

\begin{align}\notag
&E\left(\sum_{j=1}^kx_j^2 \sum_{j=1}^ky_j^2\right) = E\left(\sum_{j=1}^kx_j^2y_j^2 +  \sum_{j\neq j}^kx_j^2y_{j^\prime}^2\right) \\\notag
=&s\sum_{i=1}^D u_i^2 v_i^2 + kE\left( I_{ij} I_{i^\prime j}\right)  \sum_{i\neq i^\prime} \left(u_i^2v_{i^\prime}^2+2u_iv_i   u_{i^\prime}v_{i^\prime} \right) + k(k-1)E\left(I_{ij}I_{i^\prime j^\prime}\right) \sum_{i\neq i^\prime} u_i^2v_{i^\prime}^2, 
\end{align}
which leads to
\begin{align}\notag
&E\left(\sum_{j=1}^kx_j^2+\sum_{j=1}^ky_j^2\right)^2
=\left(s-1\right) \sum_{i=1}^D (u_i^4+v_i^4) + 2kE\left( I_{ij} I_{i^\prime j}\right)\left[ 2 -\sum_{i=1}^D (u_i^4+v_i^4)\right]  + 2\\
&\hspace{0.2in}+2s\sum_{i=1}^D u_i^2 v_i^2 + 2kE\left( I_{ij} I_{i^\prime j}\right)  \sum_{i\neq i^\prime} \left(u_i^2v_{i^\prime}^2+2u_iv_i   u_{i^\prime}v_{i^\prime} \right) + 2k(k-1)E\left(I_{ij}I_{i^\prime j^\prime}\right) \sum_{i\neq i^\prime} u_i^2v_{i^\prime}^2. \label{eqn:rho3}
\end{align}
We now analyze the third term in (\ref{eqn:rho1}). It holds that
\begin{align}\notag
E\left(\sum_{j=1}^kx_jy_j\right)\left(\sum_{j=1}^kx_j^2+\sum_{j=1}^ky_j^2\right) = E\left(\sum_{j=1}^k x_j^3y_j + \sum_{j\neq j}x_jy_jx_{j^\prime}^2\right)+E\left(\sum_{j=1}^k x_jy_j^3 + \sum_{j\neq j}x_jy_jy_{j^\prime}^2\right).
\end{align}
We have
\begin{align}\notag
&E(x_jy_j^3) = E\left(\sum_{i=1}^Du_ir_iI_{ij}\right)\left(\sum_{i=1}^Dv_ir_iI_{ij}\right)^3\\\notag
=&E\left(\sum_{i=1}^D u_iv_ir_i^2I_{ij} + \sum_{i\neq i^\prime} u_iv_{i^\prime}r_ir_{i^\prime}I_{ij}I_{i^\prime j}\right)
\left(\sum_{i=1}^D v_i^2r_i^2I_{ij} + \sum_{i\neq i^\prime} v_iv_{i^\prime}r_ir_{i^\prime}I_{ij}I_{i^\prime j}\right)\\\notag
=&E\left(\sum_{i=1}^D u_iv_ir_i^2I_{ij} \right)\left(\sum_{i=1}^D v_i^2r_i^2I_{ij}\right) + E\left(\sum_{i\neq i^\prime} u_iv_{i^\prime}r_ir_{i^\prime}I_{ij}I_{i^\prime j}\right)
\left(\sum_{i\neq i^\prime} v_iv_{i^\prime}r_ir_{i^\prime}I_{ij}I_{i^\prime j}\right)\\\notag
=&E\left(\sum_{i=1}^D u_iv_i^3r_i^4I_{ij} +\sum_{i\neq i^\prime}u_iv_iv_{i^\prime}^2r_i^2r_{i^\prime}^2I_{ij}I_{i^\prime j}\right)+ E\left(\sum_{i\neq i^\prime} u_iv_{i^\prime}r_ir_{i^\prime}I_{ij}I_{i^\prime j}\right)
\left(\sum_{i\neq i^\prime} v_iv_{i^\prime}r_ir_{i^\prime}I_{ij}I_{i^\prime j}\right)\\\notag
=&\frac{s}{k}\sum_{i=1}^D u_iv_i^3 + \sum_{i\neq i^\prime}u_iv_iv_{i^\prime}^2E(I_{ij}I_{i^\prime j})+ 2\sum_{i\neq i^\prime} u_iv_iv_{i^\prime}^2E(I_{ij}I_{i^\prime j})\\\notag
=&\frac{s}{k}\sum_{i=1}^D u_iv_i^ 3 + 3E(I_{ij}I_{i^\prime j})\sum_{i\neq i^\prime}u_iv_iv_{i^\prime}^2,
\end{align}
where we use the following computation:
\begin{align}\notag
&E\left(\sum_{i< i^\prime} u_iv_{i^\prime}r_ir_{i^\prime}I_{ij}I_{i^\prime j}
+\sum_{i> i^\prime} u_iv_{i^\prime}r_ir_{i^\prime}I_{ij}I_{i^\prime j}
\right)
\left(\sum_{i< i^\prime} v_iv_{i^\prime}r_ir_{i^\prime}I_{ij}I_{i^\prime j}+\sum_{i> i^\prime} v_iv_{i^\prime}r_ir_{i^\prime}I_{ij}I_{i^\prime j}\right)\\\notag
=&\sum_{i\neq i^\prime} u_iv_iv_{i^\prime}^2E(I_{ij}I_{i^\prime j})
+E\left(\sum_{i< i^\prime} u_iv_{i^\prime}r_ir_{i^\prime}I_{ij}I_{i^\prime j}\right)
\left(\sum_{i> i^\prime} v_iv_{i^\prime}r_ir_{i^\prime}I_{ij}I_{i^\prime j}\right)\\\notag
&\hspace{1.5in} +E\left(\sum_{i> i^\prime} u_iv_{i^\prime}r_ir_{i^\prime}I_{ij}I_{i^\prime j}
\right)\left(\sum_{i< i^\prime} v_iv_{i^\prime}r_ir_{i^\prime}I_{ij}I_{i^\prime j}\right)\\\notag
=&\sum_{i\neq i^\prime} u_iv_iv_{i^\prime}^2E(I_{ij}I_{i^\prime j})
+E\left(\sum_{i< i^\prime} u_iv_{i^\prime}r_ir_{i^\prime}I_{ij}I_{i^\prime j}\right)
\left(\sum_{i< i^\prime} v_iv_{i^\prime}r_ir_{i^\prime}I_{ij}I_{i^\prime j}\right)\\\notag
&\hspace{1.5in} +E\left(\sum_{i< i^\prime} u_{i^\prime}v_{i}r_ir_{i^\prime}I_{ij}I_{i^\prime j}
\right)\left(\sum_{i< i^\prime} v_iv_{i^\prime}r_ir_{i^\prime}I_{ij}I_{i^\prime j}\right)\\\notag
=&\sum_{i\neq i^\prime} u_iv_iv_{i^\prime}^2E(I_{ij}I_{i^\prime j})
+\sum_{i< i^\prime} u_iv_iv_{i^\prime}^2E(I_{ij}I_{i^\prime j})
+\sum_{i< i^\prime} u_{i^\prime}v_{i}^2v_{i^\prime}E(I_{ij}I_{i^\prime j})\\\notag
=&2\sum_{i\neq i^\prime} u_iv_iv_{i^\prime}^2E(I_{ij}I_{i^\prime j}).
\end{align}
Furthermore, we have
\begin{align}\notag
&E\left(x_jy_jx_{j^\prime}^2\right)
=E\left(\sum_{i=1}^Du_ir_iI_{ij}\right)\left(\sum_{i=1}^Dv_ir_iI_{ij}\right)\left(\sum_{i=1}^Du_ir_iI_{ij^\prime}\right)^2\\\notag
=&E\left(\sum_{i=1}^D u_iv_ir_i^2I_{ij}+\sum_{i\neq i^\prime}u_iv_{i^\prime}r_ir_{i^\prime}I_{ij}I_{i^\prime j}\right)
\left(\sum_{i=1}^Du_i^2r_i^2I_{ij^\prime} + \sum_{i\neq i}u_iu_{i^\prime}r_ir_{i^\prime}I_{ij^\prime}I_{i^\prime j^\prime}\right)\\\notag
=&E(I_{ij}I_{i^\prime j^\prime}) \sum_{i\neq i}u_iu_{i^\prime}^2v_i.
\end{align}
Thus, by symmetry we have
\begin{align}\notag
&E\left(\sum_{j=1}^kx_jy_j\right)\left(\sum_{j=1}^kx_j^2+\sum_{j=1}^ky_j^2\right) = E\left(\sum_{j=1}^k x_j^3y_j + \sum_{j\neq j}x_jy_jx_{j^\prime}^2\right)+E\left(\sum_{j=1}^k x_jy_j^3 + \sum_{j\neq j}x_jy_jy_{j^\prime}^2\right)\\
=&s\sum_{i=1}^D\left( u_iv_i^3+u_i^3v_i\right) + 3kE(I_{ij}I_{i^\prime j})\sum_{i\neq i^\prime}\left(u_iv_iv_{i^\prime}^2+u_iv_iu_{i^\prime}^2\right)+k(k-1)E(I_{ij}I_{i^\prime j^\prime})\sum_{i\neq i}\left(u_iu_{i^\prime}^2v_i+u_iv_{i^\prime}^2v_i\right). \label{eqn:rho4}
\end{align}

Now we combine (\ref{eqn:rho2}), (\ref{eqn:rho3}) and (\ref{eqn:rho4}) with (\ref{eqn:rho1}) to obtain
\begin{align}\notag
&\left(\sum_{j=1}^kx_jy_j -\rho/2\sum_{j=1}^kx_j^2- \rho/2\sum_{j=1}^ky_j^2\right)^2\\\notag
=&(s-1)\sum_{i=1}^D\left((1+\rho^2/2)u_i^2v_i^2+\rho^2u_i^4/4+\rho^2v_i^4/4-\rho u_iv_i^3 -\rho u_i^3v_i\right)\\\notag
&+ kE\left( I_{ij} I_{i^\prime j}\right)\left[ 1+\rho^2 -2\sum_{i=1}^D u_i^2v_i^2\right]  +  \rho^2kE\left( I_{ij} I_{i^\prime j}\right)\left[ 1 -\sum_{i=1}^D (u_i^4+v_i^4)/2\right]  \\\notag
&+\rho^2\left[3+\sum_{i=1}^D u_i^2 v_i^2 + kE\left( I_{ij} I_{i^\prime j}\right)  \sum_{i\neq i^\prime} \left(u_i^2v_{i^\prime}^2+2u_iv_i   u_{i^\prime}v_{i^\prime} \right) + k(k-1)E\left(I_{ij}I_{i^\prime j^\prime}\right) \sum_{i\neq i^\prime} u_i^2v_{i^\prime}^2 \right]/2\\\notag
&-\rho\left[\sum_{i=1}^D\left( u_iv_i^3+u_i^3v_i\right) + 3kE(I_{ij}I_{i^\prime j})\sum_{i\neq i^\prime}\left(u_iv_iv_{i^\prime}^2+u_iv_iu_{i^\prime}^2\right)+k(k-1)E(I_{ij}I_{i^\prime j^\prime})\sum_{i\neq i}\left(u_iu_{i^\prime}^2v_i+u_iv_{i^\prime}^2v_i\right)\right]\\\notag
=&(s-1)\sum_{i=1}^D\left((1+\rho^2/2)u_i^2v_i^2+\rho^2u_i^4/4+\rho^2v_i^4/4-\rho u_iv_i^3 -\rho u_i^3v_i\right)\\\notag
&+ kE\left( I_{ij} I_{i^\prime j}\right)\left[ 1+\rho^2 -2\sum_{i=1}^D u_i^2v_i^2\right]  +  \rho^2kE\left( I_{ij} I_{i^\prime j}\right)\left[ 1 -\sum_{i=1}^D (u_i^4+v_i^4)/2\right]  \\\notag
&+\rho^2\left[4+ 2kE\left( I_{ij} I_{i^\prime j}\right)\left(\rho^2-\sum_{i=1}^Du_i^2v_i^2\right) \right]/2-\rho\left[2\rho+ 2kE\left( I_{ij} I_{i^\prime j}\right)\left(2\rho -\sum_{i=1}^D\left(u_i^3v_i+u_iv_i^3\right)\right)\right]\\\notag
=&(s-1)\sum_{i=1}^D\left((1+\rho^2/2)u_i^2v_i^2+\rho^2u_i^4/4+\rho^2v_i^4/4-\rho u_iv_i^3 -\rho u_i^3v_i\right)\\\notag
&+ kE\left( I_{ij} I_{i^\prime j}\right)\left[ 1+\rho^2 -2\sum_{i=1}^D u_i^2v_i^2+ \rho^2 -\rho^2/2\sum_{i=1}^D (u_i^4+v_i^4) + \rho^4-\rho^2\sum_{i=1}^Du_i^2v_i^2-4\rho^2 +2\rho\sum_{i=1}^D\left(u_i^3v_i+u_iv_i^3\right)\right]\\\notag
=&(s-1)\sum_{i=1}^D\left((1+\rho^2/2)u_i^2v_i^2+\rho^2u_i^4/4+\rho^2v_i^4/4-\rho u_iv_i^3 -\rho u_i^3v_i\right)\\\notag
&+ kE\left( I_{ij} I_{i^\prime j}\right)\left[ (1-\rho^2)^2 -2\sum_{i=1}^D u_i^2v_i^2-\rho^2/2\sum_{i=1}^D (u_i^4+v_i^4) -\rho^2\sum_{i=1}^Du_i^2v_i^2 +2\rho\sum_{i=1}^D\left(u_i^3v_i+u_iv_i^3\right)\right]\\\notag
:=&(s-1)A+ kE\left( I_{ij} I_{i^\prime j}\right)\left[(1-\rho^2)^2-2A  \right], 
\end{align}
which gives the general expression of the variance term in Theorem~\ref{thm:rho}. Applying Lemma~\ref{lem:I} leads to the variance formula for the two binning schemes respectively. In the above calculation, we use the following facts:
\begin{align}\notag
\sum_{i\neq i^\prime} u_i^2 v_{i^\prime}^2 = \sum_{i=1}^Du_i^2\sum_{i=1}^Dv_i^2 - \sum_{i=1}^Du_i^2v_i^2 = 1-\sum_{i=1}^Du_i^2v_i^2,
\end{align}
\begin{align}\notag
\sum_{i\neq i^\prime} u_iv_i u_{i^\prime}v_{i^\prime} = \left(\sum_{i=1}^Du_iv_i\right)^2 - \sum_{i=1}^Du_i^2v_i^2 = \rho^2-\sum_{i=1}^Du_i^2v_i^2,
\end{align}
\begin{align}\notag
\sum_{i\neq i^\prime} u_iv_i v_{i^\prime}^2 = \sum_{i=1}^Du_iv_i \sum_{i=1}^Dv_i^2- \sum_{i=1}^Du_iv_i^3 = \rho-\sum_{i=1}^Du_iv_i^3,
\end{align}
\begin{align}\notag
\sum_{i\neq i^\prime} u_iv_i u_{i^\prime}^2 = \rho-\sum_{i=1}^Du_i^3v_i,
\end{align}
and
\begin{align}\notag
&kE\left( I_{ij} I_{i^\prime j}\right)  \sum_{i\neq i^\prime} \left(u_i^2v_{i^\prime}^2+2u_iv_i   u_{i^\prime}v_{i^\prime} \right) + k(k-1)E\left(I_{ij}I_{i^\prime j^\prime}\right) \sum_{i\neq i^\prime} u_i^2v_{i^\prime}^2 \\\notag
=&\left(kE\left( I_{ij} I_{i^\prime j}\right) + k(k-1)E\left(I_{ij}I_{i^\prime j^\prime}\right)\right)\left(1-\sum_{i=1}^Du_i^2v_i^2\right)+ 2kE\left( I_{ij} I_{i^\prime j}\right)\left(\rho^2-\sum_{i=1}^Du_i^2v_i^2\right)\\\notag
=&1-\sum_{i=1}^Du_i^2v_i^2+ 2kE\left( I_{ij} I_{i^\prime j}\right)\left(\rho^2-\sum_{i=1}^Du_i^2v_i^2\right),
\end{align}
\begin{align}\notag
&3kE(I_{ij}I_{i^\prime j})\sum_{i\neq i^\prime}\left(u_iv_iv_{i^\prime}^2+u_iv_iu_{i^\prime}^2\right)+k(k-1)E(I_{ij}I_{i^\prime j^\prime})\sum_{i\neq i}\left(u_iu_{i^\prime}^2v_i+u_iv_{i^\prime}^2v_i\right)\\\notag
=&\left(1+2kE\left( I_{ij} I_{i^\prime j}\right)\right)\left(2\rho -\sum_{i=1}^D\left(u_i^3v_i+u_iv_i^3\right)\right),
\end{align}
where by Lemma~\ref{lem:I} we have
\begin{align}\notag
kE\left( I_{ij} I_{i^\prime j}\right) + k(k-1)E\left(I_{ij}I_{i^\prime j^\prime}\right) = 1.
\end{align}


Lastly, we may simplify the expression of $A$ as

\begin{align}\notag
A=&\sum_{i=1}^D u_i^2v_i^2+\rho^2/4\sum_{i=1}^D (u_i^4+v_i^4) +\rho^2/2\sum_{i=1}^Du_i^2v_i^2 -\rho\sum_{i=1}^D\left(u_i^3v_i+u_iv_i^3\right)\\\notag
=&\sum_{i=1}^D u_i^2v_i^2+\rho^2/4\sum_{i=1}^D (u_i^2+v_i^2)^2  -\rho\sum_{i=1}^D\left(u_i^3v_i+u_iv_i^3\right)\\\notag
=&\sum_{i=1}^D\left(u_iv_i -\rho/2(u_i^2+v_i^2)\right)^2 + \rho(u_iv_i)(u_i^2+v_i^2) -\rho\left(u_i^3v_i+u_iv_i^3\right) \\
=&\sum_{i=1}^D\left(u_iv_i -\rho/2(u_i^2+v_i^2)\right)^2. \notag
\end{align}
This essentially completes the proof of Theorem~\ref{thm:rho}, by assuming normalized data. For un-normalized data, we need to replace $u_i$ and $v_i$ by $u_i^\prime = \frac{u_i}{\sqrt{\sum_{t=1}^D u_t^2}}$,  and 
$v_i^\prime = \frac{v_i}{\sqrt{\sum_{t=1}^D v_t^2}}$, respectively. 

\newpage

\bibliographystyle{plainnat}
\bibliography{refs_scholar}

\begin{thebibliography}{76}
\providecommand{\natexlab}[1]{#1}
\providecommand{\url}[1]{\texttt{#1}}
\expandafter\ifx\csname urlstyle\endcsname\relax
  \providecommand{\doi}[1]{doi: #1}\else
  \providecommand{\doi}{doi: \begingroup \urlstyle{rm}\Url}\fi

\bibitem[Achlioptas(2003)]{achlioptas2003database}
Dimitris Achlioptas.
\newblock Database-friendly random projections: Johnson-lindenstrauss with
  binary coins.
\newblock \emph{J. Comput. Syst. Sci.}, 66\penalty0 (4):\penalty0 671--687,
  2003.

\bibitem[AlOmar et~al.(2021)AlOmar, Aljedaani, Tamjeed, Mkaouer, and
  El{-}Glaly]{abdullah2021finding}
Eman~Abdullah AlOmar, Wajdi Aljedaani, Murtaza Tamjeed, Mohamed~Wiem Mkaouer,
  and Yasmine~N. El{-}Glaly.
\newblock Finding the needle in a haystack: On the automatic identification of
  accessibility user reviews.
\newblock In \emph{Proceedings of the Conference on Human Factors in Computing
  Systems (CHI)}, pages 387:1--387:15, Virtual Event / Yokohama, Japan, 2021.

\bibitem[Anderson(2003)]{anderson2003introduction}
Theodore~W. Anderson.
\newblock \emph{An Introduction to Multivariate Statistical Analysis}.
\newblock John Wiley \& Sons, third edition, 2003.

\bibitem[Bachrach et~al.(2014)Bachrach, Finkelstein, Gilad{-}Bachrach, Katzir,
  Koenigstein, Nice, and Paquet]{bachrach2014speeding}
Yoram Bachrach, Yehuda Finkelstein, Ran Gilad{-}Bachrach, Liran Katzir, Noam
  Koenigstein, Nir Nice, and Ulrich Paquet.
\newblock Speeding up the {Xbox} recommender system using a euclidean
  transformation for inner-product spaces.
\newblock In \emph{Proceedings of the Eighth {ACM} Conference on Recommender
  Systems (RecSys)}, pages 257--264, Foster City, CA, 2014.

\bibitem[Balle and Wang(2018)]{balle2018improving}
Borja Balle and Yu{-}Xiang Wang.
\newblock Improving the gaussian mechanism for differential privacy: Analytical
  calibration and optimal denoising.
\newblock In \emph{Proceedings of the 35th International Conference on Machine
  Learning (ICML)}, pages 403--412, Stockholmsm{\"{a}}ssan, Stockholm, Sweden,
  2018.

\bibitem[Bingham and Mannila(2001)]{bingham2001random}
Ella Bingham and Heikki Mannila.
\newblock Random projection in dimensionality reduction: Applications to image
  and text data.
\newblock In \emph{Proceedings of the Seventh {ACM} {SIGKDD} International
  Conference on Knowledge Discovery and Data Mining (KDD)}, pages 245--250, San
  Francisco, CA, 2001.

\bibitem[Broder(1997)]{broder1997resemblance}
Andrei~Z Broder.
\newblock On the resemblance and containment of documents.
\newblock In \emph{Proceedings of the Compression and Complexity of Sequences
  (SEQUENCES)}, pages 21--29, Salerno, Italy, 1997.

\bibitem[Broder et~al.(1997)Broder, Glassman, Manasse, and
  Zweig]{broder1997syntactic}
Andrei~Z. Broder, Steven~C. Glassman, Mark~S. Manasse, and Geoffrey Zweig.
\newblock Syntactic clustering of the web.
\newblock \emph{Comput. Networks}, 29\penalty0 (8-13):\penalty0 1157--1166,
  1997.

\bibitem[Broder et~al.(1998)Broder, Charikar, Frieze, and
  Mitzenmacher]{broder1998min}
Andrei~Z. Broder, Moses Charikar, Alan~M. Frieze, and Michael Mitzenmacher.
\newblock Min-wise independent permutations.
\newblock In \emph{Proceedings of the Thirtieth Annual {ACM} Symposium on the
  Theory of Computing (STOC)}, pages 327--336, Dallas, TX, 1998.

\bibitem[Brown et~al.(2020)Brown, Mann, Ryder, Subbiah, Kaplan, Dhariwal,
  Neelakantan, Shyam, Sastry, Askell, Agarwal, Herbert-Voss, Krueger, Henighan,
  Child, Ramesh, Ziegler, Wu, Winter, Hesse, Chen, Sigler, Litwin, Gray, Chess,
  Clark, Berner, McCandlish, Radford, Sutskever, and Amodei]{brown2020language}
Tom Brown, Benjamin Mann, Nick Ryder, Melanie Subbiah, Jared~D Kaplan, Prafulla
  Dhariwal, Arvind Neelakantan, Pranav Shyam, Girish Sastry, Amanda Askell,
  Sandhini Agarwal, Ariel Herbert-Voss, Gretchen Krueger, Tom Henighan, Rewon
  Child, Aditya Ramesh, Daniel Ziegler, Jeffrey Wu, Clemens Winter, Chris
  Hesse, Mark Chen, Eric Sigler, Mateusz Litwin, Scott Gray, Benjamin Chess,
  Jack Clark, Christopher Berner, Sam McCandlish, Alec Radford, Ilya Sutskever,
  and Dario Amodei.
\newblock Language models are few-shot learners.
\newblock In \emph{Advances in Neural Information Processing Systems
  (NeurIPS)}, pages 1877--1901, virtual, 2020.

\bibitem[Buhler(2001)]{buher2001effcient}
Jeremy Buhler.
\newblock Efficient large-scale sequence comparison by locality-sensitive
  hashing.
\newblock \emph{Bioinformatics}, 17\penalty0 (5):\penalty0 419--428, 2001.

\bibitem[Cand{\`{e}}s et~al.(2006)Cand{\`{e}}s, Romberg, and
  Tao]{candes2006robust}
Emmanuel~J. Cand{\`{e}}s, Justin~K. Romberg, and Terence Tao.
\newblock Robust uncertainty principles: exact signal reconstruction from
  highly incomplete frequency information.
\newblock \emph{{IEEE} Trans. Inf. Theory}, 52\penalty0 (2):\penalty0 489--509,
  2006.

\bibitem[Carter and Wegman(1977)]{carter1977universal}
Larry Carter and Mark~N. Wegman.
\newblock Universal classes of hash functions (extended abstract).
\newblock In \emph{Proceedings of the 9th Annual {ACM} Symposium on Theory of
  Computing (STOC)}, pages 106--112, Boulder, CO, 1977.

\bibitem[Chang et~al.(2020)Chang, Yu, Chang, Yang, and Kumar]{chang2020pre}
Wei{-}Cheng Chang, Felix~X. Yu, Yin{-}Wen Chang, Yiming Yang, and Sanjiv Kumar.
\newblock Pre-training tasks for embedding-based large-scale retrieval.
\newblock In \emph{Proceedings of the 8th International Conference on Learning
  Representations (ICLR)}, Addis Ababa, Ethiopia, 2020.

\bibitem[Charikar et~al.(2004)Charikar, Chen, and
  Farach-Colton]{charikar2004finding}
Moses Charikar, Kevin Chen, and Martin Farach-Colton.
\newblock Finding frequent items in data streams.
\newblock \emph{Theor. Comput. Sci.}, 312\penalty0 (1):\penalty0 3--15, 2004.

\bibitem[Charikar(2002)]{charikar2002similarity}
Moses~S Charikar.
\newblock Similarity estimation techniques from rounding algorithms.
\newblock In \emph{Proceedings of the Thiry-Fourth Annual ACM Symposium on
  Theory of Computing (STOC)}, pages 380--388, Montreal, Canada, 2002.

\bibitem[Chen et~al.(2017)Chen, Fisch, Weston, and Bordes]{chen2017reading}
Danqi Chen, Adam Fisch, Jason Weston, and Antoine Bordes.
\newblock Reading wikipedia to answer open-domain questions.
\newblock In \emph{Proceedings of the 55th Annual Meeting of the Association
  for Computational Linguistics (ACL)}, pages 1870--1879, Vancouver, Canada,
  2017.

\bibitem[Chen et~al.(2015)Chen, Wilson, Tyree, Weinberger, and
  Chen]{chen2015compressing}
Wenlin Chen, James Wilson, Stephen Tyree, Kilian Weinberger, and Yixin Chen.
\newblock {Compressing Neural Networks with the Hashing Trick}.
\newblock In \emph{Proceedings of the 32nd International Conference on Machine
  Learning (ICML)}, pages 2285--2294, Lille, France, 2015.

\bibitem[Chierichetti et~al.(2009)Chierichetti, Kumar, Lattanzi, Mitzenmacher,
  Panconesi, and Raghavan]{chierichetti2009compressing}
Flavio Chierichetti, Ravi Kumar, Silvio Lattanzi, Michael Mitzenmacher,
  Alessandro Panconesi, and Prabhakar Raghavan.
\newblock On compressing social networks.
\newblock In \emph{Proceedings of the 15th {ACM} {SIGKDD} International
  Conference on Knowledge Discovery and Data Mining (KDD)}, pages 219--228,
  Paris, France, 2009.

\bibitem[Das et~al.(2007)Das, Datar, Garg, and Rajaram]{das2007google}
Abhinandan Das, Mayur Datar, Ashutosh Garg, and Shyamsundar Rajaram.
\newblock Google news personalization: scalable online collaborative filtering.
\newblock In \emph{Proceedings of the 16th International Conference on World
  Wide Web (WWW)}, pages 271--280, Banff, Alberta, Canada, 2007.

\bibitem[Dasgupta(2000)]{dasgupta2000experiments}
Sanjoy Dasgupta.
\newblock Experiments with random projection.
\newblock In \emph{Proceedings of the 16th Conference in Uncertainty in
  Artificial Intelligence (UAI)}, pages 143--151, Stanford, CA, 2000.

\bibitem[Dasgupta and Freund(2008)]{dasgupta2008random}
Sanjoy Dasgupta and Yoav Freund.
\newblock Random projection trees and low dimensional manifolds.
\newblock In \emph{Proceedings of the 40th Annual {ACM} Symposium on Theory of
  Computing (STOC)}, pages 537--546, Victoria, Canada, 2008.

\bibitem[Datar et~al.(2004)Datar, Immorlica, Indyk, and
  Mirrokni]{datar2004locality}
Mayur Datar, Nicole Immorlica, Piotr Indyk, and Vahab~S Mirrokni.
\newblock Locality-sensitive hashing scheme based on p-stable distributions.
\newblock In \emph{Proceedings of the Twentieth Annual Symposium on
  Computational Geometry (SCG)}, pages 253--262, Brooklyn, NY, 2004.

\bibitem[Devlin et~al.(2019)Devlin, Chang, Lee, and Toutanova]{devlin2019bert}
Jacob Devlin, Ming{-}Wei Chang, Kenton Lee, and Kristina Toutanova.
\newblock {BERT:} pre-training of deep bidirectional transformers for language
  understanding.
\newblock In \emph{Proceedings of the 2019 Conference of the North American
  Chapter of the Association for Computational Linguistics: Human Language
  Technologies (NAACL-HLT)}, pages 4171--4186, Minneapolis, MN, 2019.

\bibitem[Donoho(2006)]{donoho2006compressed}
David~L. Donoho.
\newblock Compressed sensing.
\newblock \emph{{IEEE} Trans. Inf. Theory}, 52\penalty0 (4):\penalty0
  1289--1306, 2006.

\bibitem[Dwork et~al.(2006)Dwork, McSherry, Nissim, and
  Smith]{dwork2006calibrating}
Cynthia Dwork, Frank McSherry, Kobbi Nissim, and Adam~D. Smith.
\newblock Calibrating noise to sensitivity in private data analysis.
\newblock In \emph{Proceedings of the Third Theory of Cryptography Conference
  (TCC)}, pages 265--284, New York, NY, 2006.

\bibitem[Fan et~al.(2019)Fan, Guo, Zhu, Miao, Sun, and Li]{fan2019mobius}
Miao Fan, Jiacheng Guo, Shuai Zhu, Shuo Miao, Mingming Sun, and Ping Li.
\newblock {MOBIUS:} towards the next generation of query-ad matching in baidu's
  sponsored search.
\newblock In \emph{Proceedings of the 25th {ACM} {SIGKDD} International
  Conference on Knowledge Discovery {\&} Data Mining (KDD)}, pages 2509--2517,
  Anchorage, AK, 2019.

\bibitem[Fern and Brodley(2003)]{fern2003random}
Xiaoli~Zhang Fern and Carla~E. Brodley.
\newblock Random projection for high dimensional data clustering: A cluster
  ensemble approach.
\newblock In \emph{Proceedings of the Twentieth International Conference
  (ICML)}, pages 186--193, Washington, DC, 2003.

\bibitem[Giorgi et~al.(2021)Giorgi, Nitski, Wang, and Bader]{giorgi2021declutr}
John~M. Giorgi, Osvald Nitski, Bo~Wang, and Gary~D. Bader.
\newblock Declutr: Deep contrastive learning for unsupervised textual
  representations.
\newblock In \emph{Proceedings of the 59th Annual Meeting of the Association
  for Computational Linguistics and the 11th International Joint Conference on
  Natural Language Processing, {ACL/IJCNLP}}, pages 879--895, Virtual Event,
  2021.

\bibitem[Goemans and Williamson(1995)]{goemans1995improved}
Michel~X. Goemans and David~P. Williamson.
\newblock Improved approximation algorithms for maximum cut and satisfiability
  problems using semidefinite programming.
\newblock \emph{J. {ACM}}, 42\penalty0 (6):\penalty0 1115--1145, 1995.

\bibitem[Haddadpour et~al.(2020)Haddadpour, Karimi, Li, and
  Li]{haddadpour2020fedsketch}
Farzin Haddadpour, Belhal Karimi, Ping Li, and Xiaoyun Li.
\newblock Fedsketch: Communication-efficient and private federated learning via
  sketching.
\newblock \emph{arXiv preprint arXiv:2008.04975}, 2020.

\bibitem[Huang et~al.(2013)Huang, He, Gao, Deng, Acero, and
  Heck]{huang2013learning}
Po{-}Sen Huang, Xiaodong He, Jianfeng Gao, Li~Deng, Alex Acero, and Larry~P.
  Heck.
\newblock Learning deep structured semantic models for web search using
  clickthrough data.
\newblock In \emph{Proceedings of the 22nd {ACM} International Conference on
  Information and Knowledge Management (CIKM)}, pages 2333--2338, San
  Francisco, CA, 2013.

\bibitem[Huang et~al.(2019)Huang, Zhang, Li, and Li]{huang2019knowledge}
Xiao Huang, Jingyuan Zhang, Dingcheng Li, and Ping Li.
\newblock Knowledge graph embedding based question answering.
\newblock In \emph{Proceedings of the Twelfth {ACM} International Conference on
  Web Search and Data Mining (WSDM)}, pages 105--113, Melbourne, Australia,
  2019.

\bibitem[Indyk(1999)]{indyk1999sublinear}
Piotr Indyk.
\newblock Sublinear time algorithms for metric space problems.
\newblock In Jeffrey~Scott Vitter, Lawrence~L. Larmore, and Frank~Thomson
  Leighton, editors, \emph{Proceedings of the Thirty-First Annual {ACM}
  Symposium on Theory of Computing (STOC)}, pages 428--434, Atlanta, GA, 1999.

\bibitem[Johnson and Lindenstrauss(1984)]{johnson1984extensions}
William~B. Johnson and Joram Lindenstrauss.
\newblock Extensions of \text{Lipschitz} mapping into \text{Hilbert} space.
\newblock \emph{Contemporary Mathematics}, 26:\penalty0 189--206, 1984.

\bibitem[Karpathy et~al.(2014)Karpathy, Joulin, and
  Fei{-}Fei]{karpathy2014deep}
Andrej Karpathy, Armand Joulin, and Li~Fei{-}Fei.
\newblock Deep fragment embeddings for bidirectional image sentence mapping.
\newblock In \emph{Advances in Neural Information Processing Systems (NIPS)},
  pages 1889--1897, Montreal, Canada, 2014.

\bibitem[Lanchantin et~al.(2021)Lanchantin, Wang, Ordonez, and
  Qi]{lanchantin2021general}
Jack Lanchantin, Tianlu Wang, Vicente Ordonez, and Yanjun Qi.
\newblock General multi-label image classification with transformers.
\newblock In \emph{Proceedings of the {IEEE} Conference on Computer Vision and
  Pattern Recognition (CVPR)}, pages 16478--16488, virtual, 2021.

\bibitem[Li and Church(2005)]{li2005using}
Ping Li and Kenneth~Ward Church.
\newblock Using sketches to estimate associations.
\newblock In \emph{Proceedings of the Human Language Technology Conference and
  the Conference on Empirical Methods in Natural Language Processing
  (HLT/EMNLP)}, pages 708--715, Vancouver, Canada,
  \url{https://github.com/pltrees/Smallest-K-Sketch}, 2005.

\bibitem[Li and Li(2023)]{li2023differential}
Ping Li and Xiaoyun Li.
\newblock Differential privacy with random projections and sign random
  projections.
\newblock \emph{arXiv preprint}, 2023.

\bibitem[Li et~al.(2006{\natexlab{a}})Li, Hastie, and Church]{li2006improving}
Ping Li, Trevor Hastie, and Kenneth~Ward Church.
\newblock Improving random projections using marginal information.
\newblock In \emph{Proceedings of the 19th Annual Conference on Learning Theory
  (COLT)}, pages 635--649, Pittsburgh, PA, 2006{\natexlab{a}}.

\bibitem[Li et~al.(2006{\natexlab{b}})Li, Hastie, and Church]{li2006very}
Ping Li, Trevor~J Hastie, and Kenneth~W Church.
\newblock Very sparse random projections.
\newblock In \emph{Proceedings of the 12th ACM SIGKDD international conference
  on Knowledge discovery and data mining (KDD)}, pages 287--296, Philadelphia,
  PA, 2006{\natexlab{b}}.

\bibitem[Li et~al.(2008)Li, Church, and Hastie]{li2008one}
Ping Li, Kenneth Church, and Trevor Hastie.
\newblock One sketch for all: Theory and application of conditional random
  sampling.
\newblock In \emph{Advances in Neural Information Processing Systems (NIPS)},
  pages 953--960, Vancouver, Canada, 2008.

\bibitem[Li et~al.(2011)Li, Shrivastava, Moore, and K{\"{o}}nig]{li2011hashing}
Ping Li, Anshumali Shrivastava, Joshua~L. Moore, and Arnd~Christian
  K{\"{o}}nig.
\newblock Hashing algorithms for large-scale learning.
\newblock In \emph{Advances in Neural Information Processing Systems (NIPS)},
  pages 2672--2680, Granada, Spain, 2011.

\bibitem[Li et~al.(2012)Li, Owen, and Zhang]{li2012one}
Ping Li, Art~B Owen, and Cun-Hui Zhang.
\newblock One permutation hashing.
\newblock In \emph{Advances in Neural Information Processing Systems (NIPS)},
  pages 3122--3130, Lake Tahoe, NV, 2012.

\bibitem[Li et~al.(2014)Li, Mitzenmacher, and Shrivastava]{li2014coding}
Ping Li, Michael Mitzenmacher, and Anshumali Shrivastava.
\newblock Coding for random projections.
\newblock In \emph{Proceedings of the 31th International Conference on Machine
  Learning (ICML)}, pages 676--684, Beijing, China, 2014.

\bibitem[Li and Li(2019{\natexlab{a}})]{li2019generalization}
Xiaoyun Li and Ping Li.
\newblock Generalization error analysis of quantized compressive learning.
\newblock In \emph{Advances in Neural Information Processing Systems
  (NeurIPS)}, pages 15124--15134, Vancouver, Canada, 2019{\natexlab{a}}.

\bibitem[Li and Li(2019{\natexlab{b}})]{li2019random}
Xiaoyun Li and Ping Li.
\newblock Random projections with asymmetric quantization.
\newblock In \emph{Advances in Neural Information Processing Systems
  (NeurIPS)}, pages 10857--10866, Vancouver, Canada, 2019{\natexlab{b}}.

\bibitem[Li and Li(2021)]{li2021one}
Xiaoyun Li and Ping Li.
\newblock One-sketch-for-all: Non-linear random features from compressed linear
  measurements.
\newblock In \emph{Proceedings of the 24th International Conference on
  Artificial Intelligence and Statistics (AISTATS)}, pages 2647--2655, Virtual
  Event, 2021.

\bibitem[Li and Li(2022)]{li2022c}
Xiaoyun Li and Ping Li.
\newblock {C-MinHash}: Improving minwise hashing with circulant permutation.
\newblock In \emph{Proceedings of the International Conference on Machine
  Learning (ICML)}, pages 12857--12887, Baltimore, MD, 2022.

\bibitem[Malkov and Yashunin(2020)]{malkov2020efficient}
Yury~A. Malkov and Dmitry~A. Yashunin.
\newblock Efficient and robust approximate nearest neighbor search using
  hierarchical navigable small world graphs.
\newblock \emph{{IEEE} Trans. Pattern Anal. Mach. Intell.}, 42\penalty0
  (4):\penalty0 824--836, 2020.

\bibitem[Nargesian et~al.(2018)Nargesian, Zhu, Pu, and
  Miller]{nargesian2018table}
Fatemeh Nargesian, Erkang Zhu, Ken~Q. Pu, and Ren{\'{e}}e~J. Miller.
\newblock Table union search on open data.
\newblock \emph{Proc. {VLDB} Endow.}, 11\penalty0 (7):\penalty0 813--825, 2018.

\bibitem[Neelakantan et~al.(2022)Neelakantan, Xu, Puri, Radford, Han, Tworek,
  Yuan, Tezak, Kim, Hallacy, et~al.]{neelakantan2022text}
Arvind Neelakantan, Tao Xu, Raul Puri, Alec Radford, Jesse~Michael Han, Jerry
  Tworek, Qiming Yuan, Nikolas Tezak, Jong~Wook Kim, Chris Hallacy, et~al.
\newblock Text and code embeddings by contrastive pre-training.
\newblock \emph{arXiv preprint arXiv:2201.10005}, 2022.

\bibitem[Nissim et~al.(2007)Nissim, Raskhodnikova, and Smith]{nissim2007smooth}
Kobbi Nissim, Sofya Raskhodnikova, and Adam~D. Smith.
\newblock Smooth sensitivity and sampling in private data analysis.
\newblock In \emph{Proceedings of the 39th Annual {ACM} Symposium on Theory of
  Computing (STOC)}, pages 75--84, San Diego, CA, 2007.

\bibitem[Pennington et~al.(2014)Pennington, Socher, and
  Manning]{pennington2014glove}
Jeffrey Pennington, Richard Socher, and Christopher~D. Manning.
\newblock Glove: Global vectors for word representation.
\newblock In \emph{Proceedings of the 2014 Conference on Empirical Methods in
  Natural Language Processing (EMNLP)}, pages 1532--1543, Doha, Qatar, 2014.

\bibitem[Rabanser et~al.(2019)Rabanser, G{\"{u}}nnemann, and
  Lipton]{rabanser2019failing}
Stephan Rabanser, Stephan G{\"{u}}nnemann, and Zachary~C. Lipton.
\newblock Failing loudly: An empirical study of methods for detecting dataset
  shift.
\newblock In \emph{Advances in Neural Information Processing Systems
  (NeurIPS)}, pages 1394--1406, Vancouver, Canada, 2019.

\bibitem[Rahimi and Recht(2007)]{rahimi2007random}
Ali Rahimi and Benjamin Recht.
\newblock Random features for large-scale kernel machines.
\newblock In \emph{Advances in Neural Information Processing Systems (NIPS)},
  pages 1177--1184, Vancouver, Canada, 2007.

\bibitem[Ram and Gray(2012)]{ram2012maximum}
Parikshit Ram and Alexander~G Gray.
\newblock Maximum inner-product search using cone trees.
\newblock In \emph{Proceedings of the 18th {ACM} {SIGKDD} International
  Conference on Knowledge Discovery and Data Mining (KDD)}, pages 931--939,
  Beijing, China, 2012.

\bibitem[Rothchild et~al.(2020)Rothchild, Panda, Ullah, Ivkin, Stoica,
  Braverman, Gonzalez, and Arora]{rothchild2020fetchsgd}
Daniel Rothchild, Ashwinee Panda, Enayat Ullah, Nikita Ivkin, Ion Stoica,
  Vladimir Braverman, Joseph Gonzalez, and Raman Arora.
\newblock {FetchSGD}: Communication-efficient federated learning with
  sketching.
\newblock In \emph{Proceedings of the 37th International Conference on Machine
  Learning (ICML)}, pages 8253--8265, Virtual Event, 2020.

\bibitem[Shrivastava(2016)]{shrivastava2016simple}
Anshumali Shrivastava.
\newblock Simple and efficient weighted minwise hashing.
\newblock In \emph{Neural Information Processing Systems (NIPS)}, pages
  1498--1506, Barcelona, Spain, 2016.

\bibitem[Shrivastava and Li(2014)]{shrivastava2014asymmetric}
Anshumali Shrivastava and Ping Li.
\newblock Asymmetric {LSH} {(ALSH)} for sublinear time maximum inner product
  search {(MIPS)}.
\newblock In \emph{Advances in Neural Information Processing Systems (NIPS)},
  pages 2321--2329, Montreal, Canada, 2014.

\bibitem[Singhal et~al.(2021)Singhal, Sidahmed, Garrett, Wu, Rush, and
  Prakash]{singhal2021federated}
Karan Singhal, Hakim Sidahmed, Zachary Garrett, Shanshan Wu, John Rush, and
  Sushant Prakash.
\newblock Federated reconstruction: Partially local federated learning.
\newblock In \emph{Advances in Neural Information Processing Systems
  (NeurIPS)}, virtual, 2021.

\bibitem[Soria{-}Comas et~al.(2017)Soria{-}Comas, Domingo{-}Ferrer,
  S{\'{a}}nchez, and Meg{\'{\i}}as]{comas2017individual}
Jordi Soria{-}Comas, Josep Domingo{-}Ferrer, David S{\'{a}}nchez, and David
  Meg{\'{\i}}as.
\newblock Individual differential privacy: {A} utility-preserving formulation
  of differential privacy guarantees.
\newblock \emph{{IEEE} Trans. Inf. Forensics Secur.}, 12\penalty0 (6):\penalty0
  1418--1429, 2017.

\bibitem[Spillo et~al.(2022)Spillo, Musto, de~Gemmis, Lops, and
  Semeraro]{spillo2022knowledge}
Giuseppe Spillo, Cataldo Musto, Marco de~Gemmis, Pasquale Lops, and Giovanni
  Semeraro.
\newblock Knowledge-aware recommendations based on neuro-symbolic graph
  embeddings and first-order logical rules.
\newblock In \emph{Proceedings of the Sixteenth {ACM} Conference on Recommender
  Systems (RecSys)}, pages 616--621, Seattle, WA, 2022.

\bibitem[Tamersoy et~al.(2014)Tamersoy, Roundy, and Chau]{tamersoy2014guilt}
Acar Tamersoy, Kevin~A. Roundy, and Duen~Horng Chau.
\newblock Guilt by association: large scale malware detection by mining
  file-relation graphs.
\newblock In \emph{Proceedings of the 20th {ACM} {SIGKDD} International
  Conference on Knowledge Discovery and Data Mining (KDD)}, pages 1524--1533,
  New York, NY, 2014.

\bibitem[Tan et~al.(2021)Tan, Xu, Zhao, Fei, Zhou, and Li]{tan2021norm}
Shulong Tan, Zhaozhuo Xu, Weijie Zhao, Hongliang Fei, Zhixin Zhou, and Ping Li.
\newblock Norm adjusted proximity graph for fast inner product retrieval.
\newblock In \emph{Proceedings of the 27th {ACM} {SIGKDD} Conference on
  Knowledge Discovery and Data Mining (KDD)}, pages 1552--1560, Virtual Event,
  Singapore, 2021.

\bibitem[Tomita et~al.(2020)Tomita, Browne, Shen, Chung, Patsolic, Falk,
  Priebe, Yim, Burns, Maggioni, and Vogelstein]{tomita2020sparse}
Tyler~M. Tomita, James Browne, Cencheng Shen, Jaewon Chung, Jesse Patsolic,
  Benjamin Falk, Carey~E. Priebe, Jason Yim, Randal~C. Burns, Mauro Maggioni,
  and Joshua~T. Vogelstein.
\newblock Sparse projection oblique randomer forests.
\newblock \emph{J. Mach. Learn. Res.}, 21:\penalty0 104:1--104:39, 2020.

\bibitem[Wang et~al.(2019)Wang, Qi, Zhang, Zhai, Wang, Lui, and
  Guan]{wang2019memory}
Pinghui Wang, Yiyan Qi, Yuanming Zhang, Qiaozhu Zhai, Chenxu Wang, John C.~S.
  Lui, and Xiaohong Guan.
\newblock A memory-efficient sketch method for estimating high similarities in
  streaming sets.
\newblock In \emph{Proceedings of the 25th {ACM} {SIGKDD} International
  Conference on Knowledge Discovery {\&} Data Mining (KDD)}, pages 25--33,
  Anchorage, AK, 2019.

\bibitem[Weinberger et~al.(2009)Weinberger, Dasgupta, Langford, Smola, and
  Attenberg]{weinberger2009feature}
Kilian~Q. Weinberger, Anirban Dasgupta, John Langford, Alexander~J. Smola, and
  Josh Attenberg.
\newblock Feature hashing for large scale multitask learning.
\newblock In \emph{Proceedings of the 26th Annual International Conference on
  Machine Learning (ICML)}, pages 1113--1120, Montreal, Canada, 2009.

\bibitem[Wu et~al.(2019)Wu, He, and Xu]{wu2019demo}
Jun Wu, Jingrui He, and Jiejun Xu.
\newblock {DEMO-Net}: Degree-specific graph neural networks for node and graph
  classification.
\newblock In \emph{Proceedings of the 25th {ACM} {SIGKDD} International
  Conference on Knowledge Discovery {\&} Data Mining (KDD)}, pages 406--415,
  Anchorage, AK, 2019.

\bibitem[Yu et~al.(2018)Yu, Zhao, Zheng, Zhang, and You]{yu2018hierarchical}
Chaojian Yu, Xinyi Zhao, Qi~Zheng, Peng Zhang, and Xinge You.
\newblock Hierarchical bilinear pooling for fine-grained visual recognition.
\newblock In \emph{Proceedings of the 15th European Conference on Computer
  Vision (ECCV), Part {XVI}}, pages 595--610, Munich, Germany, 2018.

\bibitem[Yu et~al.(2022{\natexlab{a}})Yu, Jin, Liu, Yang, Fei, and
  Li]{yu2022boost}
Tan Yu, Zhipeng Jin, Jie Liu, Yi~Yang, Hongliang Fei, and Ping Li.
\newblock Boost {CTR} prediction for new advertisements via modeling visual
  content.
\newblock In \emph{Proceedings of the {IEEE} International Conference on Big
  Data (IEEE BigData)}, Osaka, Japan, 2022{\natexlab{a}}.

\bibitem[Yu et~al.(2022{\natexlab{b}})Yu, Liu, Yang, Li, Fei, and
  Li]{yu2022egm}
Tan Yu, Jie Liu, Yi~Yang, Yi~Li, Hongliang Fei, and Ping Li.
\newblock {EGM:} enhanced graph-based model for large-scale video advertisement
  search.
\newblock In \emph{Proceedings of the 28th {ACM} {SIGKDD} Conference on
  Knowledge Discovery and Data Mining (KDD)}, pages 4443--4451, Washington, DC,
  2022{\natexlab{b}}.

\bibitem[Zhang et~al.(2022)Zhang, Wang, Murray, and
  Koniusz]{zhang2022kernelized}
Shan Zhang, Lei Wang, Naila Murray, and Piotr Koniusz.
\newblock Kernelized few-shot object detection with efficient integral
  aggregation.
\newblock In \emph{Proceedings of the {IEEE/CVF} Conference on Computer Vision
  and Pattern Recognition (CVPR)}, pages 19185--19194, New Orleans, LA, 2022.

\bibitem[Zhang et~al.(2020)Zhang, Qi, and Wang]{zhang2020dynamic}
Zhaoqi Zhang, Panpan Qi, and Wei Wang.
\newblock Dynamic malware analysis with feature engineering and feature
  learning.
\newblock In \emph{Proceedings of the Thirty-Fourth {AAAI} Conference on
  Artificial Intelligence (AAAI)}, pages 1210--1217, New York, NY, 2020.

\bibitem[Zhao et~al.(2020)Zhao, Tan, and Li]{zhao2020song}
Weijie Zhao, Shulong Tan, and Ping Li.
\newblock {SONG:} approximate nearest neighbor search on {GPU}.
\newblock In \emph{Proceedings of the 36th {IEEE} International Conference on
  Data Engineering (ICDE)}, pages 1033--1044, Dallas, TX, 2020.

\bibitem[Zhou et~al.(2019)Zhou, Tan, Xu, and Li]{zhou2019mobius}
Zhixin Zhou, Shulong Tan, Zhaozhuo Xu, and Ping Li.
\newblock M{\"{o}}bius transformation for fast inner product search on graph.
\newblock In \emph{Advances in Neural Information Processing Systems
  (NeurIPS)}, pages 8216--8227, Vancouver, Canada, 2019.

\end{thebibliography}

\end{document}